\documentclass{article}

\PassOptionsToPackage{numbers, compress}{natbib}
\usepackage{amssymb}
\usepackage{amsmath}
\usepackage{amsthm}
\usepackage{graphicx}
\usepackage{wrapfig}
\usepackage{subcaption}
\usepackage{enumitem}
\usepackage[dvipsnames]{xcolor}
\usepackage{fontawesome5}

\newtheorem{thm}{Theorem}[section]
\newtheorem{lemma}{Lemma}[section]
\newtheorem{proposition}{Proposition}[section]

\usepackage[final]{neurips_2025}

\usepackage[utf8]{inputenc} 
\usepackage[T1]{fontenc}    
\usepackage{hyperref}       
\usepackage{url}            
\usepackage{booktabs}       
\usepackage{amsfonts}       
\usepackage{nicefrac}       
\usepackage{microtype}      
\usepackage{bm}

\hypersetup{
    colorlinks=true,
    citecolor=blue,
    linkcolor=red,
    urlcolor=teal
}

\setlist[itemize]{
  topsep=0pt,    
  itemsep=2pt,   
  parsep=0.5pt,    
  partopsep=0pt, 
  leftmargin=2.5em 
}

\newcommand*{\tran}{^{\mkern-1.5mu\mathsf{T}}}
\newcommand*{\dd}{\mathrm{d}}

\newcommand{\vx}{\bm{\mathrm{x}}} 
\newcommand{\vz}{\bm{\mathrm{z}}}
\newcommand{\vU}{\bm{\mathrm{U}}} 
\newcommand{\va}{\bm{\mathrm{a}}} 
\newcommand{\vv}{\bm{\mathrm{v}}} 
\newcommand{\vy}{\bm{\mathrm{y}}} 
\newcommand{\vA}{\bm{\mathrm{A}}} 
\newcommand{\vW}{\bm{\mathrm{W}}}
\newcommand{\vs}{\bm{\mathrm{s}}} 
\newcommand{\vB}{\bm{\mathrm{B}}}
\newcommand{\vV}{\bm{\mathrm{V}}}
\newcommand{\vJ}{\bm{\mathrm{J}}}
\newcommand{\vX}{\bm{\mathrm{X}}} 
\newcommand{\vZ}{\bm{\mathrm{Z}}} 
\newcommand{\vS}{\bm{\mathrm{S}}} 
\newcommand{\veta}{\bm{\eta}}
\newcommand{\vSigma}{\boldsymbol{\Sigma}}
\newcommand{\vI}{\bm{I}} 
\newcommand{\vxi}{\bm{\xi}}
\newcommand{\vtheta}{\bm{\theta}}
\newcommand{\vmu}{\bm{\mu}}
\newcommand{\vOmega}{\bm{\Omega}}
\newcommand{\vG}{\bm{\mathrm{G}}}
\newcommand{\vQ}{\bm{\mathrm{Q}}}
\newcommand{\vR}{\bm{\mathrm{R}}}
\newcommand{\sqSigmat}{\Sigma^{1/2}_t}
\newcommand{\sqSigma}{\Sigma^{1/2}}
\DeclareMathOperator{\Tr}{Tr}
\title{
Why Diffusion Models Don’t Memorize:  The Role of Implicit Dynamical Regularization in Training  
}

\author{%
    Tony Bonnaire$^\dagger$ \\
    LPENS \\
    Université PSL, Paris\\
    \texttt{tony.bonnaire@phys.ens.fr} \\
    \And
    Raphaël Urfin$^\dagger$ \\
    LPENS \\
    Université PSL, Paris \\
    \texttt{raphael.urfin@phys.ens.fr} \\
    \AND
    Giulio Biroli \\
    LPENS \\
    Université PSL, Paris\\
    \texttt{giulio.biroli@phys.ens.fr} \\
    \And
    Marc Mézard \\
    Department of Computing Sciences \\
    Bocconi University, Milano\\
    \texttt{marc.mezard@unibocconi.it} \\
}

\begin{document}

\def\thefootnote{$\dagger$}\footnotetext{Equal contribution.} 

\maketitle

\begin{abstract}
Diffusion models have achieved remarkable success across a wide range of generative tasks. A key challenge is understanding the mechanisms that prevent their memorization of training data and allow generalization. In this work, we investigate the role of the training dynamics in the transition from generalization to memorization. Through extensive experiments and theoretical analysis, we identify two distinct timescales: an early time $\tau_\mathrm{gen}$ at which models begin to generate high-quality samples, and a later time $\tau_\mathrm{mem}$ beyond which memorization emerges. Crucially, we find that $\tau_\mathrm{mem}$ increases linearly with the training set size $n$, while $\tau_\mathrm{gen}$ remains constant. This creates a growing window of training times with $n$ where models generalize effectively, despite showing strong memorization if training continues beyond it. It is only when $n$ becomes larger than a model-dependent threshold that overfitting disappears at infinite training times.
These findings reveal a form of implicit dynamical regularization in the training dynamics, which allow to avoid memorization even in highly overparameterized settings. Our results are supported by numerical experiments with standard U-Net architectures on realistic and synthetic  datasets, and by a theoretical analysis using a tractable random features model studied in the high-dimensional limit. \href{https://github.com/tbonnair/Why-Diffusion-Models-Don-t-Memorize}{{\textcolor{blue}\faGithub}}
\end{abstract}


\section{Introduction}

Diffusion Models \cite[DMs,][]{sohl-dickstein_15, ho2020, song2019, song2021b} achieve state-of-the-art performance in a wide variety of AI tasks such as the generation of images \cite{DALL-E}, audios \cite{survey_audio}, videos \cite{sora2024}, and scientific data \cite{Biferale_2024, Price_2025}.
This class of generative models, inspired by out-of-equilibrium thermodynamics \cite{sohl-dickstein_15}, corresponds to a two-stage process: the first one, called \emph{forward}, gradually adds noise to a data, whereas the second one, called \emph{backward}, generates new data by denoising Gaussian white noise samples.
In DMs, the reverse process typically involves solving a stochastic differential equation (SDE) with a force field called \emph{score}. However, it is also possible to define a deterministic transport through an ordinary differential equation (ODE), treating the score as a velocity field, an approach that is for instance followed in flow matching \cite{lipman_2023}.

Understanding the generalization properties of score-based generative methods is a central issue in machine learning, and a particularly important question is how memorization of the training set is avoided in practice. A model without regularization achieving zero training loss only learns the empirical score, and is bound to reproduce samples of the training dataset at the end of the backward process. This memorization regime \cite{li_2024_good_score, Biroli_2024} is empirically observed when the training set is small and disappears when it increases beyond a model-dependent threshold \cite{kadkhodaie_2024}. Understanding the mechanisms controlling this change of regimes from memorization to generalization is a central challenge for both theory and applications. 
Model regularization and inductive biases imposed by the network architecture were shown to play a role \cite{Kamb2024, Shah2025}, as well as a dynamical regularization due to the finiteness of the learning rate \cite{Wu2025}.
However, the regime shift described above is consistently observed even in models where all these regularization mechanisms are present. This suggests that the core mechanism behind the transition from memorization to generalization lies elsewhere. In this work, we demonstrate -- first through numerical experiments, and then via the theoretical analysis of a simplified model -- that this transition is driven by an implicit dynamical bias towards generalizing solutions emerging in the training, which allows to avoid the memorization phase.

\begin{figure}
    \centering
    \includegraphics[width=.49\linewidth]{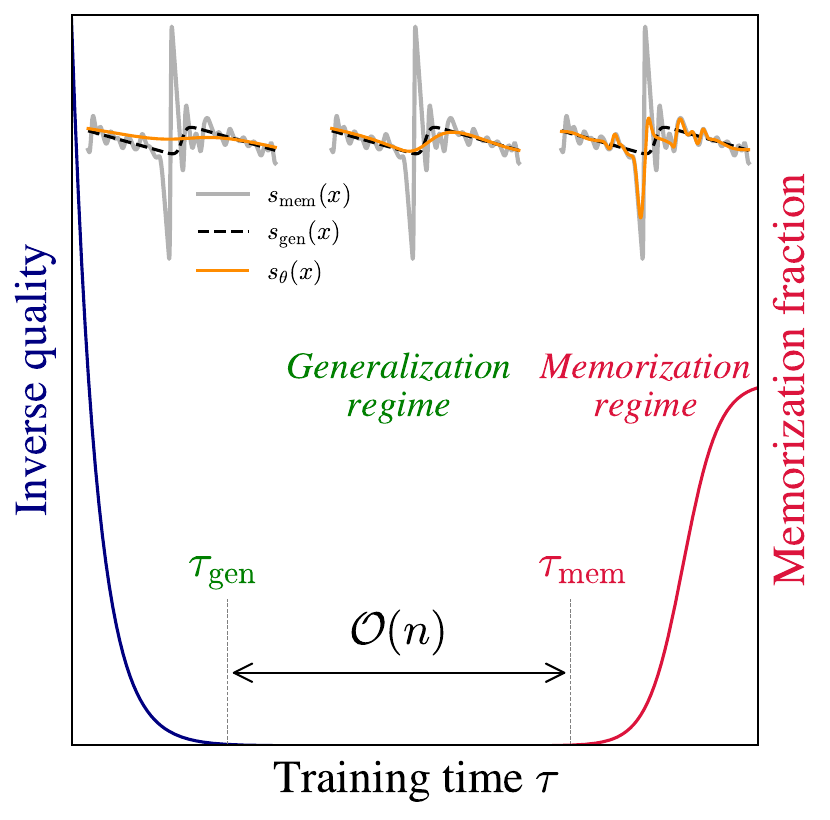}
    \includegraphics[width=.49\linewidth]{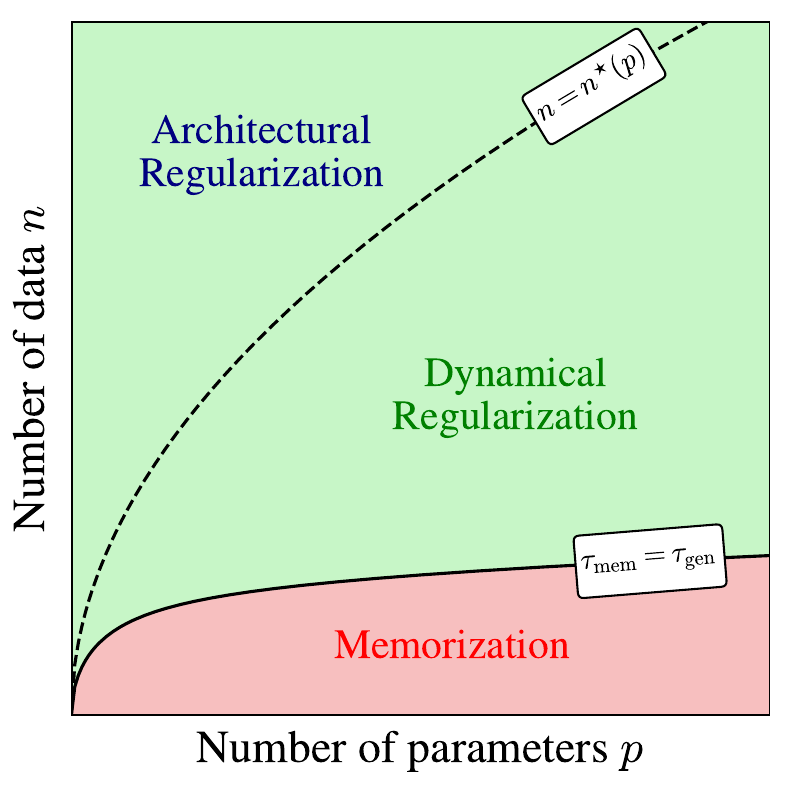}
    \caption{ \textbf{Qualitative summary of our contributions.} \textit{(Left)} Illustration of the training dynamics of a diffusion model. Depending on the training time $\tau$, we identify three regimes measured by the inverse quality of the generated samples (blue curve) and their memorization fraction (red curve). The generalization regime extends over a large window of training times which increases with the training set size $n$. On top, we show a one dimensional example of the learned score function during training (orange). The gray line gives the exact empirical score, at a given noise level, while the black dashed line corresponds to the true (population) score.
    \textit{(Right)} Phase diagram in the $(n,p)$ plane illustrating three regimes of diffusion models: \textcolor{Red}{Memorization} when $n$ is sufficiently small at fixed $p$, \textcolor{NavyBlue}{Architectural Regularization} for $n>n^{\star}(p)$ (which is model and dataset dependent, as discussed in \cite{george_2025, Kamb2024}), and \textcolor{ForestGreen}{Dynamical Regularization}, corresponding to a large intermediate generalization regime obtained when the training dynamics is stopped early, i.e. $\tau \in \left[\tau_\mathrm{gen}, \tau_\mathrm{mem}\right]$.}
    \label{fig:summary_results}
\end{figure}

\paragraph*{Contributions and theoretical picture.} We investigate  the  dynamics of score learning using gradient descent, both numerically and analytically, and study the generation properties of the score depending on the time $\tau$ at which the training is stopped. The theoretical picture built from our results and combining several findings from the recent literature is illustrated in Fig.~\ref{fig:summary_results}. The two main parameters are the size of the training set $n$ and the expressivity of the class of score functions on which one trains the model, characterized by a number of parameters $p$; when both $n$ and $p$ are large one can identify three main regimes. Given $p$, if $n$ is larger than $n^*(p)$ (which depends on the training set and on the class of scores), the score model is not expressive enough to represent the empirical score associated to $n$ data, and instead provides a smooth interpolation, approximately independent of the training set. In this regime, even with a very large training time $\tau\to\infty$, memorization does not occur because the model is regularized by its architecture and the finite number of parameters. 
When $n<n^*(p)$ the model is expressive enough to memorize, and two timescales emerge during training: one, $\tau_\mathrm{gen}$, is the minimum training time required to achieve high-quality data generation; the second, $\tau_\mathrm{mem}>\tau_\mathrm{gen}$, signals when further training induces memorization, and causes the model to increasingly reproduce the training samples (left panel).
The first timescale, $\tau_\mathrm{gen}$, is found independent of $n$, whereas the second, $\tau_\mathrm{mem}$, grows approximately linearly with $n$, thus opening a large window of training times during which the model generalizes if early stopped when $\tau \in [\tau_\mathrm{gen},\tau_\mathrm{mem}]$.
Our results shows that implicit dynamical regularization in training plays a crucial role in score-based generative models, substantially enlarging the generalization regime (see right panel of Fig.~\ref{fig:summary_results}),  and hence allowing to avoid memorization even in highly overparameterized settings. 
We find that the key mechanism behind the widening gap between $\tau_\mathrm{gen}$ and $\tau_\mathrm{mem}$ is the irregularity of the empirical score at low noise level and large $n$. In this regime the models used to approximate the score provide a smooth interpolation that remains stable for a long period of training times and closely approximates the population score, a behavior likely rooted in the spectral bias of neural networks \cite{rahaman2019spectral}. 
Only at very long training times do the dynamics converge to the low lying minimum corresponding to the empirical score, leading to memorization (as illustrated in the one-dimensional examples in the left panel of Fig.~\ref{fig:summary_results}).

The theoretical picture described above is based on our numerical and analytical results, and builds up on previous works, in particular numerical analysis characterizing the memorization--generalization transition \cite{gu2023memorization,yoon2023diffusion}, analytical works on memorization of DMs \cite{george_2025,Kamb2024, kadkhodaie_2024}, and studies on the spectral bias of deep neural networks \cite{rahaman2019spectral}.
Our numerical experiments use a class of scores based on a realistic U-Net \cite{Ronneberger2015} trained on downscaled images of the CelebA dataset \cite{CelebA}. By varying $n$ and $p$, we measure the evolution of the sample quality (through FID) and the fraction of memorization during learning, which support the theoretical scenario presented in Fig.~\ref{fig:summary_results}. Additional experimental results on synthetic data are provided in Supplemental Material (SM, Sects.~\ref{appendix:numerical_details} and~\ref{appendix:numerical_GMM}).
On the analytical side, we focus on a class of scores constructed from random features and simplified models of data, following \cite{george_2025}. 
In this setting, the timescales of training dynamics correspond directly to the inverse eigenvalues of the random feature correlation matrix. Leveraging tools from random matrix theory, we compute the spectrum in the limit of large datasets, high-dimensional data, and overparameterized models. This analysis reveals, in a fully tractable way, how the theoretical picture of Fig.~\ref{fig:summary_results} emerges within the random feature framework.

\paragraph*{Related works.}
\begin{itemize}
    \item The memorization transition in DMs has been the subject of several recent empirical investigations \cite{Carlini_2023, somepalli_2022, somepalli_2023} which have demonstrated that state-of-the-art image DMs -- including Stable Diffusion and DALL·E -- can reproduce a non-negligible portion of their training data, indicating a form of memorization. Several additional works \cite{gu2023memorization, yoon2023diffusion} examined how this phenomenon is influenced by factors such as data distribution, model configuration, and training procedure, and provide a strong basis for the numerical part of our work.
    \item A series of theoretical studies in the high-dimensional regime have analyzed the memorization--generalization transition during the generative dynamics under the empirical score assumption \cite{Biroli_2024, achilli2024, ventura2025}, showing how trajectories are attracted to the training samples. Within this high-dimensional framework, \cite{cui_2024, cui_2025, wang2024, george_2025} study the score learning for various model classes. In particular, \cite{george_2025} uses a Random Feature Neural Network \cite{Rahimi_2007}. The authors compute the asymptotic training and test losses for $\tau\rightarrow\infty$ and relate it to memorization.
    The theoretical part of our work generalizes this approach to study the role of training dynamics and early stopping in the memorization--generalization transition.
    \item Recent works have also uncovered complementary sources of implicit regularization explaining how DMs avoid memorization. Architectural biases and limited network capacity were for instance shown to constrain memorization in \cite{Kamb2024, kadkhodaie_2024}, and finiteness of the learning rate prevents the model from learning the empirical score in \cite{Wu2025}. Also related to our analysis, \cite{li2025generalizationpropertiesdiffusionmodels} provides general bounds showing the beneficial role of early stopping the training dynamics to enhance generalization for finitely supported target distributions, as well as a study of its effect for one-dimensional gaussian mixtures.
    \item Finally, previous studies on supervised learning \cite{rahaman2019spectral, Zhi_Qin_John_Xu_2020}, and more recently on DMs \cite{Wang2025}, have shown that deep neural networks display a frequency-dependent learning speed, and hence a learning bias towards low frequency functions. 
    This fact plays an important role in the results we present since the empirical score contains a low frequency part that is close to the population score, and a high-frequency part that is dataset-dependent. To the best of our knowledge, the training time to learn the high-frequency part and hence memorize, that we find to scale with $n$, has not been studied from this perspective in the context of score-based generative methods.
\end{itemize}

\paragraph*{Setting: generative diffusion and score learning.}
Standard DMs define a transport from a target distribution $P_0$ in $\mathbb{R}^d$ to a Gaussian white noise $\mathcal{N}(0, \bm{I}_d)$ through a \emph{forward process} defined as an Ornstein-Uhlenbeck (OU) stochastic differential equation (SDE):
\begin{align} \label{eq:forward}
    \dd \vx = -\vx(t)\dd t + \dd \vB(t),
\end{align}
where $\dd \vB(t)$ is square root of two times a Wiener process.
Generation is performed by time-reversing the SDE \eqref{eq:forward} using the score function
$\vs(\vx,t) = \nabla_{\vx} \log P_t(\vx)$,
\begin{align}
    -\dd \vx = \left[\vx(t) + 2 \vs(\vx,t) \right] \dd t + \dd \vB(t),
    \label{eq:backward}
\end{align}
where $P_t(\vx)$ is the probability density at time $t$ along the forward process, and the noise $\dd \vB(t)$ is also the square root of two times a Wiener process.
As shown in the seminal works \cite{hyvarinen_05, Vincent_2011}, $\vs(\vx,t)$ can be obtained by minimizing the score matching loss
\begin{align}
\label{eq:SML}
    \hat{\vs}(\vx,t)=
    \arg\min_{\vs} \mathbb{E}_{\vx \sim P_0,\vxi \sim \mathcal{N}(0, \bm{I}_d)} \left[\lVert \sqrt{\Delta_t}\vs(\vx(t), t)+\vxi\rVert^2 \right],
\end{align}
where $\Delta_t=1-e^{-2t}$. In practice, the optimization problem is restricted to a parametrized class of functions $\vs_{\vtheta}(\vx(t), t)$ defined, for example, by a neural network with parameters $\vtheta$. The expectation over $\vx$ is replaced by the empirical average over the training set ($n$ iid samples $\vx^\nu$ drawn from $P_0$),
\begin{align} \label{eq:loss_t}
    \mathcal{L}_t(\vtheta,\{\vx^\nu\}_{\nu=1}^n) = \frac{1}{n} \sum_{\nu=1}^n \mathbb{E}_{\vxi \sim \mathcal{N}(0, \bm{I}_d)} \left[\lVert \sqrt{\Delta_t}\vs_{\vtheta}(\vx^\nu(t))+\vxi\rVert^2\right],
\end{align}
where $\vx^\nu_t(\vxi)=e^{-t} \vx^\nu+\sqrt{\Delta_t}\vxi$. The loss in (\ref{eq:loss_t})
can be minimized with standard optimizers, such as stochastic gradient descent \cite[SGD,][]{robbins1951stochastic} or Adam \cite{kingma2015Adam}. In practice, a single model conditioned on the diffusion time $t$ is trained by integrating (\ref{eq:loss_t}) over time \cite{karras2022elucidatingdesignspacediffusionbased}.
The solution of the minimization of \eqref{eq:loss_t} is the so-called empirical score (e.g. \cite{Biroli_2024, li_2024_good_score}), defined as
$\vs_{\mathrm{emp}}(\vx,t) = \nabla_{\vx} \log P_t^\mathrm{emp}(\vx)$, with
\begin{equation} \label{eq:P_t_emp}
    P_t^\mathrm{emp}(\vx) = \frac{1}{n\left(2\pi\Delta_t\right)^{d/2}} \sum_{\nu=1}^n e^{-\frac{1}{2\Delta_t} \lVert \vx- \vx^{\nu} e^{-t} \rVert^2_2}.
\end{equation}
This solution is known to inevitably recreate samples of the training set at the end of the generative process (i.e., it perfectly memorizes), unless $n$ grows exponentially with the dimension $d$ \cite{Biroli_2024}. However, this is not the case in many practical applications where memorization is only observed for relatively small values of $n$, and disappears well before $n$ becomes exponentially large in $d$. The empirical minimization performed in practice, within a given class of models and a given minimization procedure, does \emph{not} drive the optimization to the global minimum of \eqref{eq:loss_t}, but instead to a smoother estimate of the score that is independent of the training set with good generalization properties \cite{kadkhodaie_2024}, as the global minimum of \eqref{eq:SML} would do. 
Understanding how it is possible, and in particular the role played by the training dynamics to avoid memorization, is the central aim of the present work.

\section{Generalization and memorization during training of diffusion models} \label{sect:Numerical_results}

\begin{figure}
    \centering
    \includegraphics[width=.378\linewidth]{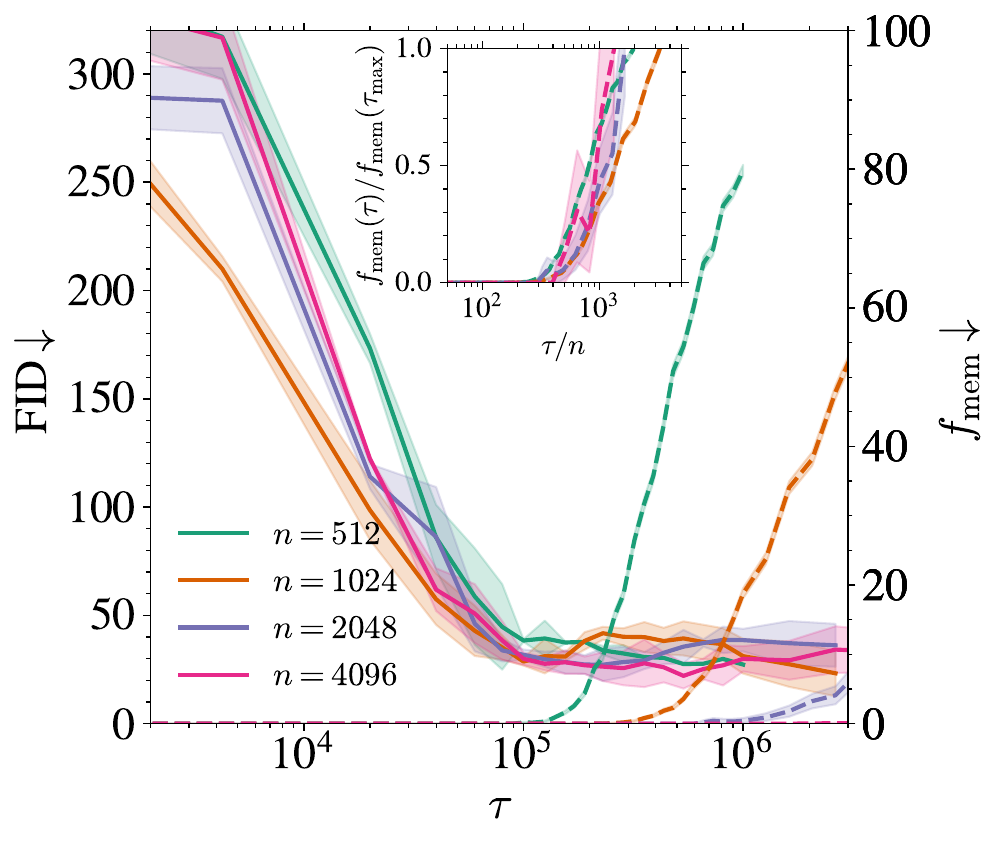}
    \includegraphics[width=.335\linewidth]{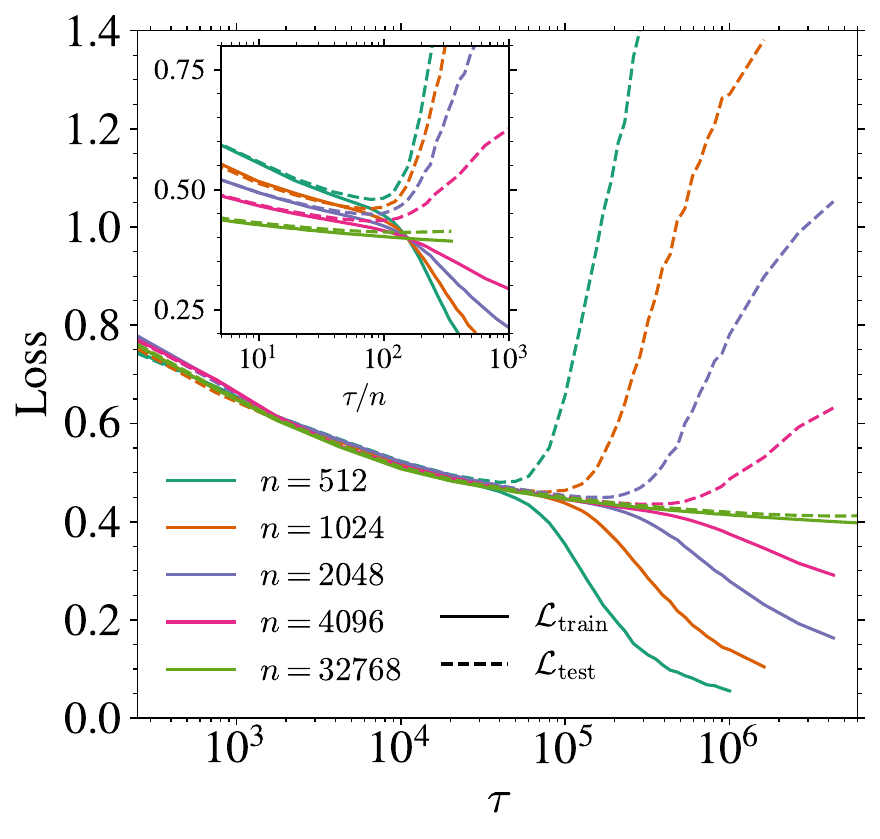}
    \includegraphics[height=0.2155\textheight]{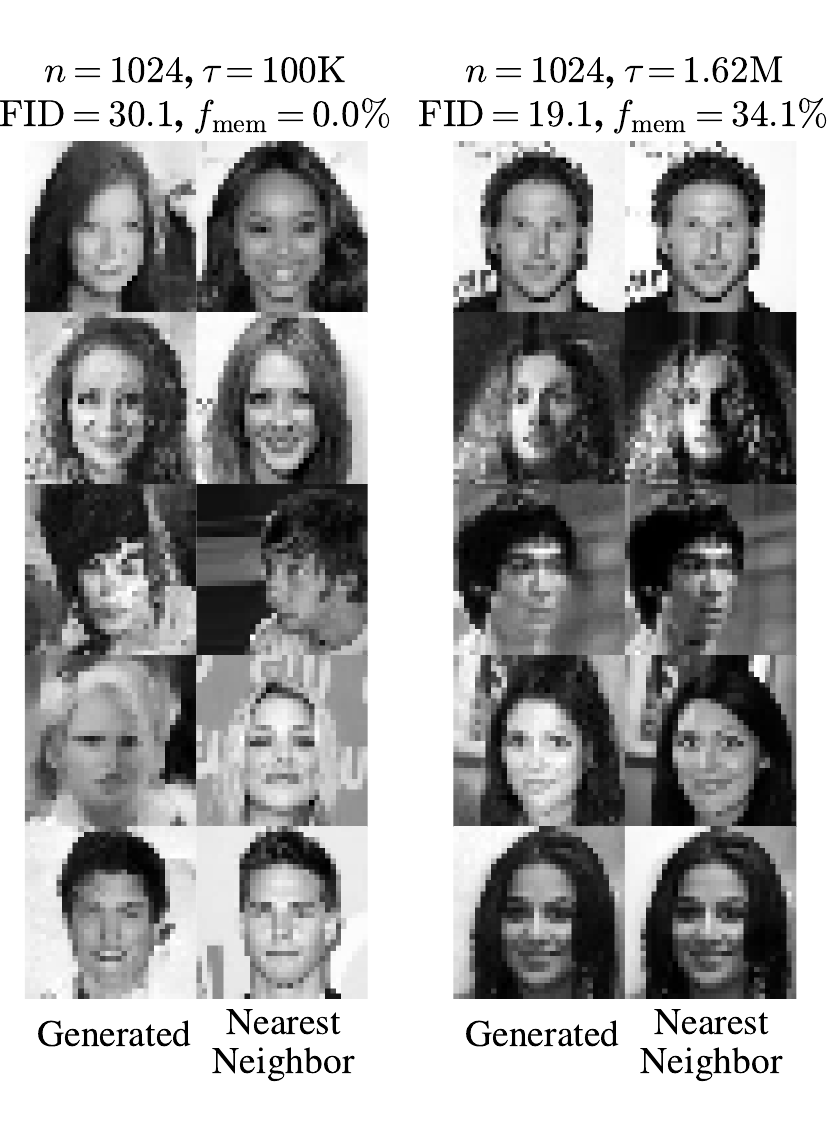}
    \caption{\textbf{Memorization transition as a function of the training set size $n$ for U-Net score models on CelebA.} \textit{(Left)} FID (solid lines, left axis) and memorization fraction $f_\mathrm{mem}$ (dashed lines, right axis) against training time $\tau$ for various $n$. Inset: normalized memorization fraction $f_\mathrm{mem}(\tau)/f_\mathrm{mem}(\tau_\mathrm{max})$ with the rescaled time $\tau/n$. \textit{(Middle)} Training (solid lines) and test (dashed lines) loss with $\tau$ for several $n$ at fixed $t=0.01$. Inset: both losses plotted against $\tau/n$. Error bars on the losses are imperceptible. \textit{(Right)} Generated samples from the model trained with $n=1024$ for $\tau=100$K or $\tau=1.62$M steps, along with their nearest neighbors in the training set.
    }
    \label{fig:numerical_memorization_n}
\end{figure}

\paragraph*{Data \& architecture.} We conduct our experiments on the CelebA face dataset \cite{CelebA}, which we convert to grayscale downsampled images of size $d=32\times32$, and vary the training set size $n$ from 128 up to 32768. Our score model has a U-Net architecture \cite{Ronneberger2015} with three resolution levels and a base channel width of $W$ with multipliers 1, 2 and 3 respectively. All our networks are DDPMs \cite{ho2020} trained to predict the injected noise at diffusion time $t$ using SGD with momentum at fixed batch size $\min(n, 512)$. The models are all conditioned on $t$, i.e. a single model approximates the score at all times, and make use of a standard sinusoidal position embedding \cite{Vaswani2017} that is added to the features of each resolution. More details about the numerical setup can be found in SM (Sect.~\ref{appendix:numerical_details}).

\paragraph*{Evaluation metrics.} 
To study the transition from generalization to memorization during training, we monitor the loss~\eqref{eq:loss_t} during training using a fixed diffusion time $t = 0.01$. At various numbers of SGD updates $\tau$, we compute the loss on all $n$ training examples (training loss) and on a held-out test set of 2048 images (test loss). 
To characterize the score obtained after a training time $\tau$, we assess the originality and quality of samples by generating 10K samples using a DDIM accelerated sampling~\cite{song2022DDIM}.
We compute (i) the Fréchet-Inception Distance \cite[FID,][]{heusel2017gans} against 10K test samples which we use to identify the generalization time $\tau_\mathrm{gen}$; and (ii) the fraction of memorized generated samples $f_\mathrm{mem}(\tau)$ granting access to $\tau_\mathrm{mem}$, the memorization time. Following previous numerical studies \cite{yoon2023diffusion, gu2023memorization}, a generated sample $\vx_\tau$ is considered memorized if
\begin{equation} \label{eq:memorization_criterion}
    \mathbb{E}_{\vx_\tau} \left[ \frac{\lVert \vx_\tau - \va^{\mu_1}\rVert_2}{\lVert \vx_\tau - \va^{\mu_2}\rVert_2} \right] < k,
\end{equation}
where $\va^{\mu_1}$ and $\va^{\mu_2}$ are the nearest and second nearest neighbors of $\vx_\tau$ in the training set in the $L_2$ sense. In what follows, we choose to work with $k=1/3$ \cite{yoon2023diffusion, gu2023memorization}, but we checked that varying $k$ to $1/2$ or $1/4$ does not impact the claims about the scaling.
Error bars in the figures correspond to twice the standard deviation over 5 different test sets for FIDs, and 5 noise realizations for $\mathcal{L}_\mathrm{train}$ and $\mathcal{L}_\mathrm{test}$. For $f_\mathrm{mem}$, we report the 95\% CIs on the mean evaluated with 1,000 bootstrap samples.

\paragraph*{Role of training set size on the learning dynamics.}
At fixed model capacity ($p=4\times 10^6$, base width $W=32$), we investigate how the training set size $n$ impacts the previous metrics. In the left panel of Fig.~\ref{fig:numerical_memorization_n}, we first report the FID (solid lines) and $f_\mathrm{mem}(\tau)$ (dashed lines) for various $n$. All trainings dynamics exhibit two phases. First, the FID quickly decreases to reach a minimum value on a timescale $\tau_\mathrm{gen}$ ($\approx100$K) that does not depend on $n$. In the right panel, the generated samples at $\tau=100$K clearly differ from their nearest neighbors in the training set, indicating that the model generalizes correctly. Beyond this time, the FID remains flat. $f_\mathrm{mem}(\tau)$ is zero until a later time $\tau_\mathrm{mem}$ after which it increases, clearly signaling the entrance into a memorization regime, as illustrated by the generated samples in the right-most panel of Fig.~\ref{fig:numerical_memorization_n}, very close to their nearest neighbors. Both the transition time $\tau_\mathrm{mem}$ and the value of the final fraction $f_\mathrm{mem}(\tau_\mathrm{max})$ (with $\tau_\mathrm{max}$ being one to four million SGD steps) vary with $n$. The inset plot shows the normalized memorization fraction $f_\mathrm{mem}(\tau)/f_\mathrm{mem}(\tau_\mathrm{max})$ against the rescaled time $\tau/n$, making all curves collapse and increase at around $\tau/n \approx 300$, showing that $\tau_\mathrm{mem} \propto n$, and demonstrating the existence of a generalization window for $\tau \in \left[\tau_\mathrm{gen}, \tau_\mathrm{mem}\right]$ that widens linearly with $n$, as illustrated in the left panel of Fig.~\ref{fig:summary_results}.

As highlighted in the introduction, memorization in DMs is ultimately driven by the overfitting of the empirical score $\vs_\mathrm{mem}(\vx,t)$. The evolution of $\mathcal{L}_\mathrm{train}(\tau)$ and $\mathcal{L}_\mathrm{test}(\tau)$ at fixed $t=0.01$ are shown in the middle panel of Fig.~\ref{fig:numerical_memorization_n} for $n$ ranging from 512 to 32768. Initially, the two losses remain nearly indistinguishable, indicating that the learned score $\vs_{\vtheta}(\vx, t)$ does not depend on the training set. Beyond a critical time, $\mathcal{L}_\mathrm{train}$ continues to decrease while $\mathcal{L}_\mathrm{test}$ increases, leading to a nonzero generalization loss whose magnitude depends on $n$. As $n$ increases, this critical time also increases and, eventually, the training and test loss gap shrinks: for $n=32768$, the test loss remains close to the training loss, even after 11 million SGD steps. The inset shows the evolution of both losses with $\tau/n$, demonstrating that the overfitting time scales linearly with the training set size $n$, just like $\tau_\mathrm{mem}$ identified in the left panel. Moreover, there is a consistent lag between the overfitting time and $\tau_\mathrm{mem}$ at fixed $n$, reflecting the additional training required for the model to overfit the empirical score sufficiently to reproduce the training samples, and therefore to impact the memorization fraction.

\paragraph*{Memorization is \emph{not} due to data repetition.} We must stress that this delayed memorization with $n$ is \emph{not} due to the mere repetition of training samples, as a first intuition could suggest. In SM Sects.~\ref{appendix:numerical_details} and~\ref{appendix:numerical_GMM}, we show that full-batch updates still yield $\tau_\mathrm{mem}\propto n$. In other words, even if at fixed $\tau$ all models have processed each sample equally often, larger $n$ consistently postpone memorization. This confirms that memorization in DMs is driven by a fundamental $n$-dependent change in the loss landscape -- not by a sample repetition during training.

\begin{figure}
    \centering
    \includegraphics[width=.48\linewidth]{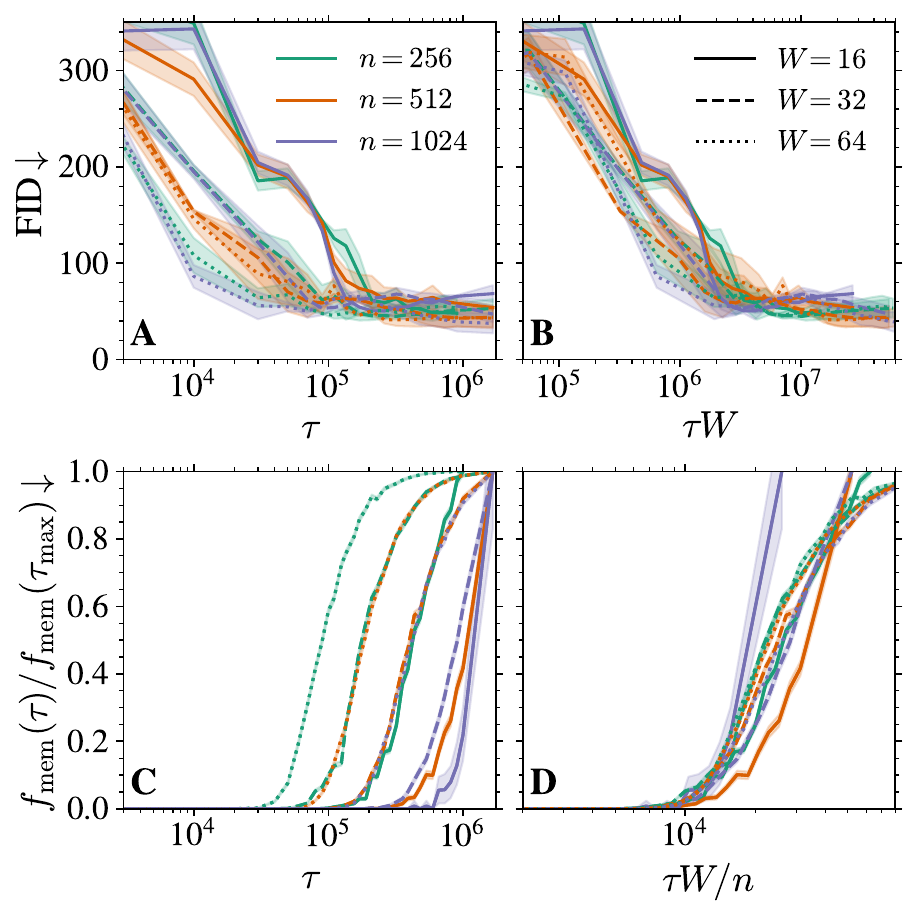}
    \includegraphics[width=.51\linewidth]{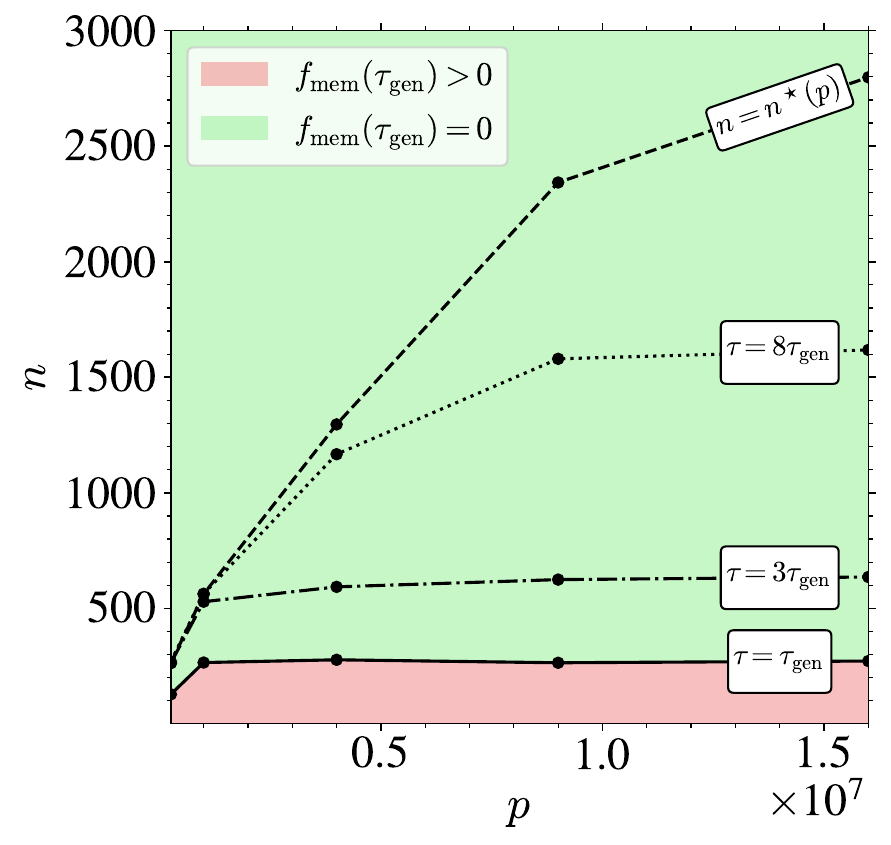}
    \caption{\textbf{Effect of the number of parameters in the U-Net architecture on the timescales of the training dynamics.} \textit{(Left)} FID (panels \textbf{A}, \textbf{B}) and normalized memorization fraction $f_\mathrm{mem}(\tau)/f_\mathrm{mem}(\tau_\mathrm{max})$ (panels \textbf{C}, \textbf{D}) for various $n$ and $W$ during training. In panels \textbf{B} and \textbf{D}, time is rescaled such that all curves collapse. \textit{(Right)} $(n,p)$ phase diagram of generalization vs memorization for U-Nets trained on CelebA. Curves show, for $\tau \in \{\tau_\mathrm{gen}, 3\tau_\mathrm{gen}, 8\tau_\mathrm{gen}\}$, the minimal dataset size $n(p)$ satisfying $f_\mathrm{mem}(\tau)=0$. The shaded background indicates the memorization--generalization boundary for $\tau=\tau_\mathrm{gen}$.
    }
    \label{fig:numerical_memorization_p}
\end{figure}

\paragraph*{Effect of the model capacity.} To study more precisely the role of the model capacity on the memorization--generalization transition, we vary the number of parameters $p$ by changing the U-Nets base width $W \in \{8, 16, 32, 48, 64\}$, resulting in a total of $p\in\{0.26, 1, 4, 9, 16\}\times10^6$ parameters. In the left panel of Fig.~\ref{fig:numerical_memorization_p}, we plot both the FID (top row) and the normalized memorization fraction (bottom row) as functions of $\tau$ for several width $W$ and training set sizes $n$. Panels \textbf{A} and \textbf{C} demonstrate that higher-capacity networks (larger $W$) achieve high-quality generation and begin to memorize \emph{earlier} than smaller ones. Panels \textbf{B} and \textbf{D} show that the two characteristic timescales simply scale as $\tau_\mathrm{gen} \propto W^{-1}$ and $\tau_\mathrm{mem} \propto nW^{-1}$. In particular, this implies that, for $W>8$, the critical training set size $n_\mathrm{gm}(p)$ at which $\tau_\mathrm{mem}=\tau_\mathrm{gen}$ is approximately independent of $p$ (at least on the limited values of $p$ we focused on).
When $n>n_\mathrm{gm}(p)$, the interval $\left[\tau_\mathrm{gen}, \tau_\mathrm{mem}\right]$ opens up, so that early stopping within this window yields high quality samples without memorization. In the right panel of Fig.~\ref{fig:numerical_memorization_p}, we display this boundary (solid line) in the $(n,p)$ plane by fixing the training time to $\tau=\tau_\mathrm{gen}$, that we identify numerically using the collapse of all FIDs at around $W\tau_\mathrm{gen}\approx 3\times 10^6$ (see panel \textbf{B}), and computing the smallest $n$ such that $f_\mathrm{mem}(\tau)=0$. The resulting solid curve delineates two regimes: below the curve, memorization already starts at $\tau_\mathrm{gen}$;  above the curve, the models generalize perfectly under early stopping. We repeat this experiment for $\tau=3\tau_\mathrm{gen}$ and $\tau=8\tau_\mathrm{gen}$, showing saturation to larger and larger $p$ as $\tau$ increases. Eventually, for $\tau \to \infty$, we expect these successive boundaries to converge to the architectural regularization threshold $n^\star(p)$, i.e. the point beyond which the network avoids memorization because it is not expressive enough, as found in \cite{george_2025} and highlighted in the right panel of Fig.~\ref{fig:summary_results}. 
In order to estimate $n^\star(p)$, we measure for a given $\tau$ the largest $n(\tau)$ yielding $f_\mathrm{mem}\approx0$. The curve $n(\tau)$ approaches $n^\star(p)$ for large $\tau$. We therefore estimate $n^\star(p)$ by measuring the asymptotic values of $n(\tau)$, which in practice is reached already at $\tau=\tau_\mathrm{max}=2$M updates for the values of $W$ we focus on.

\section{Training dynamics of a Random Features Network} \label{sect:Analytical_results}

\begin{figure}[!h]
    \centering
        \includegraphics[width=0.3\linewidth]{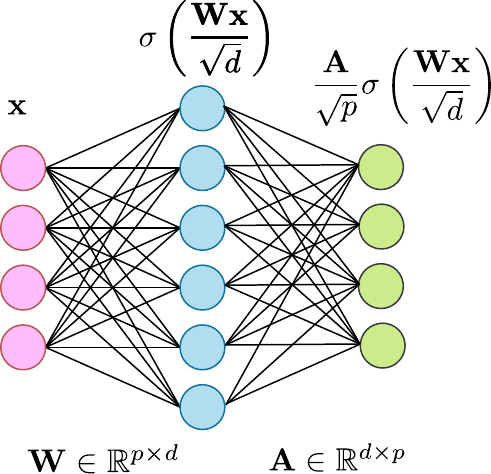} 
       \includegraphics[width=0.3\linewidth]{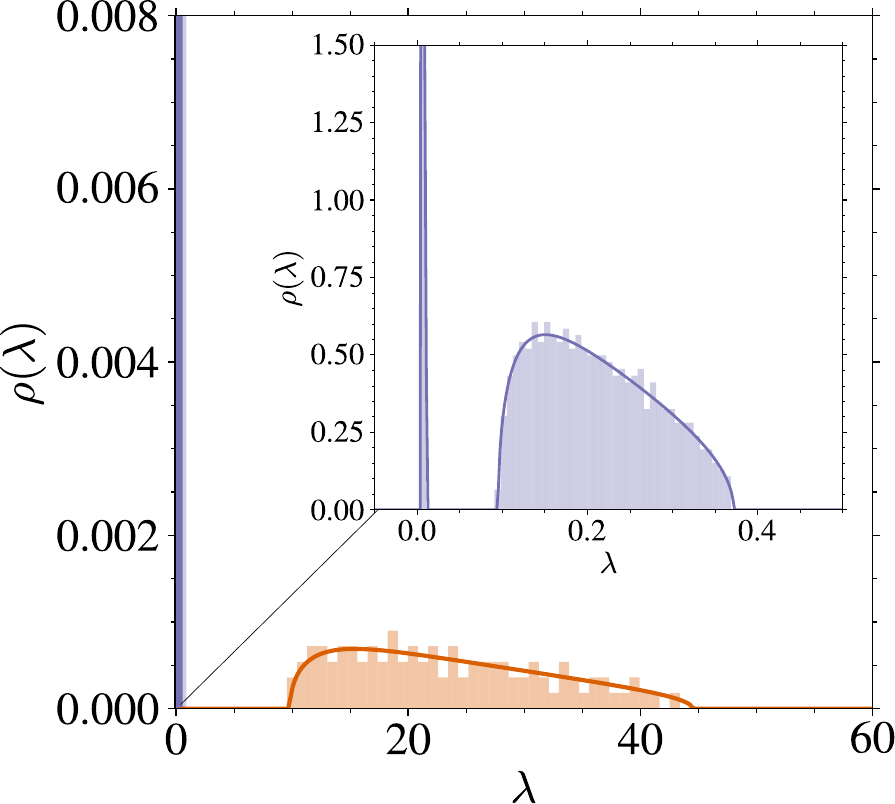} 
       \includegraphics[width=0.3\linewidth]{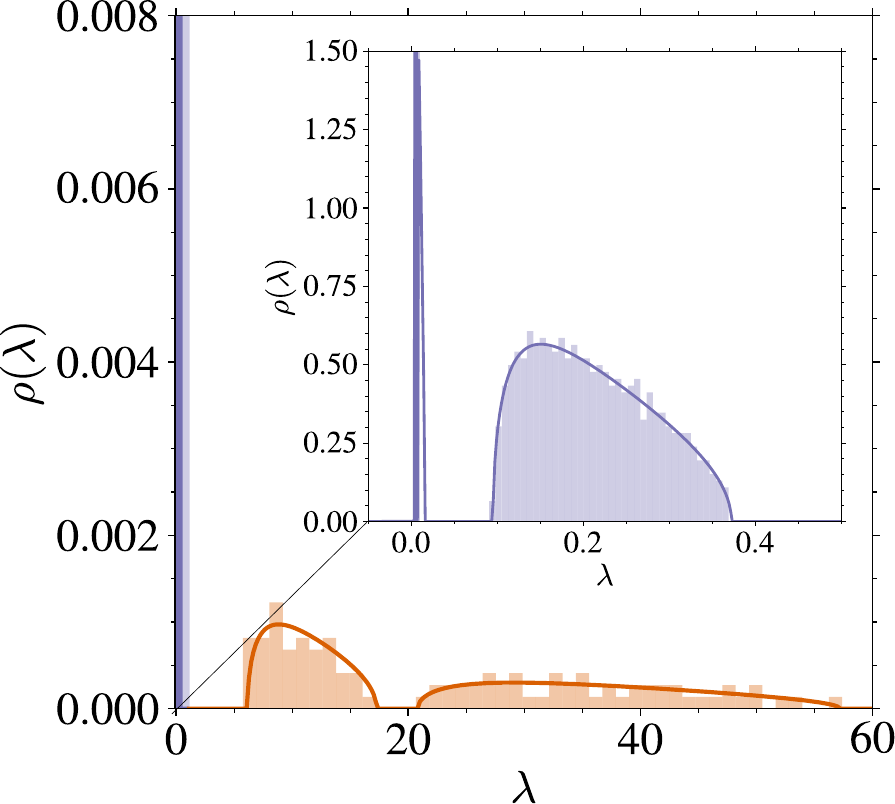} 
    \caption{(\textit{Left}) \textbf{Illustration of an RFNN.} (\textit{Middle/Right}) \textbf{Spectrum of $\vU$.}
    Density $\rho(\lambda)$ from Theorem~\ref{thm:Saddle_point_equations_new} in the overparameterized Regime~I described in Theorem~\ref{thm:Spectrum_new}, with $\psi_p = 64$, $\psi_n = 8$, $t = 0.01$, and $\rho_{\vSigma}(\lambda)=\delta(\lambda-1)$. The bulk of the spectrum (orange) is between $\lambda\approx10$ and $\lambda\approx45$. The histogram shows the eigenvalues from a single realization of $\vU$ at $d = 100$. Inset: zoom near $\lambda = 0$ (in blue) showing the first bulk $\rho_1$ and the delta peak at $\lambda = s_t^2$.
(\textit{Right}) Same as (\textit{Middle}), but with $\rho_{\vSigma}(\lambda) = \frac{1}{2}\delta(\lambda - 0.5) + \frac{1}{2}\delta(\lambda - 1.5)$. The first bulk in blue remains unchanged, as it depends only on $\sigma_{\vx}^2 = \Tr(\vSigma)/d = 1$ in both cases, while the second bulk varies with $\vSigma$.}
\label{fig:RFNN_plus_spectrum}
\end{figure}

\paragraph*{Notations.} We use bold symbols for vectors and matrices. The $L^2$ norm of a vector $\vx$ is denoted by $\lVert \vx \rVert = (\sum_i \vx_i^2)^{1/2}$. We write $f = \mathcal{O}(g)$ to mean that in the limit $n, p \to \infty$, there exists a constant $C$ such that $\lvert f \rvert \leq C \lvert g \rvert$. 
\paragraph*{Setting.}
We study analytically a model introduced in  \cite{george_2025}, where the data lie in $d$ dimensions. We
parametrize the score with a Random Features Neural Network \cite[RFNN,][]{Rahimi_2007}
\begin{align}
\label{random_features}
    \vs_{\vA}(\vx)=\frac{\vA}{\sqrt{p}}\sigma\left(\frac{\vW \vx}{\sqrt{d}}\right).
\end{align}
An RFNN, illustrated in Fig.~\ref{fig:RFNN_plus_spectrum} (left), is a two-layer neural-network whose first layer weights ($\vW\in\mathbb{R}^{p\times d}$) are drawn from a Gaussian distribution and remain frozen while the
second layer weights ($\vA \in \mathbb{R}^{d\times p}$) are learned during training. This model has already served as theoretical framework for studying several behaviors of deep neural network such as the double descent phenomenon \cite{Mei_2019, Ascoli_2020}. $\sigma$ is an element-wise non-linear activation function. We consider a training set of $n$ iid samples $\vx^\nu \sim P_{\vx}$ for $\nu = 1, \ldots, n$ and we focus on the high-dimensional limit  $d,p,n\rightarrow\infty$ with the ratios $\psi_p=p/d, \psi_n=n/d$ kept fixed. We study the training dynamics associated to the minimization of the empirical score matching loss defined in
(\ref{eq:loss_t}) at a fixed diffusion time $t$. This is a simplification compared to practical methods, which use a single model for all $t$. It has been already studied in previous theoretical works \cite{cui_2024, george_2025}. The loss (\ref{eq:loss_t}) is rescaled  by a factor $1/d$ in order to ensure a finite limit at large $d$. We also study the evolution of the test loss evaluated on test points and the distance to the exact score $\vs(\vx)=\nabla\log P_{\vx}$,
\begin{align}
    \label{test_loss_RF}
    \mathcal{L}_\mathrm{test}=\frac{1}{d}\mathbb{E}_{\vx,\vxi}\left[\lVert \sqrt{\Delta_t}\vs_{\vA}(\vx_t(\vxi))+\vxi\rVert ^2\right],\quad\mathcal{E}_\mathrm{score}=\frac{1}{d}\mathbb{E}_{\vx}\left[\lVert\vs_{\vA}(\vx)-\nabla\log P_{\vx}\rVert^2\right],
\end{align}
where the expectations $\mathbb{E}_{\vx,\vxi}$ are computed over $\vx\sim P_{\vx}$ and $\vxi \sim \mathcal{N}(0,\vI_d)$. The generalization loss, defined as $\mathcal{L}_\mathrm{gen} = \mathcal{L}_\mathrm{test} - \mathcal{L}_\mathrm{train}$, indicates the degree of overfitting in the model while the distance to the exact score $\mathcal{E}_\mathrm{score}$ measures the quality of the generation as it is an upper bound on the Kullback–Leibler divergence between the target and generated distributions \cite{song2021a, bortoli2022convergence}. The weights $\vA$ are updated via gradient descent
\begin{align}
\label{eq:GD_RF}
    \vA^{(k+1)}=\vA^{(k)}-\eta\nabla_{\vA}\mathcal{L}_\mathrm{train}(\vA^{(k)}),
\end{align}
where $\eta$ is the learning rate. In the high-dimensional limit, as the learning rate $\eta \to 0$, and after rescaling time as $\tau = k\eta / d^2$, the discrete-time dynamics converges to the following continuous-time gradient flow:
\begin{align}\label{eq:gda}
    \dot{\vA}(\tau)=-d^2\nabla_{\vA}\mathcal{L}_\mathrm{train}(\vA(\tau))
    =-2\Delta_t\frac{d}{p}\vA\vU-\frac{2d\sqrt{\Delta_t}}{\sqrt{p}}\vV^T,
\end{align}
with
\begin{align}
    \vU=\frac{1}{n}\sum_{\nu=1}^n\mathbb{E}_{\vxi}\left[\sigma\left(\frac{\vW \vx_t^\nu(\vxi)}{\sqrt{d}}\right)\sigma\left(\frac{\vW \vx_t^\nu(\vxi)}{\sqrt{d}}\right)^T\right],\quad \vV=\frac{1}{n}\sum_{\nu=1}^n\mathbb{E}_{\vxi}\left[\sigma\left(\frac{\vW \vx_t^\nu(\vxi)}{\sqrt{d}}\right)\vxi^T\right].
\end{align}

\paragraph*{Assumptions.} For our analytical results to hold, we make the following mathematical assumptions which are standard when studying Random Features \cite{Peche2019,goldt_2021, hu2023} namely (i) the activation function $\sigma$ admits a Hermite polynomial expansion $\sigma(x)=\sum_{s=0}^\infty\frac{\alpha_s}{s!}He_s(x)$; and (ii) the data distribution $P_{\vx}$ has sub-Gaussian tails and a covariance $\vSigma=\mathbb{E}_{P_{\vx}}[\vx \vx^T]$ with  bounded spectrum. We assume that the empirical 
distribution of eigenvalues of $\vSigma$ converges weakly in the high dimensional limit to a deterministic density $\rho_{\vSigma}(\lambda)$
and that $\Tr(\vSigma)/d$ converges to a finite limit  (for a more precise mathematical statement see SM Sect.~\ref{appendix:subsec:GEP}). Moreover, we make additional assumptions that are not essential to the proofs but which simplify the analysis: (iii) the activation function $\sigma$ verifies $\mu_0=\mathbb{E}_z[\sigma(z)]=0$; and (iv) the second layer $\vA$ is initialized with zero weights $\vA(\tau=0)=0$. In numerical applications, unless specified, we use $\sigma(z)=\tanh(z)$ and $P_{\vx}=\mathcal{N}(0,\vI_d)$.

\begin{figure}[!ht]
    \centering
    \begin{subfigure}[b]{0.35\linewidth}
       \centering
        \includegraphics[height=\linewidth, width=\linewidth]{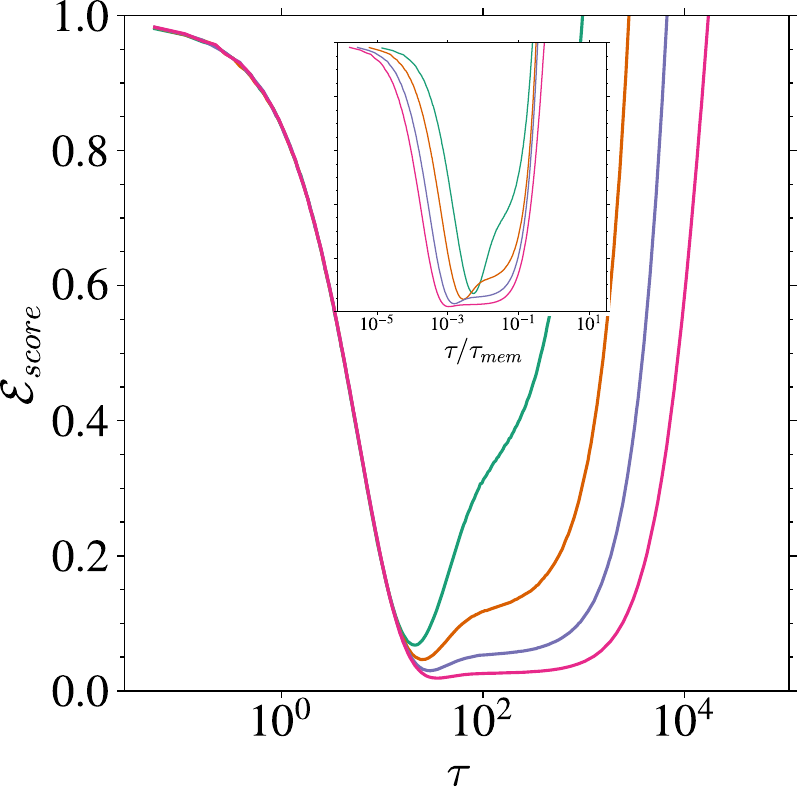} 
        \caption{}
    \end{subfigure}
    \hfill
    \begin{subfigure}[b]{0.35\linewidth}
       \centering
       \includegraphics[height=\linewidth, width=\linewidth]{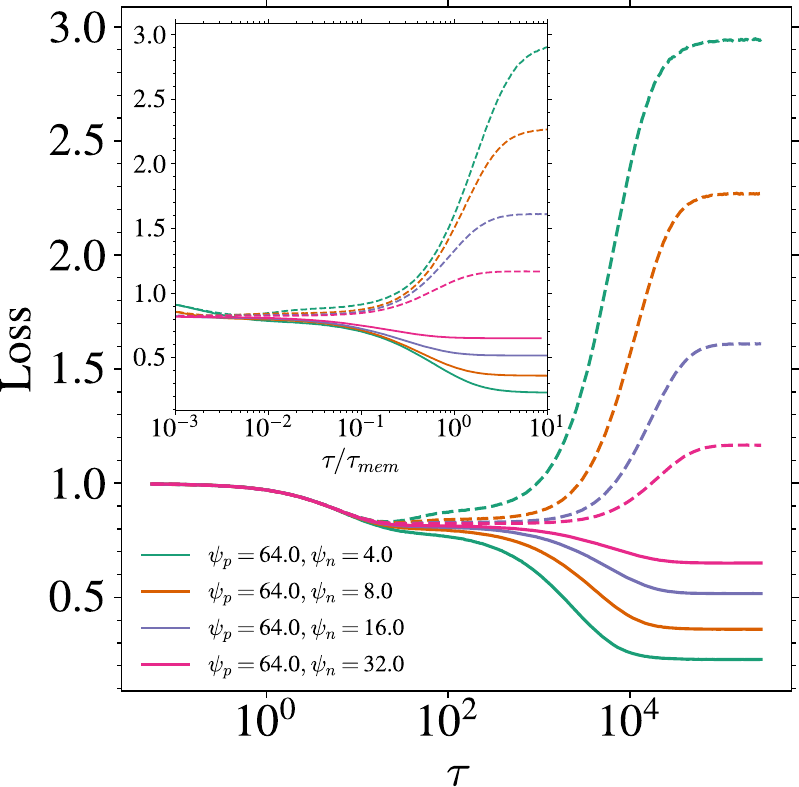} 
       \caption{}
    \end{subfigure}
    \hfill
    \begin{subfigure}[b]{0.25\linewidth}
       \centering
       \includegraphics[height=1.4\linewidth, width=\linewidth]{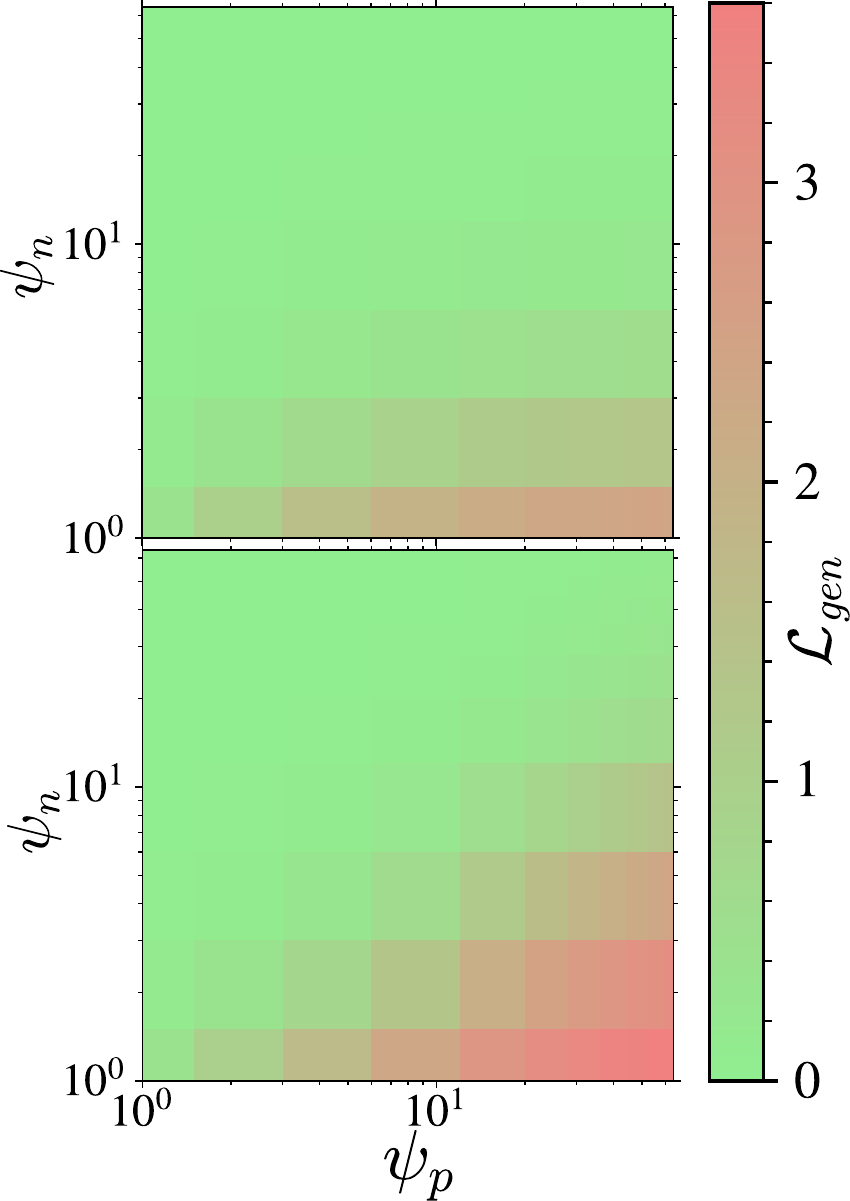} 
       
        \caption{}
     \end{subfigure}    \caption{\textbf{Evolution of the training and test losses for the RFNN.} (A) Distance to the true score $\mathcal{E}_\mathrm{score}$ against training time $\tau$ for $\psi_n=4,8,16,32$,$\psi_p=64, t=0.1$ and $d=100$. In the inset, the training time is rescaled by $\tau_\mathrm{mem}=\psi_p/\Delta_t\lambda_\mathrm{min}$.
     (B) Training (solid) and test (dashed) losses for various $\psi_n$. The inset shows both losses rescaled by $\tau_\mathrm{mem}$. (C) Heatmaps of $\mathcal{L}_\mathrm{gen}$ for $\tau=10^{3}$ (top) and $\tau=10^4$ (bottom) as a function of $\psi_n$ and $\psi_p$.
     All the curves use Pytorch \cite{pytorch_2019} gradient descent. More numerical details can be found in SM Sect.~\ref{appendix:Num_exp_RF}.
    }
    \label{fig:RF_train_test}
\end{figure}

\paragraph*{Emergence of the two timescales during training.}

We first show in Fig.~\ref{fig:RF_train_test} that the behavior of training and test losses in the RF model  mirrors the one found in realistic cases in Sect.~\ref{sect:Numerical_results}, with a separation of timescales $\tau_\mathrm{gen}$ and $\tau_\mathrm{mem}$ which increases with $n$. Equation (\ref{eq:gda}) is linear in $\vA$ and hence it can be solved exactly (see SM). The timescales of the training dynamics are given by the inverse eigenvalues of the $p\times p$ matrix $\Delta_t \vU/\psi_p$.
Building on the Gaussian Equivalence Principle \cite[GEP,][]{Gerace_2020, goldt_2021,mei2020} and the theory of linear pencils \cite{bodin2024random}, \citeauthor{george_2025} (\citeyear{george_2025}) derive a coupled system of equations characterizing the Stieltjes transform of the eigenvalue density $\rho(\lambda)$ of $\vU$ for isotropic  Gaussian data  that lie in a $D$-dimensional subspace with $D\le d$ and $D=\mathcal{O}(d)$. We offer an alternative derivation presented in SM for general variance using the replica method \cite{mezard1987spin} -- a heuristic method from the statistical physics of disordered systems -- yielding the more compact formulation for obtaining the spectrum stated in Theorem~\ref{thm:Saddle_point_equations_new}. Before stating the theorem, we introduce
\begin{align}
    &b_t=\mathbb{E}_{u,v}[v\sigma(e^{-t}\sigma_{\vx}u+\sqrt{\Delta_t}v)], \quad a_t=\mathbb{E}_{u,v}[\sigma(e^{-t}\sigma_{\vx}u+\sqrt{\Delta_t}v)\frac{u}{e^{-t}\sigma_{\vx}}],\\
    &v_t^2=\mathbb{E}_{u,v,w}[\sigma(e^{-t}\sigma_{\vx}u+\sqrt{\Delta_t}v)\sigma(e^{-t}\sigma_{\vx}u+\sqrt{\Delta_t}w)]-a_t^2e^{-2t}\sigma_{\vx}^2,\\
    &s_t^2= \mathbb{E}_u[\sigma(\Gamma_t u)^2]-a_t^2e^{-2t}\sigma_{\vx}^2-v_t^2-b_t^2,
\end{align}
    where $\sigma_{\vx}^2=\frac{\Tr(\vSigma)}{d}$, $\Gamma_t=e^{-2t}\sigma_{\vx}^2+\Delta_t=1+e^{-2t}(\sigma_{\vx}^2-1)$ and the expectation is over the $u,v,w$ random variables which are independent standard Gaussian $\mathcal{N}(0,1)$. 
\begin{thm}
\label{thm:Saddle_point_equations_new}
    Let $q(z)=\frac{1}{p}\Tr(\vU-z\vI_p)^{-1}$, $r(z)=\frac{1}{p}\Tr(\vSigma^{1/2}\vW^T(\vU-z\vI_p)^{-1}\vW\vSigma^{1/2})$ and $s(z)=\frac{1}{p}\Tr(\vW^T(\vU-z\vI_p)^{-1}\vW)$, with $z\in\mathbb{C}$. Let
    \begin{align}
        &\hat{s}(q)=b_t^2\psi_p+\frac{1}{q},\\
    &\hat{r}(r,q)=\frac{\psi_p a_t^2e^{-2t}}{1+\frac{a_t^2e^{-2t}\psi_p 
}{\psi_n }r+\frac{\psi_p v_t^2}{\psi_n }q}.
    \end{align}
    
    Then $q(z), r(z)$ and $s(z)$ satisfy the following set of three equations:
    \begin{align}
    &s=\int\dd \rho_{\vSigma}(\lambda)\frac{1}{\hat{s}(q)+\lambda\hat{r}(r,q)},\\
    &r=\int\dd \rho_{\vSigma}(\lambda)\frac{\lambda}{\hat{s}(q)+\lambda\hat{r}(r,q)},\\
    &\psi_p(s_t^2-z)+\frac{\psi_pv_t^2}{1+\frac{a_t^2e^{-2t}\psi_p} {\psi_n} r+\frac{\psi_p v_t^2}{\psi_n} q}+\frac{1-\psi_p}{q}-\frac{s}{q^2}=0,
\end{align}
The eigenvalue distribution of $\vU$, $\rho(\lambda)$, can then be obtained using the Sokhotski–Plemelj inversion formula $\rho(\lambda)=\underset{\varepsilon \rightarrow0^+}{\lim}\frac{1}{\pi}\operatorname{Im}q(\lambda+i\varepsilon)
$. 
\end{thm}

We now focus on the asymptotic regime $\psi_p,\psi_n \gg 1$, typical for strongly over‑parameterized models trained on large data sets. In this limit, the spectrum of $\vU$ can be described analytically by the following Theorem~\ref{thm:Spectrum_new}.

\begin{thm}[Informal]
\label{thm:Spectrum_new}
Let $\rho$ denote the spectral density of $\vU$.
\begin{enumerate}[leftmargin=*,itemsep=2pt]
    \item[] \textbf{Regime I (overparametrized): $\psi_p>\psi_n\gg 1$.}
    \[
        \rho(\lambda)=\Bigl(1-\frac{1+\psi_n}{\psi_p}\Bigr)\delta(\lambda-{s_t^2})
                     +\frac{\psi_n}{\psi_p}\,\rho_1(\lambda)
                     +\frac{1}{\psi_p}\,\rho_2(\lambda).
    \]
    \item[] \textbf{Regime II (underparametrized): $\psi_n>\psi_p\gg 1$.}
    \[
        \rho(\lambda)=\Bigl(1-\frac{1}{\psi_p}\Bigr)\rho_1(\lambda)
                     +\frac{1}{\psi_p}\,\rho_2(\lambda).
    \]
\end{enumerate}
where $\rho_1$ is an atomless measure with support 
\[
    \left[s_t^2 + v_t^2\left(1-\sqrt{\psi_p/\psi_n}\right)^{2},\;
          s_t^2 + v_t^2\left(1+\sqrt{\psi_p/\psi_n}\right)^{2}\right],
\] 
and $\rho_2$ coincides with the asymptotic eigenvalue bulk density of  the population covariance $\tilde{\vU}=\mathbb{E}_{\vX}[\vU]$; $\rho_2$ is independent of $\psi_n$ and its support is on the scale $\psi_p$.
The eigenvectors associated with $\delta(\lambda-{s_t^2})$ leave both training and test losses unchanged and are therefore irrelevant. In the limit $\psi_p\gg \psi_n$, the supports of $\rho_1$ and $\rho_2$ are respectively on the scales $\psi_p/\psi_n$ and $\psi_p$, i.e. they are well separated.
\end{thm}

The proofs of both theorems are shown in SM (Sect.~\ref{appendix:proofs}). We recall that training timescales are directly related to eigenvalues $\lambda$ via the relation \(\tau^{-1}= \psi_p / \Delta_t \lambda_\mathrm{\min} \). Theorem~\ref{thm:Spectrum_new} therefore demonstrates the emergence of the two training timescales \( \tau_{\mathrm{mem}} \) and \( \tau_{\mathrm{gen}} \) in the overparametrized regime of the RFNN model. They are respectively associated to the measures $\rho_1$ and $\rho_2$, which are well separated in regime I, for  $\psi_p\gg\psi_n\gg1$, as shown in 
Fig.~\ref{fig:RFNN_plus_spectrum}. \\
{\bf Generalization}: The timescale \( \tau_{\mathrm{gen}} \) on which the first relaxation takes place is associated to the formation of the generalization regime. It is related to the bulk \( \rho_2 \) and is or order $1/\Delta_t$. This regime only depends on the population covariance $\vSigma$ of the data and is independent of the specific realization of the dataset. On this timescale, which is of order one, both the training \( \mathcal{L}_{\mathrm{train}} \) and test \( \mathcal{L}_{\mathrm{test}} \) losses decrease. The generalization loss $\mathcal{L}_\mathrm{gen} = \mathcal{L}_{\mathrm{test}}-\mathcal{L}_{\mathrm{train}}$ is zero, and $\mathcal{E}_\mathrm{score}$ tends to a value that we find to scale as $\mathcal{O}(\psi_n^{-\eta})$ with $\eta\simeq0.59$ numerically (see Fig. \ref{fig:RF_train_test}).  \\
{\bf Memorization:} The timescale \( \tau_{\mathrm{mem}} \), 
   on which the second stage of the dynamics takes place, is associated to overfitting and memorization. It is related to the bulk \( \rho_1 \), and scales as \( \psi_p / \Delta_t \lambda_\mathrm{\min} \), where \( \lambda_\mathrm{\min} \) is the left edge of \( \rho_1 \). In the overparameterized regime \( p \gg n \), \( \tau_{\mathrm{mem}} \) becomes large and of order $\psi_n/\Delta_t$, thus implying a scaling of $\tau_{\mathrm{mem}}$ with $n$. On this timescale, the training loss decreases while the test loss increases, converging to their respective asymptotic values as computed in~\cite{george_2025}. Fig.~\ref{fig:RF_train_test} indeed shows that all training and test curves separate, correspondingly the generalization loss \( \mathcal{L}_{\mathrm{gen}}\) increases, at a time that scales with \( \psi_p /\Delta_t  \lambda_{\min} \), as shown in the inset.\\
   As $n$ increases, the asymptotic ($\tau \rightarrow \infty$) generalization loss \( \mathcal{L}_{\mathrm{gen}}\) decreases, indicating a reduced overfitting. 
For $n>n^*(p) = p$, although some overfitting remains (i.e., \( \mathcal{L}_{\mathrm{gen}} > 0 \)), the value of \( \mathcal{L}_{\mathrm{gen}} \) is sensibly reduced, and the model is no longer expressive enough to memorize the training data, as shown in \cite{george_2025}. This regime corresponds to the \emph{Architectural Regularization} phase in Fig.~\ref{fig:summary_results}. We show in  Fig.~\ref{fig:RF_train_test} (panel C) how the generalization loss \( \mathcal{L}_{\mathrm{gen}} \) varies in the $(n,p)$ plane depending on the time $\tau$ at which training is stopped. In agreement with the above results, we find that the generalization--memorization transition line depends on $\tau$ and moves upward for larger values of $\tau$, similarly to the numerical results exposed in Fig. \ref{fig:numerical_memorization_p} and the illustration in Fig. \ref{fig:summary_results}.

\section{Conclusions}
\label{sect:Conclusion}
We have shown that the training dynamics of neural network-based score functions display a form of implicit regularization that prevents memorization even in highly overparameterized diffusion models. Specifically, we have identified two well-separated timescales in the learning: $\tau_\mathrm{gen}$, at which models begins to generate high-quality, novel samples, and $\tau_\mathrm{mem}$, beyond which they start to memorize the training data. The gap between these timescales grows with the size of the training set, leading to a broad window where early stopped models generate novel samples of high-quality. We have demonstrated that this phenomenon happens in realistic settings, for controlled synthetic data, and in analytically tractable models. Although our analysis focuses on DMs, the underlying score‑learning mechanism we uncover is common to all score‑based generative models such as stochastic interpolants \cite{albergo2023stochastic} or flow matching \cite{lipman_2023}; we therefore expect our results to generalize to this broader class.

\paragraph*{Limitations and future works.}
\begin{itemize}[itemsep=.5em]
\item While we derived our results under SGD optimization, most DMs are trained in practice with Adam \cite{kingma2015Adam}. In SM Sects.~\ref{appendix:numerical_details} and~\ref{appendix:Num_exp_RF}, we show that the two key timescales still arise using Adam, although with much fewer optimization steps. Studying how different optimizers shift these timescales would be valuable for practical usage.
\item All experiments in Sect.~\ref{sect:Numerical_results} are conducted with unconditional DMs. We additionally verify in SM Sect.~B, using a toy Gaussian mixture dataset and classifier-free guidance \cite{ho2022classifierfree}, that the same scaling of $\tau_\mathrm{mem}$ with $n$ holds in the conditional settings. Understanding precisely how the absolute timescales $\tau_\mathrm{mem}$ and $\tau_\mathrm{gen}$ depend on the conditioning remains an open question.
\item Our numerical experiments cover a range of $p$ between 1M and 16M. Exploring a wider range is essential to map the full $(n,p)$ phase diagram sketched in Fig.~\ref{fig:summary_results} and understand the precise effect of expressivity on dynamical regularization.
\item Finally, our theoretical analysis rely on well-controlled data and score models that reproduce the core effects. Extending these analytical frameworks to richer data distributions (such as Gaussian mixtures or data from the hidden manifold model) and to structured architectures would be valuable to further characterize the implicit dynamical regularization of training score-functions. In particular investigating how heavy-tailed data distribution \cite{Adomaityte_2024} affect the picture described here could be valuable.
\item Although DMs trained on large and diverse datasets likely avoid the memorization regime we study here, some industrial models were shown to exhibit partial memorization \cite{Carlini_2023, somepalli_2022}. Our results provide practical guidelines (early-stopping, control the network capacity) to train DMs robustly and hence avoid memorization, which can be especially helpful in data-scarce domains (e.g., physical sciences).
\end{itemize}

\begin{ack}
The authors thank Valentin De Bortoli for initial motivating  discussions on memorization--generalization transitions. 
RU thanks Beatrice Achilli, Jérome Garnier-Brun, Carlo Lucibello and Enrico Ventura for insightful discussions. RU is grateful to Bocconi University for its hospitality during his stay, during which part of this work was conducted. This work was performed using HPC resources from GENCI-IDRIS (Grant 2025-AD011016319). GB acknowledges support from the French government under the management of the Agence Nationale PR[AI]RIE-PSAI (ANR-23-IACL-0008). MM acknowledges the support of the PNRR-PE-AI FAIR project funded by the NextGeneration EU program. After completing this work, we became aware that A. Favero, A. Sclocchi, and M. Wyart \cite{Favero2025_bigger} had also been investigating the memorization--generalization transition from a similar perspective.
\end{ack}

\bibliographystyle{apalike}
\bibliography{Bibliography.bib}

\begin{thebibliography}{}

\bibitem[Achilli et~al., 2024]{achilli2024}
Achilli, B., Ventura, E., Silvestri, G., Pham, B., Raya, G., Krotov, D., Lucibello, C., and Ambrogioni, L. (2024).
\newblock Losing dimensions: Geometric memorization in generative diffusion.

\bibitem[Adomaityte et~al., 2024]{Adomaityte_2024}
Adomaityte, U., Defilippis, L., Loureiro, B., and Sicuro, G. (2024).
\newblock High-dimensional robust regression under heavy-tailed data: asymptotics and universality.
\newblock {\em Journal of Statistical Mechanics: Theory and Experiment}, 2024(11):114002.

\bibitem[Albergo et~al., 2023]{albergo2023stochastic}
Albergo, M.~S., Boffi, N.~M., and Vanden-Eijnden, E. (2023).
\newblock Stochastic interpolants: A unifying framework for flows and diffusions.

\bibitem[Bach, 2023]{Bach2023Hermite}
Bach, F. (2023).
\newblock Polynomial magic iii: Hermite polynomials.
\newblock \url{https://francisbach.com/hermite-polynomials/}.
\newblock Accessed: 2025-10-09.

\bibitem[Bai and Zhou, 2008]{bai2008}
Bai, Z. and Zhou, W. (2008).
\newblock Large sample covariance matrices without independence structures in columns.
\newblock {\em Statistica Sinica}, 18(2):425--442.

\bibitem[Biroli et~al., 2024]{Biroli_2024}
Biroli, G., Bonnaire, T., de~Bortoli, V., and M{\'e}zard, M. (2024).
\newblock Dynamical regimes of diffusion models.
\newblock {\em Nature Communications}, 15(9957).
\newblock Open access.

\bibitem[Bodin, 2024]{bodin2024random}
Bodin, A. P.~M. (2024).
\newblock {\em Random Matrix Methods for High-Dimensional Machine Learning Models}.
\newblock Phd thesis, École Polytechnique Fédérale de Lausanne (EPFL), Lausanne, Switzerland.

\bibitem[Bortoli, 2022]{bortoli2022convergence}
Bortoli, V.~D. (2022).
\newblock Convergence of denoising diffusion models under the manifold hypothesis.
\newblock {\em Transactions on Machine Learning Research}.
\newblock Expert Certification.

\bibitem[Carlini et~al., 2023]{Carlini_2023}
Carlini, N., Hayes, J., Nasr, M., Jagielski, M., Sehwag, V., Tram\`{e}r, F., Balle, B., Ippolito, D., and Wallace, E. (2023).
\newblock Extracting training data from diffusion models.
\newblock In {\em Proceedings of the 32nd USENIX Conference on Security Symposium}, SEC '23, USA. USENIX Association.

\bibitem[Chen et~al., 2024]{chen2024memorizationfree}
Chen, C., Liu, D., and Xu, C. (2024).
\newblock Towards memorization-free diffusion models.

\bibitem[Cui et~al., 2024]{cui_2024}
Cui, H., Krzakala, F., Vanden-Eijnden, E., and Zdeborova, L. (2024).
\newblock Analysis of learning a flow-based generative model from limited sample complexity.
\newblock In {\em The Twelfth International Conference on Learning Representations}.

\bibitem[Cui et~al., 2025]{cui_2025}
Cui, H., Pehlevan, C., and Lu, Y.~M. (2025).
\newblock A precise asymptotic analysis of learning diffusion models: theory and insights.

\bibitem[D'Ascoli et~al., 2020]{Ascoli_2020}
D'Ascoli, S., Refinetti, M., Biroli, G., and Krzakala, F. (2020).
\newblock Double trouble in double descent: Bias and variance(s) in the lazy regime.
\newblock In III, H.~D. and Singh, A., editors, {\em Proceedings of the 37th International Conference on Machine Learning}, volume 119 of {\em Proceedings of Machine Learning Research}, pages 2280--2290. PMLR.

\bibitem[Favero et~al., 2025]{Favero2025_bigger}
Favero, A., Sclocchi, A., and Wyart, M. (2025).
\newblock Bigger isn't always memorizing: Early stopping overparameterized diffusion models.

\bibitem[George et~al., 2025]{george_2025}
George, A.~J., Veiga, R., and Macris, N. (2025).
\newblock Denoising score matching with random features: Insights on diffusion models from precise learning curves.

\bibitem[Gerace et~al., 2020]{Gerace_2020}
Gerace, F., Loureiro, B., Krzakala, F., Mezard, M., and Zdeborova, L. (2020).
\newblock Generalisation error in learning with random features and the hidden manifold model.
\newblock In III, H.~D. and Singh, A., editors, {\em Proceedings of the 37th International Conference on Machine Learning}, volume 119 of {\em Proceedings of Machine Learning Research}, pages 3452--3462. PMLR.

\bibitem[Goldt et~al., 2021]{goldt_2021}
Goldt, S., Loureiro, B., Reeves, G., Krzakala, F., Mézard, M., and Zdeborová, L. (2021).
\newblock The gaussian equivalence of generative models for learning with shallow neural networks.

\bibitem[Gu et~al., 2023]{gu2023memorization}
Gu, X., Du, C., Pang, T., Li, C., Lin, M., and Wang, Y. (2023).
\newblock On memorization in diffusion models.

\bibitem[Heusel et~al., 2017]{heusel2017gans}
Heusel, M., Ramsauer, H., Unterthiner, T., Nessler, B., and Hochreiter, S. (2017).
\newblock Gans trained by a two time-scale update rule converge to a local nash equilibrium.

\bibitem[Ho et~al., 2020]{ho2020}
Ho, J., Jain, A., and Abbeel, P. (2020).
\newblock Denoising diffusion probabilistic models.

\bibitem[Ho and Salimans, 2022]{ho2022classifierfree}
Ho, J. and Salimans, T. (2022).
\newblock Classifier-free diffusion guidance.

\bibitem[Hu and Lu, 2023]{hu2023}
Hu, H. and Lu, Y.~M. (2023).
\newblock Universality laws for high-dimensional learning with random features.
\newblock {\em IEEE Transactions on Information Theory}, 69(3):1932--1964.

\bibitem[Hyv{{\"a}}rinen, 2005]{hyvarinen_05}
Hyv{{\"a}}rinen, A. (2005).
\newblock Estimation of non-normalized statistical models by score matching.
\newblock {\em Journal of Machine Learning Research}, 6(24):695--709.

\bibitem[Kadkhodaie et~al., 2024]{kadkhodaie_2024}
Kadkhodaie, Z., Guth, F., Simoncelli, E.~P., and Mallat, S. (2024).
\newblock Generalization in diffusion models arises from geometry-adaptive harmonic representations.
\newblock In {\em The Twelfth International Conference on Learning Representations}.

\bibitem[Kamb and Ganguli, 2024]{Kamb2024}
Kamb, M. and Ganguli, S. (2024).
\newblock An analytic theory of creativity in convolutional diffusion models.

\bibitem[Karras et~al., 2022]{karras2022elucidatingdesignspacediffusionbased}
Karras, T., Aittala, M., Aila, T., and Laine, S. (2022).
\newblock Elucidating the design space of diffusion-based generative models.

\bibitem[Kibble, 1945]{kibble1945}
Kibble, W.~F. (1945).
\newblock An extension of a theorem of mehler’s on hermite polynomials.
\newblock {\em Mathematical Proceedings of the Cambridge Philosophical Society}, 41(1):12--15.

\bibitem[Kingma and Ba, 2015]{kingma2015Adam}
Kingma, D.~P. and Ba, J. (2015).
\newblock Adam: A method for stochastic optimization.
\newblock In Bengio, Y. and LeCun, Y., editors, {\em ICLR (Poster)}.

\bibitem[Li et~al., 2025]{li2025generalizationpropertiesdiffusionmodels}
Li, P., Li, Z., Zhang, H., and Bian, J. (2025).
\newblock On the generalization properties of diffusion models.

\bibitem[Li et~al., 2024a]{li_2024_good_score}
Li, S., Chen, S., and Li, Q. (2024a).
\newblock A good score does not lead to a good generative model.

\bibitem[Li et~al., 2024b]{Biferale_2024}
Li, T., Biferale, L., Bonaccorso, F., and et~al. (2024b).
\newblock Synthetic lagrangian turbulence by generative diffusion models.
\newblock {\em Nat Mach Intell}, 6:393--403.

\bibitem[Lipman et~al., 2023]{lipman_2023}
Lipman, Y., Chen, R. T.~Q., Ben-Hamu, H., Nickel, M., and Le, M. (2023).
\newblock Flow matching for generative modeling.
\newblock In {\em The Eleventh International Conference on Learning Representations}.

\bibitem[Liu et~al., 2024]{sora2024}
Liu, Y., Zhang, K., Li, Y., Yan, Z., Gao, C., Chen, R., Yuan, Z., Huang, Y., Sun, H., Gao, J., He, L., and Sun, L. (2024).
\newblock Sora: A review on background, technology, limitations, and opportunities of large vision models.

\bibitem[Liu et~al., 2015]{CelebA}
Liu, Z., Luo, P., Wang, X., and Tang, X. (2015).
\newblock Deep learning face attributes in the wild.
\newblock In {\em Proceedings of International Conference on Computer Vision (ICCV)}.

\bibitem[Mei et~al., 2019]{Mei_2019}
Mei, S., Misiakiewicz, T., and Montanari, A. (2019).
\newblock Mean-field theory of two-layers neural networks: dimension-free bounds and kernel limit.
\newblock In Beygelzimer, A. and Hsu, D., editors, {\em Proceedings of the Thirty-Second Conference on Learning Theory}, volume~99 of {\em Proceedings of Machine Learning Research}, pages 2388--2464. PMLR.

\bibitem[Mei and Montanari, 2020]{mei2020}
Mei, S. and Montanari, A. (2020).
\newblock The generalization error of random features regression: Precise asymptotics and double descent curve.

\bibitem[M{\'e}zard et~al., 1987]{mezard1987spin}
M{\'e}zard, M., Parisi, G., and Virasoro, M.~A. (1987).
\newblock {\em Spin Glass Theory and Beyond: An Introduction to the Replica Method and Its Applications}, volume~9 of {\em Lecture Notes in Physics}.
\newblock World Scientific Publishing Company, Singapore.

\bibitem[Paszke et~al., 2019]{pytorch_2019}
Paszke, A., Gross, S., Massa, F., Lerer, A., Bradbury, J., Chanan, G., Killeen, T., Lin, Z., Gimelshein, N., Antiga, L., Desmaison, A., Kopf, A., Yang, E., DeVito, Z., Raison, M., Tejani, A., Chilamkurthy, S., Steiner, B., Fang, L., Bai, J., and Chintala, S. (2019).
\newblock Pytorch: An imperative style, high-performance deep learning library.
\newblock In {\em Advances in Neural Information Processing Systems}, volume~32, pages 8024--8035. Curran Associates, Inc.

\bibitem[Potters and Bouchaud, 2020]{Potters_Bouchaud_2020}
Potters, M. and Bouchaud, J.-P. (2020).
\newblock {\em A First Course in Random Matrix Theory: for Physicists, Engineers and Data Scientists}.
\newblock Cambridge University Press.

\bibitem[Price et~al., 2025]{Price_2025}
Price, I., Sanchez-Gonzalez, A., Alet, F., and et~al. (2025).
\newblock Probabilistic weather forecasting with machine learning.
\newblock {\em Nature}, 637:84--90.

\bibitem[Péché, 2019]{Peche2019}
Péché, S. (2019).
\newblock A note on the pennington-worah distribution.
\newblock {\em Electronic Communications in Probability}, 24:1--7.

\bibitem[Rahaman et~al., 2019]{rahaman2019spectral}
Rahaman, N., Baratin, A., Arpit, D., Draxler, F., Lin, M., Hamprecht, F., Bengio, Y., and Courville, A. (2019).
\newblock On the spectral bias of neural networks.
\newblock In {\em International conference on machine learning}, pages 5301--5310. PMLR.

\bibitem[Rahimi and Recht, 2007]{Rahimi_2007}
Rahimi, A. and Recht, B. (2007).
\newblock Random features for large-scale kernel machines.
\newblock In Platt, J., Koller, D., Singer, Y., and Roweis, S., editors, {\em Advances in Neural Information Processing Systems}, volume~20. Curran Associates, Inc.

\bibitem[Robbins and Monro, 1951]{robbins1951stochastic}
Robbins, H. and Monro, S. (1951).
\newblock A stochastic approximation method.
\newblock {\em The annals of mathematical statistics}, pages 400--407.

\bibitem[Rombach et~al., 2021]{DALL-E}
Rombach, R., Blattmann, A., Lorenz, D., Esser, P., and Ommer, B. (2021).
\newblock High-resolution image synthesis with latent diffusion models.

\bibitem[Ronneberger et~al., 2015]{Ronneberger2015}
Ronneberger, O., Fischer, P., and Brox, T. (2015).
\newblock U-net: Convolutional networks for biomedical image segmentation.
\newblock In Navab, N., Hornegger, J., Wells, W.~M., and Frangi, A.~F., editors, {\em Medical Image Computing and Computer-Assisted Intervention -- MICCAI 2015}, pages 234--241, Cham. Springer International Publishing.

\bibitem[Shah et~al., 2025]{Shah2025}
Shah, K., Kalavasis, A., Klivans, A.~R., and Daras, G. (2025).
\newblock Does generation require memorization? creative diffusion models using ambient diffusion.

\bibitem[Silverstein and Bai, 1995]{SILVERSTEIN1995}
Silverstein, J. and Bai, Z. (1995).
\newblock On the empirical distribution of eigenvalues of a class of large dimensional random matrices.
\newblock {\em Journal of Multivariate Analysis}, 54(2):175--192.

\bibitem[Sohl-Dickstein et~al., 2015]{sohl-dickstein_15}
Sohl-Dickstein, J., Weiss, E., Maheswaranathan, N., and Ganguli, S. (2015).
\newblock Deep unsupervised learning using nonequilibrium thermodynamics.
\newblock In Bach, F. and Blei, D., editors, {\em Proceedings of the 32nd International Conference on Machine Learning}, volume~37 of {\em Proceedings of Machine Learning Research}, pages 2256--2265, Lille, France. PMLR.

\bibitem[Somepalli et~al., 2023a]{somepalli_2022}
Somepalli, G., Singla, V., Goldblum, M., Geiping, J., and Goldstein, T. (2023a).
\newblock Diffusion art or digital forgery? investigating data replication in diffusion models.
\newblock In {\em Proceedings of the IEEE/CVF Conference on Computer Vision and Pattern Recognition}.

\bibitem[Somepalli et~al., 2023b]{somepalli_2023}
Somepalli, G., Singla, V., Goldblum, M., Geiping, J., and Goldstein, T. (2023b).
\newblock Understanding and mitigating copying in diffusion models.
\newblock {\em Advances in Neural Information Processing Systems}, 36:47783--47803.

\bibitem[Song et~al., 2022]{song2022DDIM}
Song, J., Meng, C., and Ermon, S. (2022).
\newblock Denoising diffusion implicit models.

\bibitem[Song et~al., 2021a]{song2021a}
Song, Y., Durkan, C., Murray, I., and Ermon, S. (2021a).
\newblock Maximum likelihood training of score-based diffusion models.

\bibitem[Song and Ermon, 2019]{song2019}
Song, Y. and Ermon, S. (2019).
\newblock Generative modeling by estimating gradients of the data distribution.
\newblock In Wallach, H., Larochelle, H., Beygelzimer, A., d\textquotesingle Alch\'{e}-Buc, F., Fox, E., and Garnett, R., editors, {\em Advances in Neural Information Processing Systems}, volume~32. Curran Associates, Inc.

\bibitem[Song et~al., 2021b]{song2021b}
Song, Y., Sohl-Dickstein, J., Kingma, D.~P., Kumar, A., Ermon, S., and Poole, B. (2021b).
\newblock Score-based generative modeling through stochastic differential equations.

\bibitem[Vaswani et~al., 2017]{Vaswani2017}
Vaswani, A., Shazeer, N., Parmar, N., Uszkoreit, J., Jones, L., Gomez, A.~N., Kaiser, L.~u., and Polosukhin, I. (2017).
\newblock Attention is all you need.
\newblock In Guyon, I., Luxburg, U.~V., Bengio, S., Wallach, H., Fergus, R., Vishwanathan, S., and Garnett, R., editors, {\em Advances in Neural Information Processing Systems}, volume~30. Curran Associates, Inc.

\bibitem[Ventura et~al., 2025]{ventura2025}
Ventura, E., Achilli, B., Silvestri, G., Lucibello, C., and Ambrogioni, L. (2025).
\newblock Manifolds, random matrices and spectral gaps: The geometric phases of generative diffusion.

\bibitem[Vincent, 2011]{Vincent_2011}
Vincent, P. (2011).
\newblock A connection between score matching and denoising autoencoders.
\newblock {\em Neural Computation}, 23(7):1661--1674.

\bibitem[Wang, 2025]{Wang2025}
Wang, B. (2025).
\newblock An analytical theory of power law spectral bias in the learning dynamics of diffusion models.

\bibitem[Wang et~al., 2024]{wang2024}
Wang, P., Zhang, H., Zhang, Z., Chen, S., Ma, Y., and Qu, Q. (2024).
\newblock Diffusion models learn low-dimensional distributions via subspace clustering.

\bibitem[Wen et~al., 2024]{wen2024detecting}
Wen, Y., Liu, Y., Chen, C., and Lyu, L. (2024).
\newblock Detecting, explaining, and mitigating memorization in diffusion models.

\bibitem[Wu et~al., 2025]{Wu2025}
Wu, Y.-H., Marion, P., Biau, G., and Boyer, C. (2025).
\newblock Taking a big step: Large learning rates in denoising score matching prevent memorization.

\bibitem[Yoon et~al., 2023]{yoon2023diffusion}
Yoon, T., Choi, J.~Y., Kwon, S., and Ryu, E.~K. (2023).
\newblock Diffusion probabilistic models generalize when they fail to memorize.
\newblock In {\em ICML 2023 Workshop on Structured Probabilistic Inference {\&} Generative Modeling}.

\bibitem[Zhang et~al., 2023]{survey_audio}
Zhang, C., Zhang, C., Zheng, S., Zhang, M., Qamar, M., Bae, S.-H., and Kweon, I.~S. (2023).
\newblock A survey on audio diffusion models: Text to speech synthesis and enhancement in generative ai.

\bibitem[Zhi-Qin John~Xu et~al., 2020]{Zhi_Qin_John_Xu_2020}
Zhi-Qin John~Xu, Z.-Q. J.~X., Yaoyu~Zhang, Y.~Z., Tao~Luo, T.~L., Yanyang~Xiao, Y.~X., and Zheng~Ma, Z.~M. (2020).
\newblock Frequency principle: Fourier analysis sheds light on deep neural networks.
\newblock {\em Communications in Computational Physics}, 28(5):1746–1767.

\end{thebibliography}

\newpage
\appendix
\onecolumn
\begin{center}
    {\Large Why Diffusion Models Don’t Memorize:\\ The Role of Implicit Dynamical Regularization in Training} \\ \vspace{1ex} {\large \bf Supplementary Material (SM)}
    \\ \vspace{3ex} {\large Tony Bonnaire$^\dagger$, Raphaël Urfin$^\dagger$, Giulio Biroli, Marc Mézard}
\end{center}

This document provides detailed derivations and additional experiments supporting the main text (MT). In Sect.~\ref{appendix:numerical_details}, we give details about the numerical experiments carried out in Sect.~\ref{sect:Numerical_results}. In Sect.~\ref{appendix:numerical_GMM} we provide additional numerical experiments on simplified score and data models. Sect.~\ref{appendix:proofs} gives formal proofs of the main theorems of Sect.~\ref{sect:Analytical_results}. Finally, Sect.~\ref{appendix:Num_exp_RF} exposes more details on the numerical experiments of Sect.~\ref{sect:Analytical_results}. 

\section{Numerical experiments on CelebA} \label{appendix:numerical_details}

\subsection{Details on the numerical setup}

\begin{figure}
    \centering
    \includegraphics[width=.65\linewidth]{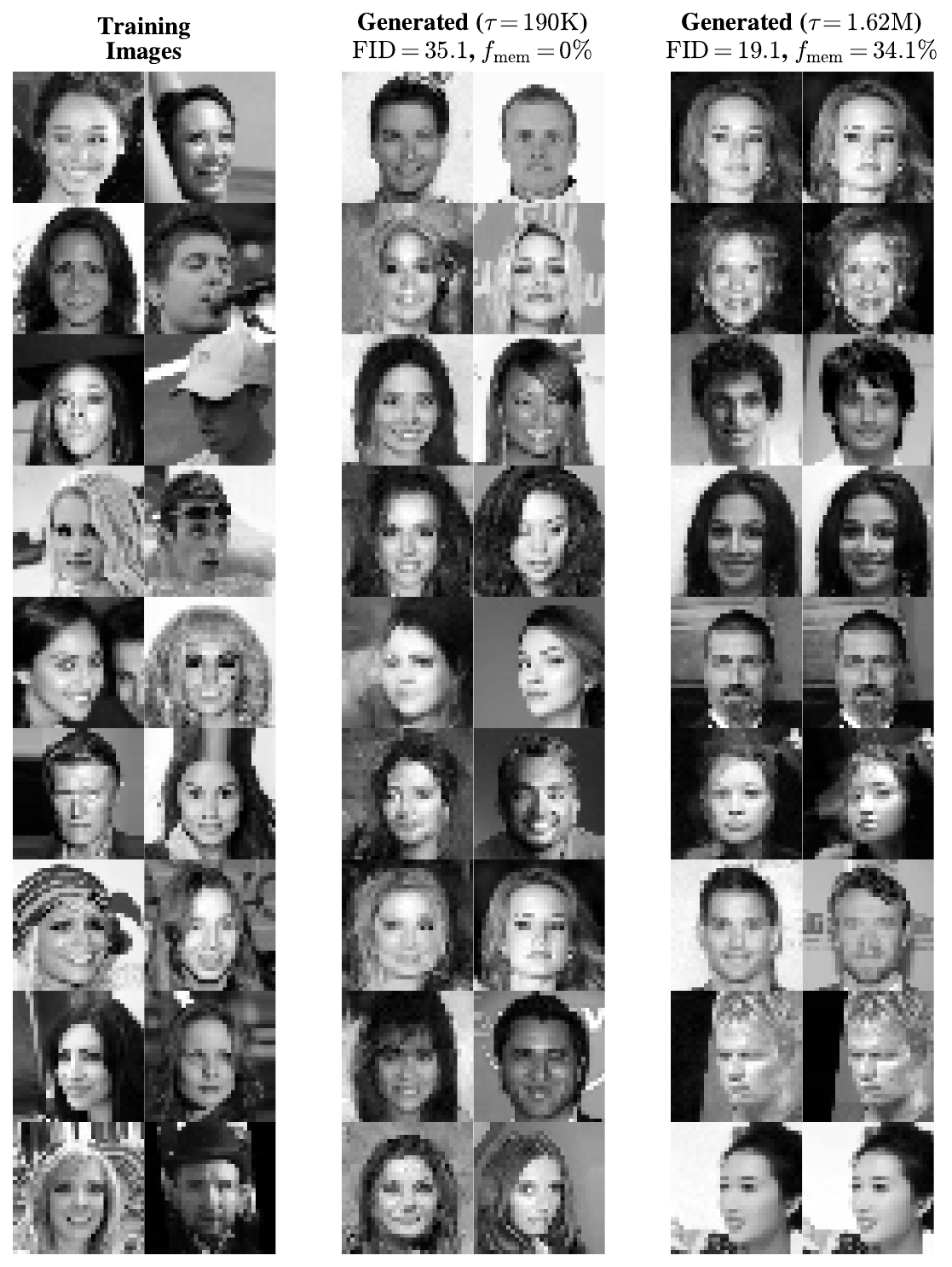}
    \caption{\textbf{Training and generation on CelebA.} The left-most block shows random training images. Middle and right blocks show generated samples in the left column (after $\tau=190$K and $\tau=1.62$M SGD updates respectively), alongside each sample's nearest neighbor in the training set in the right column. All generated images come from model trained on $n=1024$ with base width $W=32$.
    }
    \label{fig:images_CelebA}
\end{figure}

\paragraph*{Dataset.} All numerical experiments in Sect.~\ref{sect:Numerical_results} of the MT use the CelebA face dataset \cite{CelebA}. We center-crop each RGB image to $32\times 32$ pixels and convert to grayscale images in order to accelerate the training of our Diffusion Models (DMs). To precisely control the samples seen by a model, no data augmentation is applied, and we vary the training set size $n$ in the window $\left[128, 32768\right]$. Examples of training samples are shown in the left-most block of Fig.~\ref{fig:images_CelebA}.

\paragraph*{Architecture.} As commonly done in DDPMs implementations \cite[e.g.,][]{ho2020, song2021b}, the network approximating the score function is a U-Net \cite{Ronneberger2015} made of three resolution levels, each containing two residual blocks with channel multipliers $\{1, 2, 4\}$ respectively. We apply attention to the two coarsest resolutions, and embed the diffusion time via sinusoidal position embedding \cite{Vaswani2017}. The base channel width $W$ varies from $16$ to $64$ depending on the experiment, resulting in a total of $1$ to $16$ million trainable parameters. 

\paragraph*{Time reparameterization.} Compared to the framework presented in the MT, the DDPMs we train make use of a time reparameterization of the forward and backward processes with a variance schedule $\{\beta_{t'}\}_{t'=1}^T$, where $T$ is the time horizon given as a number of steps, fixed to 1000 in our experiments. The variance is evolving linearly from $\beta_1 = 10^{-4}$ to $\beta_{1000} = 2 \times 10^{-2}$. A sample at time $t'$, denoted $\vx(t')$, can be expressed from $\vx(0)$ as the following interpolation
\begin{equation}
    \vx(t') = \sqrt{\overline{\alpha}(t')} \vx(0) + \sqrt{1 - \overline{\alpha}(t')} \vxi,
\end{equation}
where $\overline{\alpha}(t') = \prod_{s=1}^{t'} (1-\beta_s)$, and $\vxi$ is a standard and centered Gaussian noise. This is a reparameterization of the Ornstein-Uhlenbeck process from Eq.~\ref{eq:forward} defined through time $t$ in the MT, with
\begin{equation}
    t = -\frac{1}{2} \log \left( \overline{\alpha}(t')\right).
\end{equation}

\paragraph*{Training.} All DMs are trained with Stochastic Gradient Descent (SGD) at fixed learning rate $\eta = 0.01$, fixed momentum $\beta=0.95$ and batch size $B=\min(n, 512)$. We focus on SGD to facilitate the analysis of time scaling, avoiding problems that may cause alternative adaptive optimization schemes like Adam \cite{kingma2015Adam}. We train each model for at least 2M SGD steps, sometimes more for large values of $n$ displaying memorization only later. We do not employ exponential moving average or learning-rate warm-up.

\paragraph*{Generation.} To accelerate sampling while preserving FID, we employ the DDIM sampler of \citeauthor{song2022DDIM} (\citeyear{song2022DDIM}) which replaces the Markovian reverse SDE with a deterministic, non-Markovian update. Given a trained denoiser $\vxi_{\vtheta}(\vx_t, t)$, we iterate for $t=T', \ldots,1$
\begin{equation}
  \vx_{t-1} = \sqrt{\overline{\alpha}(t-1)}\ \frac{\vx_t - \sqrt{1-\overline{\alpha}(t)}\;\vxi_{\vtheta}(\vx_t,t)}{\sqrt{\overline{\alpha}(t)}} + \sqrt{1-\overline{\alpha}(t-1)}\vxi_{\vtheta}(\vx_t, t),
\end{equation}
with $T'=200$. During training, we generate at 40 milestones a set of 10,000 samples to assess generalization and memorization. Examples of samples obtained from a model trained on $n=1024$ samples with base width $W=32$ are shown in the middle and right blocks from Fig.~\ref{fig:images_CelebA} for two training times, $\tau=190$K and $\tau=1.62$M. At $\tau=190$K the model generalizes ($f_\mathrm{mem}=0\%$) and achieve a test FID of 35.1. After too much training, memorization sets in and, by $\tau=1.62$M steps, nearly half the generated samples reproduce training images ($f_\mathrm{mem}=47.2\%$).

\paragraph*{Statistical evaluation.} FIDs \cite{heusel2017gans} are computed\footnote{Using the \href{https://github.com/mseitzer/pytorch-fid}{pytorch-fid} Python package.} using 10,000 generated samples and 10,000 test samples, averaged over 5 independent runs with disjoint test sets. Error bars in the MT denote twice the standard deviation. Training and test losses are estimated similarly over 5 repeated evaluation on $n$ training samples and 2048 test samples, and give negligible confidence intervals. For the memorization fraction $f_\mathrm{mem}(\tau)$, we report the standard error on the mean obtained via bootstrap resampling of the 10,000 generated samples. We also verified that the scaling in the memorization time $\tau_\mathrm{mem}$ is insensitive to the choice of the threshold $k$ used to define $f_\mathrm{mem}$ in Eq.~\ref{eq:memorization_criterion} by testing larger and lower values.

\paragraph*{Computing resources.} Most trainings were performed on Nvidia H100 GPUs (80GB of memory). A typical run of 2M steps takes approximately 50 hours on two GPUs and vary with the model size (defined through its base width $W$). In total, we train 18 distinct models for the several $n,W$ configurations of the MT. The longest training ($n=32768$ and $W=32$ in Fig.~\ref{fig:numerical_memorization_n}) ran for 11M steps. The generation of $10,000$ samples over 40 training times takes around an additional hour per model on the same hardware support.

\begin{figure}
    \centering
    \includegraphics[width=.49\linewidth]{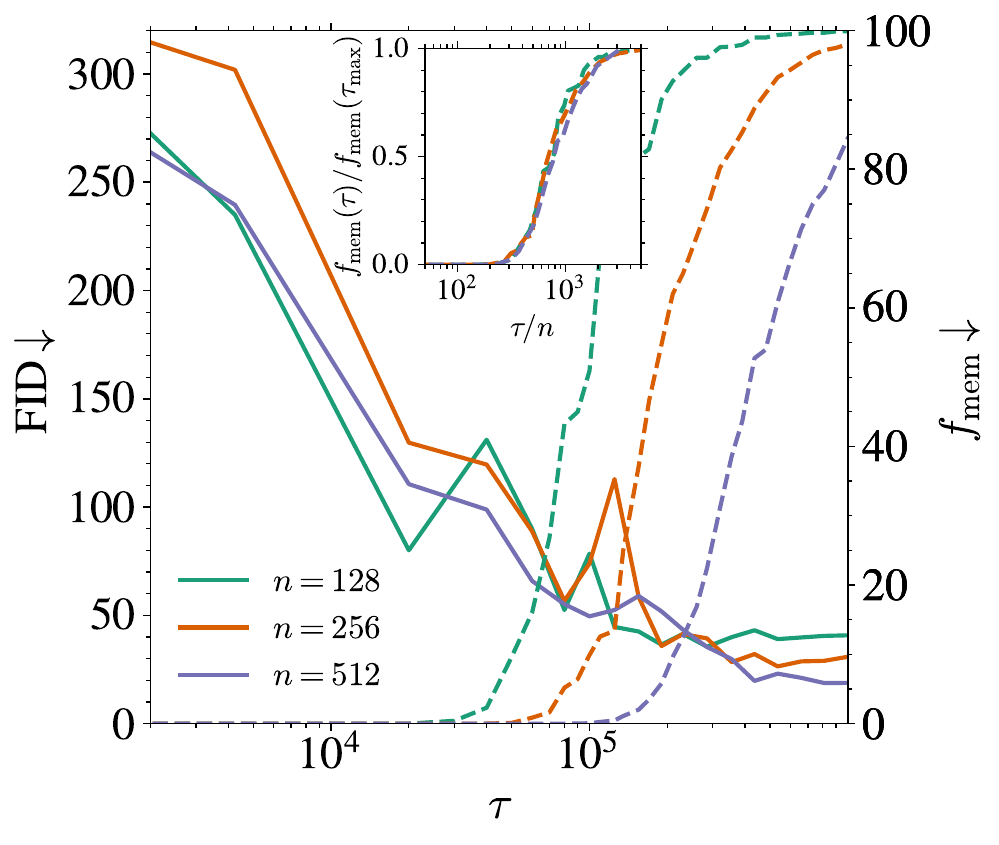}
    \includegraphics[width=.49\linewidth]{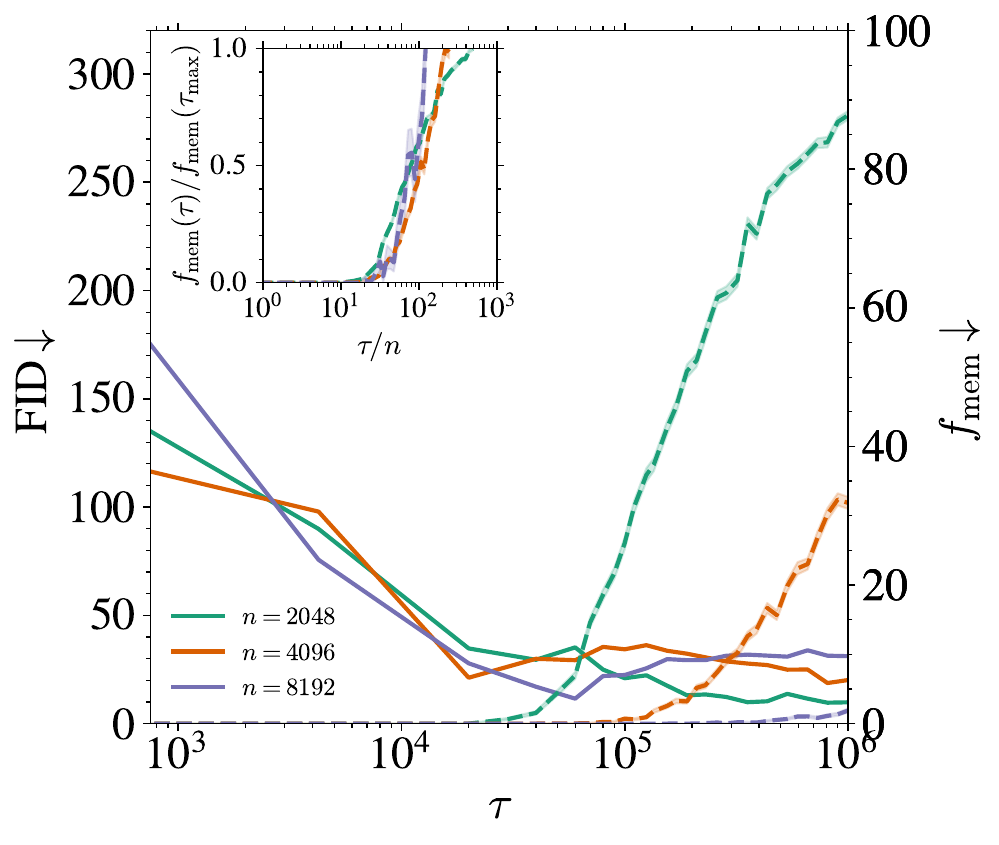}
    \caption{\textbf{Impact of batch size and optimizer on the scaling of $\tau_\mathrm{mem}$.} FID (solid lines, left axis) and memorization fraction $f_\mathrm{mem}$ (in \%, dashed lines, right axis) against training time $\tau$ for various $n$. Inset: normalized memorization fraction $f_\mathrm{mem}(\tau)/f_\mathrm{mem}(\tau_\mathrm{max})$ with the rescaled time $\tau/n$. \textit{(Left)} Memorization scaling for $B=n$. \textit{(Right)} Generalization--Memorization transition with Adam optimizer for $W=64$.
    }
    \label{fig:CelebA_batch_adam}
\end{figure}

\subsection{Batch-size effect: repetition vs. memorization}

All the experiments in the MT use a fixed batch size $B=512$, and in Sect~\ref{sect:Numerical_results} we emphasize that the observed $\mathcal{O}(n)$ scaling of $\tau_\mathrm{mem}$ cannot be explained by repetition over training samples. To validate this statement, the left panel of Fig.~\ref{fig:CelebA_batch_adam} shows FID and memorization fraction curves when we train the models with full-batch updates ($B=n$) for $n\in\left[128, 512\right]$. At any fixed $\tau$, every sample has been seen exactly $\tau$ times. Yet $\tau_\mathrm{mem}$ continues to grow linearly with $n$, as shown in the inset. This demonstrates that larger datasets reshape the loss landscape -- requiring proportionally more updates to overfit -- rather than simply increasing memorization through repeated exposure of training samples.

\subsection{What about Adam?}
We conclude this section by repeating our analysis at fixed $W=64$ using the Adam optimizer \cite{kingma2015Adam} instead of SGD with momentum. The learning rate is $\eta = 1\times10^{-4}$, gradient averages take values $(\beta_1, \beta_2)=(0.9, 0.999)$, and batch size $B=\min(512, n)$. We keep all other settings and evaluation metrics as above. As shown in the right panel of Fig.~\ref{fig:CelebA_batch_adam}, Adam yields the same two-phase training dynamics with first a generalization regime with $f_\mathrm{mem}=0$ and good performances (small FID), and later a memorization phase at $\tau_\mathrm{mem}\propto n$, as shown in the inset. The only difference is that both $\tau_\mathrm{gen}$ and $\tau_\mathrm{mem}$ occur after much fewer steps compared to SGD. This also points out that the emergence of the two well-separated timescales and their scaling is a fundamental property of the loss landscape.

\section{Generalization--memorization transition in the Gaussian Mixture Model} \label{appendix:numerical_GMM}

The aim of this section is to show our results hold for other data distributions than natural images, and alternative score model that U-Net architectures.

\subsection{Settings}

\paragraph*{Data distribution.} We focus on data iid sampled from a $d$-dimensional Gaussian Mixture Model (GMM) made of two balanced Gaussians centered on $\pm \vmu$ with unit covariance, i.e.,
\begin{equation}
    \mathbb{P}_0 = \frac{1}{2} \mathcal{N}(\vmu, \bm{I}_d) + \frac{1}{2}\mathcal{N}(-\vmu, \bm{I}_d).
\end{equation}
In what follows, we choose to work with $\vmu = \bm{1}_d$, with $\bm{1}_d = \left[1, \ldots, 1\right]\tran \in \mathbb{R}^d$. In this controlled setup, the generalization score can be computed analytically from $\mathbb{P}_0$ and reads
\begin{equation}
    \vs_\mathrm{gen}(\vx_t, t) = \vmu e^{-t} \tanh\left(\vx_t \cdot \vmu e^{-t}\right) - \vx_t.
\end{equation}

\begin{figure}
    \centering
    \includegraphics[width=0.8\linewidth]{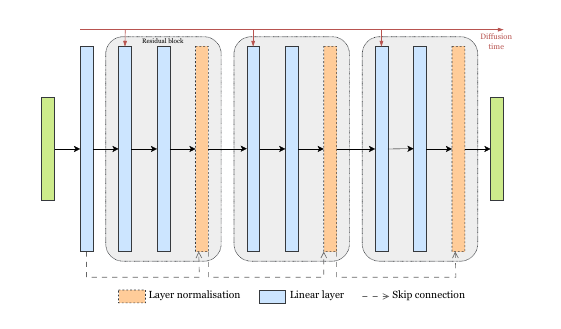}
    \caption{\textbf{Basic ResNet architecture of the GMM numerical experiments.} Residual network with three residual blocks, each made of two fully-connected layers followed by a layer normalization. The width of the network is $W$, and the input and output sizes are $d$.
    }
    \label{fig:architecture}
\end{figure}

\paragraph*{Score model.} The denoise $\vxi_{\vtheta}(\vx_t, t)$ is implemented as a lightweight residual multi-layer neural network (see Fig.~\ref{fig:architecture}): an input layer projecting $\mathbb{R}^d \to \mathbb{R}^W$, followed by three identical residual blocks and an output layer projecting back to $\mathbb{R}^d$. Each block consists of two fully connected layers of width $W$, a skip connection, and a layer normalization. We encode the diffusion time $t$ via sinusoidal position embedding and add it to the first feature of each block. The total number of parameter in the network is $p(d, W) = W(2d + 13) + d + 6W^2$. For $d=8$, and $W=128$, the reference setting of this section, this yields $p=102,024$ trainable parameters.

\paragraph*{Training and computing resources.} Unless otherwise specified, we train every model of this section with SGD at fixed learning rate $\eta = 6\times 10^{—3}$ and momentum $\beta=0.95$ using full-batch updates $B=n$ for $n\in\{128, 256, 512, 1024, 2048, 4096\}$, running for up to 4M updates. All experiments are executed on an Nvidia RTX 2080 Ti, with the largest $n=4096$ requiring around 10 hours to complete.

\paragraph*{Generalization and memorization metrics.} In addition to the memorization fraction $f_\mathrm{mem}(\tau)$, we exploit this controlled setting where we know the true data distribution $\mathbb{P}_0$ to directly measure how closely it matches the generated distribution $\mathbb{P}_{\vtheta}$ via the Kullback-Leibler (KL) divergence
\begin{equation}
    D_\mathrm{KL}(\mathbb{P}_{\vtheta} | \mathbb{P}_0) = \int \dd \vx \mathbb{P}_{\vtheta}(\vx) \log \mathbb{P}_{\vtheta}(\vx) - \int \dd \vx \mathbb{P}_{\vtheta}(\vx) \log \mathbb{P}_{0}(\vx).
\end{equation}
The cross-entropy term $\mathbb{E}_{\mathbb{P}_{\vtheta}}\left[\log\mathbb{P}_0\right]$ is easy to estimate using Monte Carlo,
\begin{equation}
    \int \dd \vx \mathbb{P}_{\vtheta}(\vx) \log \mathbb{P}_{0}(\vx) \approx \frac{1}{N} \sum_{\mu=1}^N \log \mathbb{P}_0(\tilde{\vx}_\mu),
\end{equation}
where $\{\tilde{\vx}_\mu\}_{\mu=1}^N$ are $N=10,000$ samples drawn from the model with parameters $\vtheta(\tau)$ at training time $\tau$. Estimating the negative entropy term $\mathbb{E}_{\mathbb{P}_{\vtheta}}\left[\log\mathbb{P}_{\vtheta}\right]$ is more challenging, since DMs only give access to the score function $\vs_{\vtheta}(\vx, t) = \nabla_{\vx} \log \mathbb{P}_{\vtheta}(\vx)$ and not the underlying probability distribution $\mathbb{P}_{\vtheta}$. We can however employ time integration to express it as a function of the score only,
\begin{equation}
    \mathbb{E}_{\mathbb{P}_{\vtheta}}\left[\log\mathbb{P}_{\vtheta}\right] \approx \int_{0}^T \dd t I(t) - \frac{d}{2} \log \left(2\pi e\right),
\end{equation}
with
\begin{equation}
    I(t) = \frac{\beta_t}{2N}\sum_{\mu=1}^N \left[ \tilde{\vx}_\mu \vs_{\vtheta}(\tilde{\vx}_\mu, t) + \vs_{\vtheta}(\tilde{\vx}_\mu, t)^2 \right].
\end{equation}
This expression assumes that the model learns an accurate representation of the score function. It is noteworthy to mention that samples are generated using standard Euler-Maruyama discretization of the backward process~\ref{eq:backward} of the MT over $T=1000$ timesteps.

\subsection{Scaling of $\tau_\mathrm{mem}$ and $\tau_\mathrm{gen}$ with $n$ and $W$}

\begin{figure}
    \centering
    \includegraphics[width=.49\linewidth]{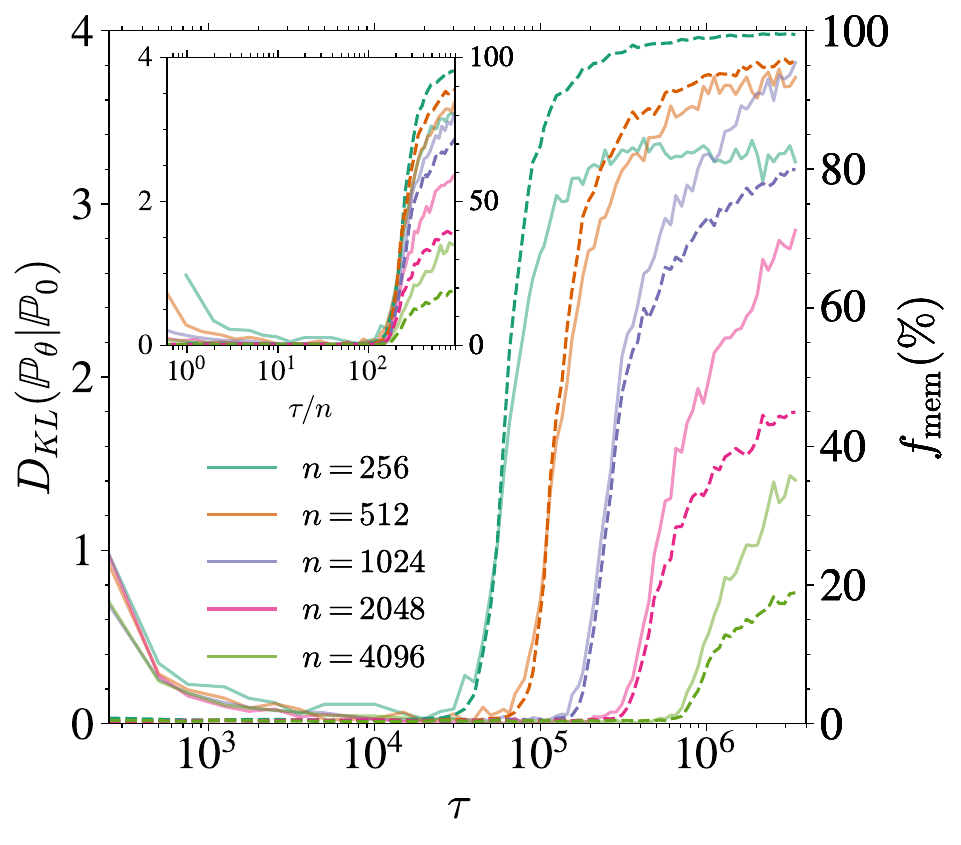}
    \includegraphics[width=.49\linewidth]{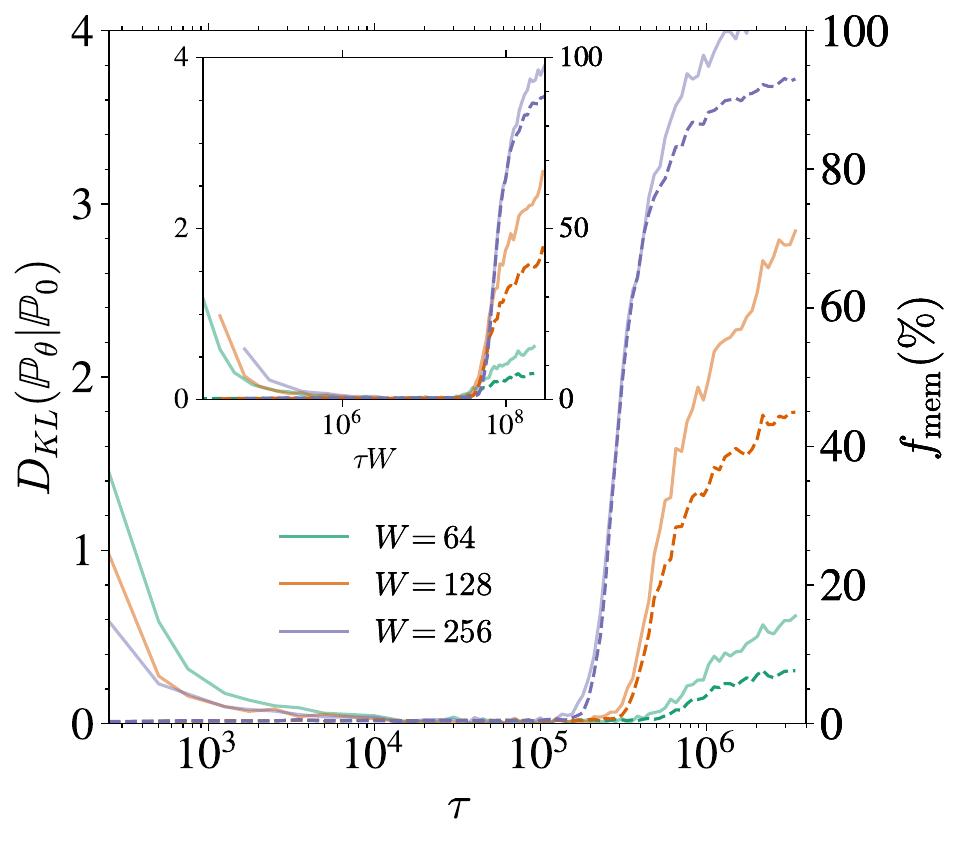}
    
    \caption{\textbf{Generalization--Memorization transition as a function of the training set size $n$ and width $W$ for ResNet score models on GMM ($d=8$).} KL divergences (solid lines, left axis) and memorization fraction $f_\mathrm{mem}$ (in \%, dashed lines, right axis) against training time $\tau$ for various \textit{(Left)} $n \in \{256, 512, 1024, 2048, 4096\}$ at fixed $W=128$. \textit{(Right)} $W \in \{64, 128, 256\}$ at fixed $n=2048$. Insets: $D_\mathrm{KL}(\mathbb{P}_{\vtheta} | \mathbb{P}_0)$ and $f_\mathrm{mem}$ against the rescaled time $\tau/n$ (left) and $\tau W$ (right).
    }
    
    \label{fig:GMM_scalings_n_W}
\end{figure}

In Fig.~\ref{fig:GMM_scalings_n_W}, the left panel shows how the KL divergence and memorization fraction evolve with training time $\tau$ for different training set sizes $n$ at fixed width $W=128$, while the right panel fixes $n=2048$ and varies $W$. In both cases, we observe two distinct phases. First, the KL divergence decreases to near zero on a timescale $\tau_\mathrm{gen}$ independent of $n$ during which the model fully generalizes ($f_\mathrm{mem}=0$). Beyond $\tau_\mathrm{gen}$, both $D_\mathrm{KL}(\mathbb{P}_{\vtheta}|\mathbb{P}_0)$ and $f_\mathrm{mem}$ begin to rise at a time $\tau_\mathrm{mem}$ that scales linearly with $n$, as highlighted by the inset of the left panel. In contrast, $\tau_\mathrm{mem}$ scales with $W^{-1}$, as shown in the inset of the right panel. While the precise dependence of $\tau_\mathrm{gen}$ with $W$ remains inconclusive in this setting and require a more careful analysis, these results on the GMM mirror the main findings of the MT: the training dynamics of DMs unfolds first in a generalization phase and only later -- at $\tau_\mathrm{mem} \propto n/W$ -- memorization begins.

\subsection{Discussion on conditional diffusion models}

\begin{figure}
    \centering
    \includegraphics[width=.55\linewidth]{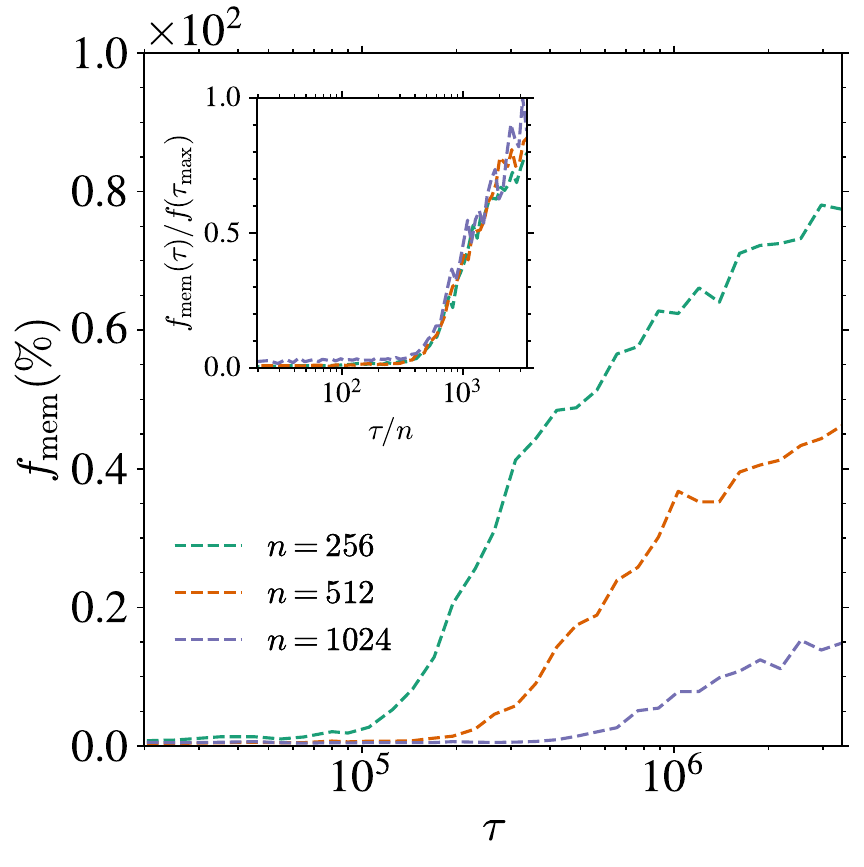}
    \caption{\textbf{Effect of guidance on $\tau_\mathrm{mem}$.} Evolution of $f_\mathrm{mem}$ as a function of $\tau$ for $n \in \{256, 512, 1024\}$ at fixed $W=64$.
    }
    \label{fig:GMM_guidance}
\end{figure}

Conditional generation aims to sample from distributions of the form $p(\vx | \vy)$, where $\vy$ denotes a conditioning variable such as a class label, a text embedding, or any other contextual information. DMs can naturally be extended to this setting using for instance classifier-free guidance \cite{ho2022classifierfree}. Although conditioning often improves sample quality in practice, memorization effects have also been observed in models trained conditionally \cite{somepalli_2023, wen2024detecting, chen2024memorizationfree}. 
Our analysis does not rely on the model being unconditional since these variables typically enter the model as additional inputs and we expect our result to hold in this setting as well.
To investigate it, we train a classifier-free guidance model to generate sample from our Gaussian mixture conditionally on the class label, and compute the memorized fraction as a function of $\tau$ that we report in Fig. \ref{fig:GMM_guidance}. In the inset, when rescaling the training time by $n$, the curves for $n \in \{256, 512, 1024\}$ all collapse perfectly, confirming that the phenomenon persists in the conditional setting.
For more complex datasets, $\tau_\mathrm{mem}$ and $\tau_\mathrm{gen}$ may in fact depend on the conditioning variable and intermediate regimes could exist where certain classes have already entered the generalization (or memorization) phase while others have not yet.

\section{Proofs of the analytical results} \label{appendix:proofs}
In the following we provide the mathematical arguments and the proofs for the statement in the MT. The section using the replica method is not mathematically rigorous but uses a well established method of theoretical physics, which has been already shown to provide correct results in several cases. The final result is rigorous, since it can alternatively be obtained from the rigorous free random matrix approach of \cite{george_2025}, as shown in Sect. \ref{proof-thm31Sigma}.

\subsection{Notations}
\label{sect:Notations}
We recall here the notations used throughout Sect.~\ref{sect:Analytical_results} of the MT and Sect.~\ref{appendix:proofs} of the SM.
\begin{align}
    &d: \text{Data dimension}\\
    &n:\text{Numbers of data points}\\
    &p:\text{Dimension of the hidden layers of the RFNN}\\
    &\bm{I}_d:\text{Identity matrix in dimension } d\\
    &\sigma(x):\text{Activation function of the model}\\
    &P_{\vx}:\text{Distribution of the data points}\\
    &P_{t}:\text{Distribution of the noisy data points at diffusion time $t$.}\\
    &\psi_n=\frac{n}{d}\\
    &\psi_p=\frac{p}{d}\\
    &\Delta_t=1-e^{-2t}\\
    &\vSigma=\mathbb{E}_{\vx\sim P_{\vx}}[\vx\vx^T]\\
    &\vSigma_t=e^{-2t}\vSigma+\Delta_t\vI_d\\
    &\Gamma_t^2=\frac{\Tr(\vSigma_t)}{d}\\
    &\sigma_{\vx}^2=\frac{\Tr(\vSigma)}{d}
\end{align}
\begin{align}
    &\lVert \sigma\rVert^2=\mathbb{E}_z[\sigma(\Gamma_t z)^2]\\
    &b_t^2=\left(\mathbb{E}_{u,v}[v\sigma(e^{-t}\sigma_{\vx}u+\sqrt{\Delta_t}v)] \right)^2\\
    &a_t=\mathbb{E}_{u,v}[\sigma(e^{-t}\sigma_{\vx}u+\sqrt{\Delta_t}v)\frac{u}{e^{-t}\sigma_{\vx}}]\\
    &v_t^2=\mathbb{E}_{u,v,w}[\sigma(e^{-t}\sigma_{\vx}u+\sqrt{\Delta_t}v)\sigma(e^{-t}\sigma_{\vx}u+\sqrt{\Delta_t}w)]-a_t^2e^{-2t}\sigma_{\vx}^2\\
    &s_t^2=\lVert \sigma \rVert^2-a_t^2e^{-2t}\sigma_{\vx}^2-v_t^2-b_t^2\\
    &\mu_1(t)=\mathbb{E}_{u}[\sigma(\Gamma_tu)u]=\sqrt{e^{-2t}\sigma_{\vx}^2a_t^2+b_t^2}.
\end{align}
Unless specified, all the expectation values are taken for standard Gaussian variables. We will denote
\begin{align}
    \vX=[\vx^1\lvert...\rvert\vx^n]\in\mathbb{R}^{d\times n}
\end{align}
the matrix whose columns are the data point vectors and likewise we decompose $\vW$ as
\begin{align}
    \vW =
    \begin{bmatrix}
        (\vW_1)^{\!\top} \\
        \vdots \\
        (\vW_p)^{\!\top}
    \end{bmatrix}
    \in \mathbb{R}^{p\times d},
\end{align}
where $\vW_i \in \mathbb{R}^{ d}$ denotes the $i$th row of $\vW$.
We recall the definitions of the matrices $\vU$ and $\vV$
\begin{align}
    &\vU=\frac{1}{n}\sum_{\nu=1}^n\mathbb{E}_{\vxi}\left[\sigma\left(\frac{\vW \vx_t^\nu(\vxi)}{\sqrt{d}}\right)\sigma\left(\frac{\vW \vx_t^\nu(\vxi)}{\sqrt{d}}\right)^T\right],\\
    &\vV=\frac{1}{n}\sum_{\nu=1}^n\mathbb{E}_{\vxi}\left[\sigma\left(\frac{\vW \vx_t^\nu(\vxi)}{\sqrt{d}}\right)\vxi^T\right].
\end{align}

\subsection{Closed form of the learning dynamics}

\begin{proposition}
\label{prop:A_tau}
    Let $\vA(\tau)$ be the solution of the gradient flow (\ref{eq:gda}) defined in the MT with initial conditions $\vA(\tau=0)=\vA_0$, then
    \begin{align}
    \frac{\vA(\tau)}{\sqrt{p}}=-\frac{1}{\sqrt{\Delta_t}}\vV^T\vU^{-1}+(\frac{1}{\sqrt{\Delta_t}}\vV^T\vU^{-1}+\frac{\vA_0}{\sqrt{p}})e^{-\frac{2\Delta_t}{\psi_p}\vU \tau}
\end{align}
with
\begin{align}
    \vV=\frac{1}{n}\sum_{\nu=1}^n\mathbb{E}_{\vxi}[\sigma(\frac{\vW \vx_t^\nu(\vxi)}{\sqrt{d}})\vxi^T].
\end{align}
\end{proposition}
\begin{proof}
    We expand the square in the training loss
    \begin{align}
    \mathcal{L}_\mathrm{train}(\vA)=1+\frac{\Delta_t}{d} \Tr(\frac{\vA^T}{\sqrt{p}}\frac{\vA}{\sqrt{p}}\vU)+\frac{2\sqrt{\Delta_t}}{d}\Tr(\frac{\vA}{\sqrt{p}}\vV)
\end{align}
and compute the gradient
\begin{align}
    \nabla_{\vA}\mathcal{L}_\mathrm{train}(\vA(\tau))=\frac{2\Delta_t}{d}\frac{\vA}{p}\vU+\frac{2\sqrt{\Delta_t}}{d\sqrt{p}}\vV^T.
\end{align}
Solving this ordinary differential equation yields the desired result. Consequently, the timescales of the dynamics of $\vA(\tau)$ is determined by the inverse of the eigenvalues of $\Delta_t \vU / \psi_p$.
\end{proof}

\subsection{Gaussian Equivalence Principle} \label{appendix:subsec:GEP}
As explained in~\cite{Peche2019, goldt_2021, hu2023}, the Gaussian Equivalence Principle which applies in the high dimensional setting considered here establishes an equivalence between the spectral properties of the original model and those of a Gaussian covariate model in which the nonlinear activation function is replaced by a linear term and a nonlinear term that acting as noise,
\begin{align}
     \sigma\left(\frac{ \vW\vx}{\sqrt d}\right) \to \kappa_1 \frac{\vW\vx'}{\sqrt d} + \kappa_* \veta,\quad \vx'\sim\mathcal{N}(0,\mathbb{E}_{\vx}[\vx\vx^T]), \quad \veta\sim \mathcal{N}(0,\vI_p),
    \label{eq:gep}
\end{align}
where $\kappa_1, \kappa_*$ are constants that depend on the distribution of the data and on the activation function $\sigma$ whose formula we recall
\begin{align}
    &\kappa_1=\mathbb{E}_z[\sigma(\sigma_{\vx}z)\frac{z}{\sigma_{\vx}}],\\
    &\kappa_*=\sqrt{\mathbb{E}_z[\sigma(\sigma_{\vx}z)^2]-\kappa_1^2\sigma_{\vx}^2}.
\end{align}
The expectation function are with respect to $z\sim \mathcal{N}(0,1)$ and $\sigma^2_{\vx}=\Tr(\vSigma)/d$. The Gaussian Equivalence Principle (GEP) holds if the distribution $P_{\vx}$ of the vector $\vx$ verifies
\begin{enumerate}
    \item[(i)] $P_{\vx}(\vx)$ has sub-Gaussian tails: there exists a constant $C>0$ such that for all $A \ge 0$ and each entry $\vx_i$,
\begin{align}
    \mathbb{P}(\lvert \vx_i \rvert \ge A) \le 2 \, e^{-A^2 / C}.
\end{align}
    \item[(ii)] The data covariance matrix $\vSigma=\mathbb{E}_{\vx\sim P_{\vx}}[\vx\vx^T]$ is bounded: there exists a constant $K>0$ independent of $d$ such that $\lambda_{\textrm{max}}(\vSigma)<K$ and $\frac{\Tr \vSigma}{d}<K$ where $\lambda_{\textrm{max}}(\vSigma)$ denotes the spectral norm of $\vSigma$.
\end{enumerate}

In this section, we outline the derivation of the Gaussian Equivalence Principle (GEP) for the matrices $\vU, \tilde{\vU},\vV$ and $\tilde{\vV}$ under arbitrary input covariance. This generalizes the approach developed in \cite{george_2025}, which considered only the case of data drawn from $\mathcal{N}(0,\vI_d)$. A more rigorous approach, which would consist in following \cite{mei2020}, is left for future works.
We will make use of the Mehler kernel formula~\cite{kibble1945} which states that for $f$ a test function defined on $\mathbb{R}^2$,
\begin{align}
\mathbb{E}_{u,v\sim P^\gamma}[f(u,v)]=\sum_{s=1}^\infty \frac{\gamma^s}{s!}\mathbb{E}_{u,v\sim\mathcal{N}(0,\vI_2)}[He_s(u)He_s(v) f(u,v)],
\end{align}
where the expectation on the left-hand side is taken over jointly Gaussian random variables $u$ and $v$ with zero mean, unit variance, and correlation $\gamma$, while on the right-hand side the expectation is taken over independent standard Gaussian variables. $He_s$ denotes the $s$-th Hermite polynomial. We recall some useful properties of the Hermite polynomials~\cite{Bach2023Hermite}:
\begin{itemize}
    \item They form an orthogonal base of $L^2(\mathbb{R},\frac{e^{-x^2/2}}{\sqrt{2\pi}}\dd x)$.
    \item The first Hermite polynomials are $He_0(x)=1,He_1(x)=x$.
\end{itemize}
\begin{lemma}[Gaussian Equivalence Principle for $\vU$]
\label{lem:GEP_U}
    In the limit $n,p,d\rightarrow\infty$ with $\psi_p=p/d, \psi_n=n/d$ and with a dataset $\{\vx^\nu\}_{\nu=1}^n$ sampled from a distribution $P_{\vx}$ which verifies assumptions (i) and (ii) with $\vSigma=\mathbb{E}_{P_{\vx}}[\vx\vx^T]$, the matrix
    \begin{align}
    \vU=\frac{1}{n}\sum_{\nu=1}^n\mathbb{E}_{\vxi}[\sigma(\frac{\vW \vx_t^\nu(\vxi)}{\sqrt{d}})\sigma(\frac{\vW \vx_t^\nu(\vxi)}{\sqrt{d}})^T]
\end{align}
has the same spectrum as its Gaussian equivalent
\begin{align}
    \vU=\frac{\vG}{\sqrt{n}}\frac{\vG^T}{\sqrt{n}}+b_t^2\frac{\vW\vW^T}{d}+s_t^2\vI _p
\end{align}
where
\begin{align}
    \vG=e^{-t} a_t\frac{\vW}{\sqrt{d}}\vX'+v_t\vOmega,
\end{align}
$\vX'\in\mathbb{R}^{d\times n}$ is a matrix whose columns $\vx'^\nu$ are sampled according to $\mathcal{N}(0,\vSigma)$ and $\vOmega\in \mathbb{R}^{p\times d}$ has gaussian entries independent of $\vX$ and $\vW$.
\end{lemma}

\begin{proof}
For the sake of clarity, in this proof we explicitly make the covariance of the data $\vSigma$ appear by writing the data points are written as $\vx^\nu=\vSigma^{1/2}\vz^\nu$ where the vectors $\vz^\nu$ have variance 1. Let us focus on the element of $\vU$ in position $(i,j)$
\begin{align}
    \vU_{ij}=\frac{1}{n}\sum_{\nu=1}^n\mathbb{E}_{\vxi}[\sigma\left(\frac{\vW_{ik} (e^{-t}(\sqSigma)_{kl}\vz^\nu_l+\sqrt{\Delta_t}\vxi_k)}{\sqrt{d}}\right)\sigma\left(\frac{\vW_{jk'} (e^{-t}(\sqSigma)_{k'l'}\vz^\nu_{l'}+\sqrt{\Delta_t}\vxi_{k'})}{\sqrt{d}}\right)],
\end{align}
where repeated indices mean that there is a hidden sum. We introduce the random variable $\chi_i^\nu=\frac{\vW_{ik} (e^{-t}(\sqSigma)_{kl}\vz^\nu_l+\sqrt{\Delta_t}\vxi_k)}{\sqrt{d}}$. In the high dimensional limit it converges to a Gaussian random variable by the Central Limit Theorem (since the tails of the data distribution are sub-Gaussian). If $i=j$, the diagonal terms concentrate with respect to the data points and we can thus replace the sum by an average
\begin{align}
    \vU_{ii}=\mathbb{E}_{\chi}[\sigma(\chi)^2]+\mathcal{O}(1/n).
\end{align}
The finite $n$ corrections can be discarded because they cannot change the spectrum of $\vU$. $\chi$ can be taken Gaussian with mean 0 and covariance $\mathbb{E}_{\vW_i,\vz,\xi}[\chi^2]=\mathbb{E}_{\vW_i,\vz,\xi}[\frac{\vW_i^T\vSigma_t\vW_i}{d}]=\frac{\Tr(\vSigma_t)}{d}=\Gamma_t^2$ hence
\begin{align}
    \vU_{ii}=\mathbb{E}_{\chi}[\sigma(\chi)^2]+\mathcal{O}(1/n)=\mathbb{E}_{u\sim\mathcal{N}(0,1)}[\sigma(\Gamma_tu)^2]=\lvert \lvert \sigma\rvert\rvert^2.
\end{align}
If $i\neq j$, we denote $\eta_i^\nu=e^{-t}\frac{\vW_i^T\sqSigma\vz}{\sqrt{d}}$. For now we consider $\vW$ and the $\vz^\nu$ fixed and look at $\xi$. We use the Mehler Kernel formula for the random variables $u=\vW_i^T\vxi/\sqrt{d}$ and $v=\vW_j^T\vxi/\sqrt{d}$ that have correlation $\mathbb{E}_{\vxi}[uv]=\frac{\vW_i^T\vW_j}{d}$
\begin{align}
    &\mathbb{E}_{u,v}[\sigma\left(\eta_i^\nu+\sqrt{\Delta_t}u\right)\sigma\left(\eta_j^\nu+\sqrt{\Delta_t}v\right)]\\
    &=\sum_{s=1}^\infty\frac{(\vW_i^T\vW_j/d)^s}{s!}\mathbb{E}_u[He_s(u)\sigma\left(\eta_i^\nu+\sqrt{\Delta_t}u\right)]\mathbb{E}_u[He_s(v)\sigma\left(\eta_j^\nu+\sqrt{\Delta_t}v\right)].
\end{align}
We truncate at order $s=1$ since the corrections are order $\mathcal{O}(1/d)$.
\begin{align}
    \vU_{ij}&=\frac{1}{n}\sum_{\nu=1}^n \mathbb{E}_u[\sigma\left(\eta_i^\nu+\sqrt{\Delta_t}u\right)]\mathbb{E}_v[\sigma\left(\eta_j^\nu+\sqrt{\Delta_t}v\right)]\\
    &+\frac{1}{n}\sum_{\nu=1}^n\frac{\vW_i^T\vW_j}{d}\mathbb{E}_u[u\sigma\left(\eta_i^\nu+\sqrt{\Delta_t}u\right)]\mathbb{E}_v[v\sigma\left(\eta_j^\nu+\sqrt{\Delta_t}v\right)]\\
    &=\frac{1}{n}\sum_{\nu=1}^n \mathbb{E}_u[\sigma\left(\eta_i^\nu+\sqrt{\Delta_t}u\right)]\mathbb{E}_v[\sigma\left(\eta_j^\nu+\sqrt{\Delta_t}v\right)]\\
    &+\frac{\vW_i^T\vW_j}{d}\mathbb{E}_{\veta}[\mathbb{E}_u[u\sigma\left(\veta_i+\sqrt{\Delta_t}u\right)]\mathbb{E}_v[v\sigma\left(\veta_j+\sqrt{\Delta_t}v\right)]].
\end{align}
by neglecting $\mathcal{O}(1/d)$ corrections and where the law of $\eta$ can be considered Gaussian with zero mean correlation $\mathbb{E}[\eta_i^\nu \eta_j^\nu]=\frac{e^{-2t}\Tr(\vSigma)}{d}\delta_{ij}=e^{-2t}\sigma_{\vx}^2\delta_{ij}$. The coefficient in front of $\frac{\vW_i^T\vW_j}{d}$ is therefore
\begin{align}
    b_t^2=(\mathbb{E}_{u,v}[v\sigma(e^{-t}\sigma_{\vx}u+\sqrt{\Delta_t}v)])^2.
\end{align}
Denote $\sigma_0(\eta)=\mathbb{E}_u[\sigma(\eta+\sqrt{\Delta_t}u)]$. We now focus on
\begin{align}
\frac{1}{n}\sum_\nu \sigma_0(\eta_i^\nu)\sigma_0(\eta_j^\nu).
\end{align}
We use the GEP on $\sigma_0$ 
\begin{align}
     \sigma_0\left(\frac{e^{-t} \vW_i^T\vx^\nu}{\sqrt d}\right) \to a_t e^{-t}\frac{\vW_i^T\vx'^\nu}{\sqrt d} + \
     v_t \vOmega_i^\nu,\quad \vx'^\nu\sim\mathcal{N}(0,\vSigma), \quad \vOmega_i^\nu\sim \mathcal{N}(0,\vI_{p}),
\end{align}
with $a_t=\mathbb{E}_u[\sigma_0(e^{-t}\sigma_{\vx}u)\frac{u}{e^{-t}\sigma_{\vx}}]=\mathbb{E}_{u,v}[\sigma(e^{-t}\sigma_{\vx} u+\sqrt{\Delta_t }v)\frac{u}{e^{-t}\sigma_{\vx}}]$ and $v_t^2=\mathbb{E}_u[\sigma_0(e^{-t}\sigma_{\vx}u)^2]-a_t^2e^{-2t}\sigma_{\vx}^2=\mathbb{E}_{u,v,w}[\sigma(e^{-t}\sigma_{\vx}u+\sqrt{\Delta_t}v)\sigma(e^{-t}\sigma_{\vx}u+\sqrt{\Delta_t}w)]-a_t^2e^{-2t}\sigma_{\vx}^2$. Hence the truncated expansion yields for $i\neq j$
\begin{align}
    \vU_{ij}=\frac{1}{n}\sum_{\nu=1}^n\left( a_te^{-t}\frac{\vW_i^T\vx'^\nu}{\sqrt{d}}+v_t\vOmega_i^\nu\right)\left(a_te^{-t}\frac{\vW_j^T\vx'^\nu}{\sqrt{d}}+v_t\vOmega_j^\nu \right)^T+b_t^2\frac{\vW_i^T\vW_j}{d}.
\end{align}
Now we need to deal with the diagonal term. We need to substract 
\begin{align}
    \left(a_t^2e^{-2t}\sigma_{\vx}^2+v_t^2+b_t^2\right)\vI_p.
\end{align}
The Gaussian equivalent of $\vU$ reads
\begin{align}
    \vU=\frac{\vG}{\sqrt{n}}\frac{\vG^T}{\sqrt{n}}+b_t^2\frac{\vW\vW^T}{d}+s_t^2\vI_p,
\end{align}
with $s_t^2=\lVert \sigma \rVert^2- a_t^2e^{-2t}\sigma_{\vx}^2-v_t^2-b_t^2$.

\end{proof}
\begin{lemma}[GEP for $\tilde{U}$]
    Let 
    \begin{align}
        \tilde{\vU}=\mathbb{E}_{\vy}[\sigma(\frac{\vW \vy}{\sqrt{d}})\sigma(\frac{\vW \vy}{\sqrt{d}})^T],
    \end{align}
    where the expectation value is taken $\vy\sim P_t$. Then the GEP of $\tilde{U}$ reads
    \begin{align}
        \mu_1^2(t)\frac{\vW\vSigma_t\vW^T}{d}+\left( \lVert \sigma \rVert^2-\mu_1^2(t)\right)\vI_p,
    \end{align}
    where $\mu_1^2(t)$ and $\lVert \sigma \rVert^2$ are defined in Sect.~\ref{sect:Notations}.
\end{lemma}
\begin{proof}
For a vector $\vy$ sampled from $P_t$, the $\frac{\vW_i^T\vy}{\sqrt{d}}$ are asymptotically Gaussian with 0 mean, variance $\mathbb{E}_{\vy}[\frac{\vW_i^T\vy}{\sqrt{d}} \frac{\vW_i^T\vy}{\sqrt{d}}]=\frac{\vW_i^T\vSigma_t\vW_i}{d}\sim\Gamma_t^2$ and correlation $\mathbb{E}_{\vy}[\frac{\vW_i^T\vy}{\sqrt{d}} \frac{\vW_j^T\vy}{\sqrt{d}}]=\frac{\vW_i^T\vSigma_t\vW_j}{d}$. We apply Mehler Kernel formula to $\tilde{\vU}$
\begin{align}
    \tilde{\vU}_{ij}=\sum_s\frac{1}{s!}\left(\frac{\vW_{ik}(\vSigma_t)_{kl}\vW_{jl}}{\Gamma_t^2d}\right)^s \mathbb{E}_u[\sigma(\Gamma_tu)He_s(u)]\mathbb{E}_v[\sigma(\Gamma_tv)He_s(v)],
\end{align}
where the expectation on $u$ and $v$ is standard Gaussian. We keep only terms at order $\mathcal{O}(1/\sqrt{d})$. If $i\neq j$ we keep the terms up to order $s=1$.
\begin{align}
     \tilde{\vU}_{ij}=\left(\frac{\vW_{ik}(\vSigma_t)_{kl}\vW_{jl}}{\Gamma_t^2d}\right)\mathbb{E}_u[\sigma(\Gamma_tu)u]^2.
\end{align}
For $i=j$ we cannot truncate because all terms are $\mathcal{O}_d(1)$. Hence the diagonals terms are asymptotically
\begin{align}
    \tilde{\vU}_{ii}=\mathbb{E}_{u\sim\mathcal{N}(0,1)}[\sigma^2(\Gamma_t z)]=\lVert\sigma\rVert^2.
\end{align}
Taking care of the diagonal terms, the Gaussian Equivalent matrix reads
\begin{align}
    \tilde{\vU}=\frac{\mu_1^2(t)}{\Gamma_t^2}\frac{\vW\vSigma_t\vW}{d}+\left(\lVert \sigma \rVert^2-\mu_1^2(t)\right)\vI_p
\end{align}
where $\mu_1(t)=\mathbb{E}_{u}[\sigma(\Gamma_tu)u]$.

\end{proof}
Building on the GEP of $\tilde{\vU}$, we prove the following lemma on the scaling of the eigenvalues in the bulk.
\begin{lemma}[Scaling of the bulk of $\tilde{\vU}$]
\label{lem:bulk_U_tilde}
    We assume that $\vSigma$ is positive definite and that the spectral norm $\lambda_{\mathrm{max}}(\vSigma)$ stays $\mathcal{O}_d(1)$. In the high dimensional limit $p>d\gg 1$, the spectrum of $\tilde{\vU}$ is asymptotically equal to
    \begin{align}
        \left(1-\frac{1}{\psi_p}\right)\delta(\lambda-(\lVert \sigma \rVert^2-\mu_1^2(t)))+\frac{1}{\psi_p}\rho_{\mathrm{bulk}}(\lambda),
    \end{align}
    where $\rho_{\mathrm{bulk}}(\lambda)$ is an atomless measure whose support is of order $\mathcal{O}(\psi_p)$.
\end{lemma}
\begin{proof}
    Since $p>d$ and $\vW\in\mathbb{R}^{p\times d}$ and $\vSigma\in \mathbb{R}^{d\times d}$, the spectrum admits a Dirac mass at $\lambda=\lVert \sigma \rVert^2-\mu_1^2(t)$ with weight $(p-d)/p$. For the order of magnitude of the eigenvalues in the bulk, let us first observe that the bulk of $\frac{\vW^T\vSigma_t\vW}{d}$ is the same as the one of $\frac{\vSigma_t^{1/2}\vW\vW^T\vSigma_t^{1/2}}{d}$. We can bound the spectral norm of the product by the product of the spectral norms
    \begin{align}
        \lambda_{\mathrm{max}}(\frac{\vSigma_t^{1/2}\vW\vW^T\vSigma_t^{1/2}}{d})\le  \lambda_{\mathrm{max}}(\frac{\vW\vW^T}{d}) \lambda_{\mathrm{max}}(\vSigma_t)\lesssim\mathcal{O}(\psi_p),
    \end{align}
    since we assumed that $\lambda_{\mathrm{max}}(\vSigma_t)=e^{-2t}\lambda_{\mathrm{max}}(\vSigma_t)+\Delta_t=\mathcal{O}(1)$ and since $\lambda_{\mathrm{max}}(\frac{\vW\vW^T}{d})=\mathcal{O}(\psi_p)$ is given by the Marchenko-Pastur law \cite{Potters_Bouchaud_2020}. To bound the norm from below we use the following inequality
    \begin{align}
        \lambda_{\mathrm{min}}(\vSigma_t)\lambda_{\mathrm{min}}(\frac{\vW\vW^T}{d})\le \lambda_{\mathrm{min}}(\frac{\vSigma_t^{1/2}\vW\vW^T\vSigma_t^{1/2}}{d}).
    \end{align}
    Since $\vSigma_t$ is positive definite, the bound is also of order $\psi_p$. This concludes that the support of the bulk is of order $\psi_p$.
\end{proof}

\begin{lemma}[GEP for $\vV$ and $\tilde{\vV}$]
Let
\begin{align}
    &\vV=\frac{1}{n}\sum_{\nu=1}^n\mathbb{E}_{\vxi}[\sigma(\frac{\vW \vx_t^\nu(\vxi)}{\sqrt{d}})\vxi^T],\\
    &\tilde{\vV}=\mathbb{E}_{\vx,\vxi}[\sigma(\frac{\vW \vx_t(\vxi)}{\sqrt{d}})\vxi^T].
\end{align}
    They can be replaced by their Gaussian Equivalence Principle in the train and test losses.
    \begin{align}
    \tilde{\vV}=\vV=\frac{\mu_1(t)\sqrt{\Delta_t}}{\Gamma_t}\frac{\vW}{\sqrt{d}}.
\end{align}
\end{lemma}
\begin{proof}
    The two matrices only differ element-wise by quantity of order $\mathcal{O}(1/n)$ and therefore have the same Gaussian Equivalent matrix. We focus on $\tilde{\vV}$. Introduce the random variable $\veta_i=\frac{\vW_{ik}(e^{-t}\vx_k+\sqrt{\Delta_t}\vxi_k)}{\sqrt{d}}$. Its has 0 mean, covariance $\mathbb{E}_{\vx, \vxi}[\veta_i^2]=\frac{\vW_i^T\vSigma_t\vW_i}{d}\sim\Gamma_t^2$ and correlation with $\vxi$ $\gamma_{ij}=\mathbb{E}_{\vx,\vxi}[\veta_i\vxi_j]=\frac{\sqrt{\Delta_t}\vW_{ij}}{\sqrt{d}}$. We apply the Mehler Kernel formula
    \begin{align}
        \tilde{\vV}_{ij}&=\mathbb{E}_{\vx,\vxi}[\sigma\left(\Gamma_t(\frac{\vW_{ik}(e^{-t}(\vSigma_t)_{kl}\vz_l+\sqrt{\Delta_t}\xi_l)}{\Gamma_t\sqrt{d}})\right)\xi_{j}]\\
        &=\sum_s\frac{1}{s!}\left( \frac{\vW_{ij}\sqrt{\Delta_t}}{\Gamma_t\sqrt{d}}\right)^s\mathbb{E}_u[\sigma(\Gamma_t u)He_s(u)]\mathbb{E}_v[vHe_s(v)]\\
        &=0+\frac{\sqrt{\Delta_t}}{\Gamma_t}\frac{\vW_{ij}}{\sqrt{d}}\mathbb{E}_u[\sigma(\Gamma_t u)u]\mathbb{E}_v[v^2]+\mathcal{O}(\frac{1}{d})\\
&=\frac{\sqrt{\Delta_t}\mu_1(t)}{\Gamma_t}\frac{\vW_{ij}}{\sqrt{d}}.
    \end{align}
\end{proof}


\subsection{Proof of Theorem~\ref{thm:Saddle_point_equations_new}} 
\label{proof-thm31Sigma}
We recall the Theorem~\ref{thm:Saddle_point_equations_new} of the MT.

\begin{thm}
\label{thm:Saddle_point_equations_new2}
    Let $q(z)=\frac{1}{p}\Tr(\vU-z\vI_p)^{-1}$, $r(z)=\frac{1}{p}\Tr(\vSigma^{1/2}\vW^T(\vU-z\vI_p)^{-1}\vW\vSigma^{1/2})$ and $s(z)=\frac{1}{p}\Tr(\vW^T(\vU-z\vI_p)^{-1}\vW)$, with $z\in\mathbb{C}$. Let
    \begin{align}
        &\hat{s}(q)=b_t^2\psi_p+\frac{1}{q},\\
    &\hat{r}(r,q)=\frac{\psi_p a_t^2e^{-2t}}{1+\frac{a_t^2e^{-2t}\psi_p 
}{\psi_n }r+\frac{\psi_p v_t^2}{\psi_n }q}.
    \end{align}
    
    Then $q(z), r(z)$ and $s(z)$ satisfy the following set of three equations:
    \begin{align}
    &s=\int\dd \rho_{\vSigma}(\lambda)\frac{1}{\hat{s}(q)+\lambda\hat{r}(r,q)},\\
    &r=\int\dd \rho_{\vSigma}(\lambda)\frac{\lambda}{\hat{s}(q)+\lambda\hat{r}(r,q)},\\
    &\psi_p(s_t^2-z)+\frac{\psi_pv_t^2}{1+\frac{a_t^2e^{-2t}\psi_p} {\psi_n} r+\frac{\psi_p v_t^2}{\psi_n} q}+\frac{1-\psi_p}{q}-\frac{s}{q^2}=0,
\end{align}
The eigenvalue distribution of $\vU$, $\rho(\lambda)$, can then be obtained using the Sokhotski–Plemelj inversion formula $\rho(\lambda)=\underset{\varepsilon \rightarrow0^+}{\lim}\frac{1}{\pi}\operatorname{Im}q(\lambda+i\varepsilon)
$. 
\end{thm}

We first show that the equations of the Stieltjes transform of $\rho$ found in Ref.~\cite{george_2025} with linear pencils \cite{bodin2024random} in the case $P_{\vx}=\mathcal{N}(0,\vI_d)$ i.e. $\rho_{\vSigma}(\lambda)=\delta(\lambda-1)$ can be reduced to the equations of Theorem~\ref{thm:Saddle_point_equations_new} with our definitions of $\mu_1(t), s_t$ and $v_t$. The equations of \cite{george_2025} read
\begin{align}
    &\zeta_1(s_t^2-z+e^{-2t}\mu_1^2\zeta_2\zeta_3+v_t^2\zeta_2+\Delta_t\mu_1^2\zeta_4)-1=0\\
    &\zeta_2(\psi_n+v_t^2\psi_p\zeta_1-e^{-t}\mu_1\zeta_3)-\psi_n=0\\
    &e^{-t}\mu_1\psi_p\zeta_1(1+e^{-t}\mu_1\zeta_2\zeta_3)+(1+(\Delta_t\mu_1^2\psi_p\zeta_1)\zeta_3)=0\\
    &e^{-2t}\mu_1^2\psi_p\zeta_1\zeta_2\zeta_4+(1+\Delta_t\mu_1^2\psi_p\zeta_1)\zeta_4-1=0,
\end{align}
with $\zeta_1=q$ and $\zeta_{2,3,4}$ auxiliary variables. We make the following change of variables $r=-\frac{\zeta_3}{e^{-t}\mu_1\psi_p}$. The second equations relates $\zeta_2$ to $q$ and $r$
\begin{align}
    \zeta_2=\frac{1}{1+\frac{e^{-2t}\mu_1^2\psi_p}{\psi_n}r+\frac{v_t^2\psi_p}{\psi_n}q}.
\end{align}
Injecting this into the second equations gives the second equation of Theorem~\ref{thm:Saddle_point_equations_new}. The fourth equation gives 
\begin{align}
    \zeta_4=\frac{1}{1+\mu_1^2\psi_pq(\Delta_t+e^{-2t}\zeta_2)}.
\end{align}
Injecting this into the first equation gives
\begin{align}
    q(s_t^2-z+e^{-2t}\mu_1^2\zeta_2r(-e^{-t}\mu_1\psi_p)+v_t^2\zeta_2+\Delta_t\mu_1^2\frac{1}{1+\mu_1^2\psi_pq(\Delta_t+e^{-2t}\zeta_2)})-1=0.
\end{align}
After some massaging we find back the first equation of Theorem~\ref{thm:Saddle_point_equations_new}.\\
We now prove Theorem~\ref{thm:Saddle_point_equations_new} using a replica computation, inspired by the calculation done in Ref.~\cite{Ascoli_2020}.
\begin{proof}
Our goal is to compute the Stieltjes transform of the matrix $\vU$.
   \begin{align}
       q&=\underset{p\rightarrow\infty}{\lim}\frac{1}{p}\mathbb{E}_{\vW,\vX,\vOmega}[\Tr(\vU-z\vI_p)^{-1}]\\
       &=-\partial_z \underset{p\rightarrow\infty}{\lim}\frac{1}{p}\mathbb{E}_{\vW,\vX,\vOmega}[\log\det (\vU-z\vI_p)]\\
&=2\partial_z\underset{p\rightarrow\infty}{\lim}\frac{1}{p}\mathbb{E}_{\vW,\vX,\vOmega}[\log\det (\vU-z\vI_p)^{-1/2}].
\end{align}
The so-called \textit{replica trick} consists of replacing the $\log x$ by $\underset{s\rightarrow\infty}{\lim}\frac{x^s-1}{s}$. Applying this identity, we obtain 
\begin{align}
       q=2\partial_z\underset{s\rightarrow0}{\lim}\underset{p\rightarrow\infty}{\lim}\frac{1}{ps}\mathbb{E}_{\vW,\vX,\vOmega}[\det (\vU-z\vI_p)^{-s/2}-1],
   \end{align}
where as usual with replica computations we have inverted the order of the limits $p\rightarrow \infty$ and $s\rightarrow 0$. We define the partition function $\mathcal{Z}$ as
\begin{align}
   \mathcal{Z}= \det(\vU-z\vI_p)^{-1/2}=\int \frac{\dd\phi}{(2\pi)^{p/2}}e^{-\frac{1}{2}{\phi}^T(\vU-z\vI_p)\phi}.
\end{align}
We replace $\vU$ by its Gaussian equivalent proved in Lemma~\ref{lem:GEP_U} and write the partition function for an arbitrary integer $s$
\begin{align}
    \mathbb{E}_{\vW,\vX,\vOmega}[\mathcal{Z}^s]&=\int\prod_{a=1}^s \frac{\dd\phi^a}{(2\pi)^{p/2}}\mathbb{E}_{\vW,\vX,\vOmega}[e^{-\frac{1}{2}{\phi^a}^T(\vU-z\vI_p)\phi^a}]\\
    &=\int\prod_{a=1}^s \frac{\dd\phi^a}{(2\pi)^{p/2}}e^{\frac{1}{2}{\phi^a}^T(z-s_t^2)\phi^a}\nonumber \\
    &\ \  \ \mathbb{E}_{\vW,\vX,\vOmega}[e^{-\frac{1}{2n}{\phi^a}^T\left(a_te^{-t}\frac{\vW\vX^\nu}{\sqrt{d}}+v_t\vOmega^\nu\right)\left(a_te^{-t}\frac{\vW\vX^\nu}{\sqrt{d}}+v_t\vOmega^\nu\right)^T\phi^a}e^{-\frac{b_t^2}{2d}{\phi^a}^T\vW\vW^T\phi^a}].
\end{align}
We first perform the computation for integer values of $s$, and then analytically continue the result to the limit $s \to 0$. To compute the expectation over $\vX$, $\vW$, and $\vOmega$, we need the following standard result from Gaussian integration
\begin{align}
    \int \dd \vx e^{-\frac{1}{2}\vx\vG\vx^T+\vJ\vx^T}=e^{-\frac{1}{2}\log \det\vG+\frac{1}{2}\vJ\vG^{-1}\vJ^T},
\end{align}
where $\vG$ is a square matrix and $\vJ$ a vector.
\paragraph*{Averaging over the data set.}
The dataset dependence enters through 
\begin{align}
    &\mathbb{E}_{\vX}[e^{-\frac{1}{2n}{\phi^a}^T\left(a_te^{-t}\frac{\vW\vX^\nu}{\sqrt{d}}+v_t\vOmega^\nu\right)\left(a_te^{-t}\frac{\vW\vX^\nu}{\sqrt{d}}+v_t\vOmega^\nu\right)^T\phi^a}]\nonumber \\
    &=\mathbb{E}_{\vX}[e^{-\frac{a_t^2e^{-2t}}{2nd}{\phi^a}^T \vW \vX^\nu {\vX^\nu}^T\vW^T\phi^a}e^{-\frac{a_te^{-t}v_t}{2\sqrt{d}n}{\phi^a}^T(\vW \vX^\nu\vOmega^T+\vOmega {\vX^\nu}^T\vW^T)\phi^a}]e^{-\frac{v_t^2}{2n}{\phi^a}^T\vOmega \vOmega^T{\phi^a}}.
\end{align}
We introduce for each replica $\phi^a$ a Fourier transform of the delta function by using the auxiliary variables $\omega^a, \hat{\omega}^a \in \mathbb{R}^{d}$ as\footnote{Throughout the computation, we discard non-exponential prefactors, as they give subleading contributions.}
\begin{align}
    \int \dd\hat{\omega}^a e^{i\hat{\omega}^a(\sqrt{p}\omega^a- {\phi^a}^T\vW\sqSigma)}=1. 
\end{align}
In the following, we do the change of variable $\vX^\nu=\sqSigma \vZ^\nu$ with $\vZ^\nu$ a $d$ dimensional Gaussian random variable with unit variance.
\begin{align}
    &\mathbb{E}_{\vX}[e^{-\frac{1}{2n}{\phi^a}^T\left(a_te^{-t}\frac{\vW\vX^\nu}{\sqrt{d}}+v_t\vOmega^\nu\right)\left(a_te^{-t}\frac{\vW\vX^\nu}{\sqrt{d}}+v_t\vOmega^\nu\right)^T\phi^a}]\nonumber \\
    &=\mathbb{E}_{\vZ}[e^{-\frac{a_t^2e^{-2t}p}{2nd}{\omega^a}^T\vZ^\nu {\vZ^\nu}^T\omega^a}e^{-\frac{a_t e^{-t}v_t\sqrt{p}}{\sqrt{d}n}\sum_{a,\nu}\vOmega^\nu \phi^a \omega^a\cdot \vZ^\nu}]e^{-\frac{v_t^2}{2n}{\phi^a}^T\vOmega \vOmega^T{\phi^a}}.
\end{align}
Denote $\vG_{\vZ}=\frac{a_t^2p}{dn}\sum_a\omega^a{\omega^a}^T$ and $(\vJ_{\vZ})^\nu=\frac{a_te^{-t}v_t\sqrt{p}}{\sqrt{d}n}\sum_{a}(\vOmega^\nu\cdot \phi^a) \omega^a$, then
\begin{align}
    &\mathbb{E}_{\vX}[e^{-\frac{1}{2n}{\phi^a}^T\left(a_te^{-t}\frac{\vW\vX^\nu}{\sqrt{d}}+v_t\vOmega^\nu\right)\left(a_te^{-t}\frac{\vW\vX^\nu}{\sqrt{d}}+v_t\vOmega^\nu\right)^T\phi^a}]=\nonumber \\
    &e^{-\frac{n}{2}\log \det(1+\vG_{\vZ})}e^{\frac{a_te^{-t}v_tp}{2dn^2} \sum_\nu (\vOmega^\nu\cdot \phi^a)(\vOmega^\nu\cdot \phi^b) {\omega^a}_h(1+\vG_{\vZ})^{-1}_{k,l} {\omega^b}_l}e^{-\frac{v_t^2}{2n}{\phi^a}^T\vOmega \vOmega^T{\phi^a}},
\end{align}
where repeated indices mean that there is an implicit summation.
\paragraph*{Averaging over $\vOmega$.}
The terms that depend on $\vOmega$ are 
\begin{align}
    &\mathbb{E}_{\vOmega}[e^{\frac{a_te^{-t}v_t}{2dn^2} \sum_\nu (\vOmega^\nu\cdot \phi^a)(\vOmega^\nu\cdot \phi^b) {\omega^a}_k(1+\vG_{\vX})^{-1}_{k,l} {\omega^b}_l}e^{-\frac{v_t^2}{2n}{\phi^a}^T\vOmega \vOmega^T{\phi^a}}]\nonumber \\
    &=(\mathbb{E}_{\vOmega^\nu}[e^{\frac{a_te^{-t}v_tp}{2dn^2} (\vOmega^\nu\cdot \phi^a)(\vOmega^\nu\cdot \phi^b) {\omega^a}_k(1+\vG_{\vX})^{-1}_{k,l} {\omega^b}_l}e^{-\frac{v_t^2}{2n}{\phi^a}^T\vOmega^\nu {\vOmega^\nu}^T{\phi^a}}])^n\\
    &=e^{-\frac{n}{2}\log \det (1+\vG_{\vOmega})},
\end{align}
with
\begin{align}
    (\vG_{\vOmega})_{k,l}=\phi^a( \frac{v_t^2}{n}\delta_{ab} -\frac{a_te^{-t}v_tp}{dn^2} {\omega^a}_k(1+\vG_{\vZ})^{-1}_{k,l}{\omega^b}_{l}){\phi^b}^T.
\end{align}
We are left with
\begin{align}
    &\mathbb{E}_{\vW,\vX,\vOmega}[\mathcal{Z}^s]=\int\prod_{a=1}^s \frac{\dd\phi^a}{(2\pi)^{p/2}}\dd\omega^a d\hat{\omega}^a  e^{\frac{1}{2}(z-s_t^2)\phi^a {\phi^a}^T}e^{-\frac{b_t^2p}{2d}{\omega^a}^T\vSigma^{-1}\omega^a}\nonumber \\
    &\mathbb{E}_{\vW}[e^{i\hat{\omega}^a(\sqrt{p}\omega^a- {\phi^a}^T\vW\sqSigma)}\ e^{-\frac{n}{2}\log \det(\vI_d+\vG_{Z})}e^{-\frac{n}{2}\log \det (\vI_d+\vG_{\vOmega})}].
\end{align}

\paragraph*{Averaging over the random features $\vW$.}
$\vW$ only appears through $e^{-i\hat{\omega}^a\vW^T{\phi^a}\sqSigma}$.
\begin{align}
   \mathbb{E}_{\vW}[ e^{i\sum_a\hat{\omega}^a(\sqrt{p}\omega^a-\vW^T{\phi^a} \sqSigmat)}]&=e^{i\sqrt{p}\sum_a\hat{\omega}^a\cdot\omega^a}(\mathbb{E}_{\vW}[ e^{-i \hat{\omega}_k^a{\phi^a}_i \vW_{li}(\sqSigmat)_{kl}}])\\
   &=e^{i\sqrt{p}\hat{\omega}^a\cdot\omega^a}e^{-\frac{1}{2}\hat{\omega}^a_k(\vSigma)_{kl}\hat{\omega}^b_{l}\phi^a_i\phi^b_i}\\
   &=e^{i\sqrt{p}\sum_a\hat{\omega}^a\cdot\omega^a} e^{-\frac{1}{2}\sum_{a,b} \hat{\omega}^a  \vSigma \hat{\omega}^b \phi^a\cdot \phi^b}.
\end{align}
We end up with 
\begin{align}
    \mathbb{E}_{\vW,\vX,\vOmega}[\mathcal{Z}^s]&=\int\prod_{a=1}^s \dd\phi^a\dd\omega^a d\hat{\omega}^a  e^{\frac{1}{2}(z-s_t^2)\phi^a {\phi^a}^T}e^{-\frac{b_t^2p}{2d}{\omega^a}^T\vSigma^{-1}\omega^a}e^{i\sqrt{p}\sum_a\hat{\omega}^a\cdot\omega^a}\nonumber \\
& e^{-\frac{1}{2}\sum_{a,b} \hat{\omega}^a  \vSigma \hat{\omega}^b \phi^a\cdot \phi^b} e^{-\frac{n}{2}\log \det(\vI_d+\vG_{Z})}e^{-\frac{n}{2}\log \det (\vI_d+\vG_{\vOmega})}.
\end{align}

\paragraph*{Averaging over the $\hat{\omega}^a$.}
We can integrate with respect to $\hat{\omega}$. The only terms that appear with it are
\begin{align}
    \int \prod_a \dd\hat{\omega}^a e^{i\sqrt{p} \sum_a\hat{\omega}^a\cdot \omega^a} e^{-\frac{1}{2}\sum_{a,b} \hat{\omega}^a \vSigma \hat{\omega}^b \phi^a\cdot \phi^b}.
\end{align}
Denote $\vJ_i^a=i\sqrt{p}\omega_i^a$ and $\vG_{kl}^{ab}=\vSigma_{kl} \ \phi^a\cdot\phi^b$, then the integral is of the form
\begin{align}
    \int \prod_a \dd\hat{\omega}^a e^{\sum_{i,a}\vJ_i^a \hat{\omega}_i^a} e^{-\frac{1}{2}\sum_{i,j,a,b} \hat{\omega}_i^a \vG_{ij}^{ab}\hat{\omega}_j^b}
    =e^{-\frac{1}{2}\log \det(\vG)+\frac{1}{2}\vJ^T\vG^{-1}\vJ}.
\end{align}
This gives 

\begin{align}
    \mathbb{E}_{\vW,\vX,\vOmega}[\mathcal{Z}^s]&=\int\prod_{a=1}^s \dd\phi^a\dd\omega^a    e^{\frac{1}{2}(z-s_t^2)\phi^a {\phi^a}^T}e^{-\frac{b_t^2p}{2d}{\omega^a}^T\vSigma^{-1}\omega^a}e^{-\frac{n}{2}\log \det(\vI_d+\vG_{Z})}\nonumber \\
& e^{-\frac{n}{2}\log \det (\vI_d+\vG_{\vOmega})}e^{-\frac{1}{2}\log \det(\vG)+\frac{1}{2}\vJ^T\vG^{-1}\vJ}.
\end{align}
\paragraph*{Introducing the order parameters.}
We define the order parameters as $\vQ^{ab} = \frac{1}{p} \phi^a \cdot \phi^b$ and $\vR^{ab} = \frac{1}{d} \omega^a \cdot \omega^b$. To enforce these constraints, we use the following delta function representations 
\begin{align}
    &1=\int \dd\vQ^{ab}\dd\hat{\vQ}^{ab} e^{\frac{1}{2}\hat{\vQ}^{ab}(p\vQ^{ab}-\phi^a \cdot \phi^b)},\\
    &1=\int \dd\vR^{ab}\dd\hat{\vR}^{ab} e^{\frac{1}{2}\hat{\vR}^{ab}(d\vR^{ab}-\omega^a \cdot \omega^b)},
\end{align}
\begin{align}
    \mathbb{E}_{\vW,Y,\vOmega}[\mathcal{Z}^s]&=\int\prod_{a=1}^s \dd\phi^a\dd\omega^a\dd\vQ^{ab}\dd\hat{\vQ}^{ab}    \dd\vR^{ab}\dd\hat{\vR}^{ab}\nonumber \\
    &e^{\frac{1}{2}\hat{\vQ}^{ab}(p\vQ^{ab}-\phi^a \cdot \phi^b)}e^{\frac{1}{2}\hat{\vR}^{ab}(d\vR^{ab}-\omega^a \cdot \omega^b)}\nonumber \\
    &e^{\frac{p}{2}(z-s_t^2)\Tr \vQ}e^{-\frac{n}{2}\log \det (\vI_m+\frac{a_t^2e^{-2g}p}{n}\vR)}e^{-\frac{b_t^2p}{2d}{\omega^a}^T\vSigma^{-1}\omega^a}\nonumber \\
& e^{-\frac{n}{2}\log(1+\frac{p}{n}(v_t^2-\frac{a_t^2e^{-2t}v_t^2}{n}\vR(1+\frac{a_t^2e^{-2t}p}{n}\vR)^{-1})\vQ)}\nonumber \\
&e^{-\frac{1}{2}\log \det(\vSigma \otimes \vQ)}e^{-\frac{1}{2}\omega_k^a\vSigma^{-1}_{kl}(\vQ^{-1})_{ab}\omega_l^b}.
\end{align}
We also introduce $\vS^{ab}=\omega_k^a \vSigma^{-1}\omega_l^b/d$.
\begin{align}
    \mathbb{E}_{\vW,\vX,\vOmega}[\mathcal{Z}^s]&=\int\prod_{a=1}^s \dd\phi^a\dd\omega^a\dd\vQ^{ab}\dd\hat{\vQ}^{ab}    \dd\vR^{ab}\dd\hat{\vR}^{ab}\dd \vS^{ab}\dd \hat{\vS}^{ab}\nonumber \\
    &e^{\frac{1}{2}\hat{\vQ}^{ab}(p\vQ^{ab}-\phi^a \cdot \phi^b)}e^{\frac{1}{2}\hat{\vR}^{ab}(d\vR^{ab}-\omega^a \cdot \omega^b)}e^{\frac{1}{2}\hat{\vS}^{ab}(dS^{ab}-\omega^a\vSigma^{-1}\omega^b)}\nonumber \\
    &e^{\frac{p}{2}(z-s_t^2)\Tr \vQ}e^{-\frac{n}{2}\log \det (\vI_m+\frac{a_t^2e^{-2t}p}{n}\vR)}e^{-\frac{b_t^2 p}{2}\Tr(\vS)}\nonumber \\
& e^{-\frac{n}{2}\log(1+\frac{p}{n}(v_t^2-\frac{a_t^2e^{-2t}v_t^2}{n}\vR(1+\frac{a_t^2v_t^2p}{n}\vR)^{-1})\vQ)}\nonumber \\
&e^{-\frac{1}{2}\log \det(\vSigma \otimes \vQ)}e^{-\frac{d}{2}\Tr (\vS\vQ^{-1})}.
\end{align}
The integration over $\dd \phi^a$ and $\dd \omega^a$ gives
\begin{align}
    \mathbb{E}_{\vW,\vX,\vOmega}[\mathcal{Z}^s]&=\int\prod_{a=1}^s \dd\vQ^{ab}\dd\hat{\vQ}^{ab}    \dd\vR^{ab}\dd\hat{\vR}^{ab}\dd \vS^{ab}\dd \hat{\vS}^{ab}\nonumber \\
    &e^{\frac{p}{2}\Tr(\hat{\vQ}\vQ)}e^{-\frac{p}{2}\log \det \hat{\vQ}}e^{\frac{d}{2}\hat{\vR}^{ab}\vR^{ab}}e^{\frac{d}{2}\hat{\vS}^{ab}\vS^{ab}}\nonumber \\
    &e^{-\frac{1}{2}\log\det(\hat{\vR}\otimes\vI_d+\hat{\vS}\otimes\vSigma^{-1})}\nonumber \\
    &e^{\frac{p}{2}(z-s_t^2)\Tr \vQ}e^{-\frac{n}{2}\log \det (\vI_m+\frac{a_t^2e^{-2t}p}{n}\vR)}e^{-\frac{b_t^2 p}{2}\Tr(\vS)}\nonumber \\
& e^{-\frac{n}{2}\log(1+\frac{p}{n}(v_t^2-\frac{a_t^2e^{-2t}v_t^2}{n}\vR(1+\frac{a_t^2e^{-2t}p}{n}\vR)^{-1})\vQ)}\nonumber \\
&e^{-\frac{1}{2}\log \det(\vSigma \otimes \vQ)}e^{-\frac{d}{2}\Tr (\vS\vQ^{-1})}.
\end{align}
We need to combine $e^{-\frac{1}{2}\log \det(\vSigma \otimes \vQ)}$ and $e^{-\frac{1}{2}\log\det(\hat{\vR}\otimes\vI_d+\hat{\vS}\otimes\vSigma^{-1})}$,
\begin{align}
    e^{-\frac{1}{2}\log \det(\vSigma \otimes \vQ)}e^{-\frac{1}{2}\log\det(\hat{\vR}\otimes\vI_d+\hat{\vS}\otimes\vSigma^{-1})}
    &=e^{-\frac{1}{2}\log \det(\vQ\hat{\vS}\otimes\vI_d+\vQ\hat{\vR}\otimes\vSigma)}\\
    &=e^{-\frac{d}{2}\log \det(\vQ\hat{\vS})}e^{-\frac{1}{2}\log \det(\vI_m\otimes\vI_d+\hat{\vR}\hat{\vS}^{-1}\otimes\vSigma)}
\end{align}
Then for $e^{-\frac{1}{2}\log \det(\vI_m\otimes\vI_d+\hat{\vR}\hat{\vS}^{-1}\otimes\vSigma)}$, we can introduce $\rho_{\vSigma}(\lambda)$ the density of eigenvalues of $\vSigma$
\begin{align}
    -\frac{1}{2}\log \det(\vI_m\otimes\vI_d+\hat{\vR}\hat{\vS}^{-1}\otimes\vSigma)=
    &-\frac{1}{2}\Tr \log(\vI_m\otimes\vI_d+\hat{\vR}\hat{\vS}^{-1}\otimes\vSigma)\\
    &=-\frac{1}{2}\sum_{l\ge0}\frac{(-1)^l}{l!}(\hat{\vR}\hat{\vS}^{-1})^l\otimes \vSigma^l\\
    &=-\frac{d}{2}\int\dd\lambda\rho_{\vSigma}(\lambda)\sum_{l\ge0}\frac{(-1)^l}{l!}\Tr((\hat{\vR}\hat{\vS}^{-1})^l)\lambda^l\\
    &=-\frac{d}{2}\int\dd\lambda\rho_{\vSigma}(\lambda)\Tr\log(\vI_m\otimes\vI_d+\lambda\hat{\vR}\hat{\vS}^{-1}).
\end{align}
We end up with
\begin{align}
    \mathbb{E}_{\vW,\vX,\vOmega}[\mathcal{Z}^m]=\int \dd \vQ \dd\hat{\vQ} \dd\vR\dd\hat{\vR} \dd \vS \dd\hat{\vS}e^{-\frac{d}{2}S(\vQ,\hat{\vQ}, \vR, \hat{\vR},\vS, \hat{\vS})},
\end{align}
where the action reads
\begin{align}
    S(\vQ,\hat{\vQ}, \vR, \hat{\vR},\vS, \hat{\vS})&=\psi_p \log \det\hat{\vQ}-\psi_p\Tr(\vQ\hat{\vQ})-\Tr(\vR\hat{\vR})-\Tr(\vS\hat{\vS})\nonumber \\
    &-\psi_p(z-s_t^2)\Tr\vQ+\psi_n\log \det(\vI_s+\frac{a_t^2e^{-2t}p}{n}\vR)+b_t^2\psi_p\Tr\vS\nonumber \\
    &+\psi_n \log(\vI_s+\frac{p}{n}(v_t^2-\frac{a_t^2e^{-2t}v_t^2}{n}\vR(\vI_s+\frac{a_t^2e^{-2t}p}{n}\vR)^{-1})\vQ)\nonumber \\
    &+\log \det(\vQ\hat{\vS})+\int\dd\lambda\rho_{\vSigma}(\lambda)\Tr\log(\vI_m\otimes\vI_d+\lambda\hat{\vR}\hat{\vS}^{-1})
    +\Tr(\vS\vQ^{-1}).
\end{align}
In the high dimensional limit, the partition function is dominated by the saddle point. By derivating with respect to $\hat{\vQ}$ we get
\begin{align}
    \hat{\vQ}^{-1}=\vQ,
\end{align}
which yields
\begin{align}
    S(\vQ, \vR, \hat{\vR},\vS, \hat{\vS})&=-\psi_p \log \det\vQ-\Tr(\vR\hat{\vR})-\Tr(\vS\hat{\vS})\nonumber \\
    &-\psi_p(z-s_t^2)\Tr\vQ+\psi_n\log \det(\vI_s+\frac{a_t^2e^{-2t}p}{n}\vR)+b_t^2\psi_p\Tr \vS\nonumber \\
    &+\psi_n \log(\vI_s+\frac{p}{n}(v_t^2-\frac{a_t^2e^{-2t}v_t^2}{n}\vR(\vI_s+\frac{a_t^2e^{-2t}p}{n}\vR)^{-1})\vQ)\nonumber \\
    &+\log \det(\vQ\hat{\vS})+\int\dd\lambda\rho_{\vSigma}(\lambda)\Tr\log(\vI_m\otimes\vI_d+\lambda\hat{\vR}\hat{\vS}^{-1})\nonumber \\
    &+\Tr(\vS\vQ^{-1}).
\end{align}
As a sanity check, if $\vSigma=\vI_d$, differentiation with respect to $\hat{\vR}$ and $\hat{\vS}$ yields
\begin{align}
    \vR=\vS=(\hat{\vS}+\hat{\vR})^{-1},
\end{align}
and we find back the same action as before. 
\paragraph*{RS Ansatz.} As before we introduce a RS ansatz for all the the matrices and moreover suppose that only the diagonal terms are non vanishing i.e. they are of the form $\vQ=q\vI_s$. This ansatz yields
\begin{align}
    S(q,r,\hat{r}, s,\hat{s})/s&=-\psi_p \log q-r\hat{r}-s\hat{s}\nonumber \\
    &-\psi_p(z-s_t^2)q+\psi_n\log(1+\frac{a_t^2e^{-2t}p}{n}r+\frac{pv_t^2}{n}q)+b_t^2\psi_p s\nonumber \\
    &+\log (q)+\int\dd\lambda\; \rho_{\vSigma}(\lambda)\log(\hat{s}+\lambda\hat{r})+\frac{s}{q}.
\end{align}
Let us differentiate with respect to the 5 variables
\begin{align}
    &\frac{\partial S}{\partial s }=-\hat{s}+b_t^2\psi_p+\frac{1}{q},\\
    &\frac{\partial S}{\partial r }=-\hat{r}+\frac{\psi_p a_t^2e^{-2t}}{1+\frac{a_t^2e^{-2t}p}{n}r+\frac{pv_t^2}{n}q},\\
    &\frac{\partial S}{\partial \hat{s}}=-s+\int\dd\lambda \rho_{\vSigma}(\lambda)\frac{1}{\hat{s}+\lambda\hat{r}},\\
    &\frac{\partial S}{\partial \hat{r}}=-r+\int\dd\lambda \rho_{\vSigma}(\lambda)\frac{\lambda}{\hat{s}+\lambda\hat{r}},\\
    &\frac{\partial S}{\partial q}=-\frac{\psi_p}{q}-\psi_p(z-s_t^2)+\frac{\psi_pv_t^2}{1+\frac{a_t^2e^{-2t}p}{n}r+\frac{pv_t^2}{n}q}+\frac{1}{q}-\frac{s}{q^2}.
\end{align}
Hence the saddle point equations read
\begin{align}
        &\hat{s}=b_t^2\psi_p+\frac{1}{q},\\
    &\hat{r}=\frac{\psi_p a_t^2e^{-2t}}{1+\frac{a_t^2e^{-2t}p}{n}r+\frac{pv_t^2}{n}q},\\
    &s=\int\dd \rho_{\vSigma}(\lambda)\frac{1}{\hat{s}+\lambda\hat{r}},\\
    &r=\int\dd \rho_{\vSigma}(\lambda)\frac{\lambda}{\hat{s}+\lambda\hat{r}},\\
    &\psi_p(s_t^2-z)+\frac{\psi_pv_t^2}{1+\frac{a_t^2e^{-2t}p}{n}r+\frac{pv_t^2}{n}q}+\frac{1-\psi_p}{q}-\frac{s}{q^2}=0.
\end{align}
Finally, we observe that the solution $q^*$ to the saddle point equations corresponds to the Stieltjes transform of $\rho$.
\begin{align}
    2\partial_z\frac{1}{p}\frac{\mathbb{E}[\mathcal{Z}^s]-1}{s}=2\partial_z\frac{1}{p}\frac{e^{-\frac{d}{2}S(q^*,r^*)}-1}{m}\underset{m\rightarrow0}{\rightarrow}-2\partial_z\frac{1}{p}\frac{d}{2}S(q^*,r^*)=q^*.
\end{align}
\end{proof}

\subsection{Proof of Theorem~\ref{thm:Spectrum_new}} 
We recall Theorem~\ref{thm:Spectrum_new} of the MT.
\begin{thm}[Informal]
\label{thm:Spectrum_new2}
Let $\rho$ denote the spectral density of $\vU$.
\begin{enumerate}[leftmargin=*,itemsep=2pt]
    \item[] \textbf{Regime I (overparametrized): $\psi_p>\psi_n\gg 1$.}
    \[
        \rho(\lambda)=\Bigl(1-\frac{1+\psi_n}{\psi_p}\Bigr)\delta(\lambda-{s_t^2})
                     +\frac{\psi_n}{\psi_p}\,\rho_1(\lambda)
                     +\frac{1}{\psi_p}\,\rho_2(\lambda).
    \]
    \item[] \textbf{Regime II (underparametrized): $\psi_n>\psi_p\gg 1$.}
    \[
        \rho(\lambda)=\Bigl(1-\frac{1}{\psi_p}\Bigr)\rho_1(\lambda)
                     +\frac{1}{\psi_p}\,\rho_2(\lambda).
    \]
\end{enumerate}

where $\rho_1$ is a atomless measure with support 
\[
    \left[s_t^2 + v_t^2\left(1-\sqrt{\psi_p/\psi_n}\right)^{2},\;
          s_t^2 + v_t^2\left(1+\sqrt{\psi_p/\psi_n}\right)^{2}\right],
\] 
and $\rho_2$ coincides with the asymptotic eigenvalue bulk density of  the population covariance $\tilde{\vU}=\mathbb{E}_{\vX}[\vU]$; $\rho_2$ is independent of $\psi_n$ and its support is on the scale $\psi_p$.
The eigenvectors associated with $\delta(\lambda-{s_t^2})$ leave both training and test losses unchanged and are therefore irrelevant. In the limit $\psi_p\gg \psi_n$, the supports of $\rho_1$ and $\rho_2$ are respectively on the scales $\psi_p/\psi_n$ and $\psi_p$, i.e. they are well separated.

\end{thm}

We now proceed to prove Theorem~\ref{thm:Spectrum_new}.
\begin{proof}
\textbf{Delta peak.} We first account for the delta peak in the spectrum. We use the Gaussian equivalence for $\vU$ computed in Lemma~\ref{lem:GEP_U}. Let $\vOmega^\nu\in\mathbb{R}^p$ be the $\nu$th column of $\vOmega$ and $\vW_i\in\mathbb{R}^p$ the $i$th row of $\vW$.  Suppose a vector $\vv\in\mathbb{R}^p$ lies in the kernel of all these
\begin{align}
    &\forall \nu=1,\dots,n,\quad\sum_{i=1}^p\vOmega^\nu_i \vv_i=0,\\
     &\forall k=1,\ldots,d,\quad\sum_{k=1}^p\vW_{ik}\vv_i=0.
\end{align}
then $\vU\vv=s_t^2\vv$. These are $n+d$ linear constraints on a vector of size $p$ hence there are non trivial solutions for $n+d\le p$. Hence a delta‐peak at $s_t^2$ appears as soon as 
$\psi_p  \ge  \psi_n+1 $.  Next, we extract its weight.  Recall that the Stieltjes transform satisfies
\[
q(z) \;=\;\int \frac{\rho(\lambda)}{\lambda - z}\,\dd\lambda,
\]
and a point mass of weight \(f\) at \(\lambda = s_t^2\) contributes
\(\tfrac{-f}{z - s_t^2}\approx \tfrac{f}{\varepsilon}\)
as \(z \to s_t^2 - \varepsilon\).  Meanwhile
\[
s(z)
\;=\;\frac1p \Tr\!\bigl[\vW^T(\vU - z \vI)^{-1}\vW\bigr],\quad r(z)
\;=\;\frac1p \Tr\!\bigl[{\vSigma}^{1/2}\vW^T(\vU - z \vI)^{-1}\vW{\vSigma}^{1/2}\bigr]
\]
remain finite in that limit, since the corresponding eigenvectors satisfy \(\vW\,\vv=0\).  We substitute this Ansatz into the equations of Theorem~\ref{thm:Saddle_point_equations_new}. The first equation reads
\begin{align}
    \psi_n\frac{\frac{pv_t^2}{n}}{1+\frac{e^{-2t}\mu_1^2p\sigma_{\vx}^2}{n}r+\frac{p}{n}v_t^2q}+\psi_p(s_t^2-z)+\frac{1-\psi_p}{q}-\frac{s}{q^2}=0,
\end{align}
and simplifies to
\begin{align}
    \frac{\psi_n\varepsilon}{f}+\psi_p\varepsilon+\frac{(1-\psi_p)\varepsilon}{f}=0.
\end{align}
It readily gives
\begin{align}
    f=1-\frac{1}{\psi_p}-\frac{\psi_n}{\psi_p}.
\end{align}
Thus the point mass at \(s_t^2\) has weight $1-\frac{1}{\psi_p}-\frac{\psi_n}{\psi_p}$, in agreement with the counting of degrees of freedom presented above.\\
Finally, one checks that these isolated eigenvalues do not contribute to the train and test losses. After expanding the square they read
\begin{align}
    &\mathcal{L}_\mathrm{train}(\vA)=1+\frac{\Delta_t}{d} \Tr(\frac{\vA^T}{\sqrt{p}}\frac{\vA}{\sqrt{p}}\vU)+\frac{2\sqrt{\Delta_t}}{d}\Tr(\frac{\vA}{\sqrt{p}}\vV)\\
    &\mathcal{L}_\mathrm{test}(\vA)=1+\frac{\Delta_t}{d} \Tr(\frac{\vA^T}{\sqrt{p}}\frac{\vA}{\sqrt{p}}\tilde{\vU})+\frac{2\sqrt{\Delta_t}}{d}\Tr(\frac{\vA}{\sqrt{p}}\tilde{\vV}) 
\end{align}
The terms that appear in the loss are of the form $\Tr(\vA^T\vA...)$ and $\Tr(\vA\vW)$. The trace can be decomposed on the basis of eigenvectors of $\vU$. The eigenvectors associated with the delta peak satisfy $\vW^T\vv=0$. Looking at the expression of the matrix $\vA=\vW^T...+\vA_0$, one can easily see that, for initial conditions $\vA_0=0$, one has $\vv^T\vA^T=0$ and the subspace corresponding to these isolated eigenvalues does not contribute to the loss.\\
\textbf{First bulk.} Using the expression for $q=\frac{1}{p}\Tr\frac{1}{\vU-z\vI_p}$ and $r(z)=\frac{1}{p}\Tr(\vSigma^{1/2}\vW^T(\vU-z\vI)^{-1}\vW\vSigma^{1/2})$ we make the following Ansatz in the large $\psi_p$ limit:
\begin{align}
    q=\mathcal{O}_{\psi_p}(1), \quad r=\mathcal{O}_{\psi_p}(\frac{1}{\psi_p}).
\end{align}
In this limit the saddle point equations becomes at leading order in $\psi_p$
\begin{align}
    &\hat{s}=b_t^2\psi_p\\
    &\hat{r}=\frac{\psi_pa_t^2e^{-2t}}{1+\frac{v_t^2p}{n}r}\\
    &s=\mathcal{O}(1/\psi_p)\\
    &r=\mathcal{O}(1/\psi_p)\\
    &(s_t^2-z)+\frac{ v_t^2}{1+\frac{pv_t^2}{n}q}-\frac{1}{q}=0.
\end{align}
We can focus only on the last equation on $q$ only. This is a quadratic polynomial in $q$. If its discriminant is negative then the solutions are imaginary and thus the density of eigenvalues is non-zero. The edge of the bulk are where the discriminant vanishes
\begin{align}
    \Delta=(s_t^2-\lambda(1-\frac{p}{n})v_t^2)^2+4(s_t^2-\lambda)\frac{p}{n}v_t^2=0.
\end{align}
It vanishes for 
\begin{align}
    \lambda_{\pm}=s_t^2+v_t^2\left(1\pm\sqrt{\frac{p}{n}}\right)^2
\end{align}
which are the edges of the first bulk $\rho_1$. We have checked this result, and hence validated the Ansatz  solving numerically the equations on $r,q$. Interestingly at leading order the expression of the first bulk is independent of $\rho_{\vSigma}$.\\
\textbf{Second Bulk.} We scale $q=\mathcal{O}_{\psi_p}(1/\psi_p)$ and $r=\mathcal{O}_{\psi_p}(1/\psi_p)$. The equations on $\hat{s}$ and $\hat{r}$ lead to 
\begin{align}
    &\hat{s}=\psi_pb_t^2+\frac{1}{q}\\
    &\hat{r}=\psi_pa_t^2e^{-2t}.
\end{align}
This yields the following equation on $q$
\begin{align}
    \psi_p(s_t^2-z)+\psi_pv_t^2+\frac{1-\psi_p}{q}-\frac{1}{q}\int\frac{\dd\rho_{\vSigma}(\lambda)}{1+q\psi_p(b_t^2+\lambda a_t^2e^{-2t})}=0.
\end{align}
We denote the shifted variable $z'=z-s_t^2-v_t^2$. This yields
\begin{align}
     -\psi_pz'+\frac{1-\psi_p}{q}-\frac{1}{q}\int\frac{\dd\rho_{\vSigma}(\lambda)}{1+q\psi_p(b_t^2+\lambda a_t^2e^{-2t})}=0.
\end{align}
We decompose the integral
\begin{align}
    \int\frac{\dd\rho_{\vSigma}(\lambda)}{1+q\psi_p(b_t^2+\lambda a_t^2e^{-2t})}&=\int\frac{\dd\rho_{\vSigma}(\lambda)(1+q\psi_p(b_t^2+\lambda a_t^2e^{-2t})-q\psi_p(b_t^2+\lambda a_t^2e^{-2t}))}{1+q\psi_p(b_t^2+\lambda a_t^2e^{-2t})}\\
    &=1-q\psi_p\int\frac{\dd\rho_{\vSigma}(\lambda)(b_t^2+\lambda a_t^2e^{-2t})}{1+q\psi_p(b_t^2+\lambda a_t^2e^{-2t})}
\end{align}
By plugging this back in the equation we find
\begin{align}
    q=-\left(z'-\int \frac{\dd \rho_{\vSigma}(\lambda)(b_t^2+\lambda a_t^2e^{-2t})}{1+\psi_p q(b_t^2+\lambda a_t^2e^{-2t})}\right)^{-1}.
\end{align}
We do the change of variable $\mu=b_t^2+\lambda a_t^2e^{-2t}$. This yields
\begin{align}
    q=-\left(z'-\frac{1}{a_t^2e^{-2t}}\int \frac{\dd\mu \rho_{\vSigma}(\frac{\mu-b_t^2}{a_t^2e^{-2t}})\mu}{1+\psi_p q\mu}\right)^{-1}.
\end{align}
An integration by parts give that $b_t^2=\Delta_t\mu_1^2(t)$ $a_t^2=\mu_1^2(t)/\sigma_{\vx}^2$. We thus realize that the integral is over the eigenvalue distribution of $\mu_1^2(t)(e^{-2t}\vSigma+\Delta_t\vI_d)$,
\begin{align}
    q=-\left(z'-\int \frac{\dd\mu \rho_{\mu_1^2(t)\vSigma_t}(\mu)\mu}{1+\psi_p q\mu}\right)^{-1}.
\end{align}
We recognize the Bai-Silverstein equations \cite{SILVERSTEIN1995, bai2008} for the eigenvalue density of the matrix 
\begin{align}
    \tilde{\vU}=\mu_1^2(t)\frac{\vW\vSigma_t\vW^T}{d}+(s_t^2+v_t^2)\vI_p=\mathbb{E}_{\vx}[\vU]
\end{align}
which is the population version of $\vU$ and is thus independent of $n$. Lemma~\ref{lem:bulk_U_tilde} concludes on the order of the eigenvalues in the bulk of $\rho_2$.\\

\end{proof}

\subsection{Dynamics on the fast timescales}

In the following we denote for a matrix $\vA\in \mathbb{R}^{p\times p}$,
\begin{align}
\lVert \vA\rVert_{\mathrm{op}}=\underset{\vv\in\mathbb{R}^p,\lVert \vv\rVert=1}{\sup}\lVert\vA\vv\rVert
\end{align}
the operator norm and 
\begin{align}
\lVert\vA\rVert_\mathrm{F}=(\sum_{i,j=1}^p\vA_{ij}^2)^{1/2}
\end{align}
the Frobenius norm. Before deriving the fast‐time behavior, we need the following lemma.
\begin{lemma}
\label{lem:Eigenvectors_U}
The operator norm of $\vU-\tilde{\vU}$ satisfies 
\begin{align}
     \lVert (\vU-\tilde{\vU})\rVert_{\mathrm{op}}=\mathcal{O}(\frac{\psi_p}{\sqrt{\psi_n}}),
\end{align}
when $p\gg n\gg d$.
\end{lemma}
\begin{proof}
     On the one hand,
    \begin{align}
        \vU=e^{-2t}a_t^2\frac{\vW\vX\vX^T\vW^T}{d}+v_t^2\frac{\vOmega\vOmega^T}{n}+\frac{e^{-t}a_tv_t}{n\sqrt{d}}\left(\vW \vX\vOmega^T+\vOmega\vX^T\vW^T\right)+(s_t^2+v_t^2)\vI_p
    \end{align}
    and on the other hand,
    \begin{align}
        \tilde{\vU}=\mu_1^2e^{-2t}\frac{\vW\vSigma\vW^T}{d}+\Delta_t\mu_1^2\frac{\vW\vW^T}{d}+(s_t^2+v_t^2)\vI_p.
    \end{align}
    We also note the identities $b_t^2=\Delta_t\mu_1^2(t)$ and $a_t^2=\mu_1^2(t)$.
    \begin{align}
    \vU-\tilde{\vU}=a_t^2e^{-2t}\frac{\vW}{\sqrt{d}}(\frac{\vX\vX^T}{n}-\vSigma)\frac{\vW^T}{\sqrt{d}}+v_t^2(\frac{\vOmega\vOmega^T}{n}-\vI_p)+\frac{a_tv_te^{-t}}{n\sqrt{d}}(\vOmega \vX^T\vW^T+\vW \vX\vOmega^T).
\end{align}
We can bound its operator norm
\begin{align}
    \lVert (\vU-\tilde{\vU})&\rVert_{\mathrm{op}}\le C_1 \lVert \frac{\vW}{\sqrt{d}}(\frac{\vX\vX^T}{n}-\vSigma)\frac{\vW^T}{\sqrt{d}}\rVert_{\mathrm{op}}+C_2\lVert (\frac{\vOmega\vOmega^T}{n}-\vI_p)\rVert_{\mathrm{op}} \nonumber \\
    &\qquad+\frac{C_3}{n\sqrt{d}}\lVert\vOmega \vX^T\vW^T+\vW \vX\vOmega^T\rVert_{\mathrm{op}},
\end{align}
where $C_1,C_2,C_3$ are constants independent of $p,n,d$. We bound each of the three terms on the right hand side. We will use the fact that for a symmetric matrix, the operator norm $\lVert .\rVert_{\mathrm{op}}$is equal to its largest eigenvalue. 

\paragraph*{First term.} 
\begin{align}
    \lVert \frac{\vW}{\sqrt{d}}(\frac{\vX\vX^T}{n}-\vSigma)\frac{\vW^T}{\sqrt{d}}\rVert_{\mathrm{op}}.
\end{align}
We observe that $\frac{\vW}{\sqrt{d}}(\frac{\vX\vX^T}{n}-\vSigma)\frac{\vW^T}{\sqrt{d}}$ and $\frac{\vW^T}{\sqrt{d}}\frac{\vW}{\sqrt{d}}(\frac{\vX\vX^T}{n}-\vSigma)$ have the same eigenvalues up to the multiplicity of 0\footnote{They both have the same moments $\Tr(.)^k$ owing to the cyclicity of the trace.}. We then use the sub-multiplicativity of the operator norm
\begin{align}
    \lVert \frac{\vW}{\sqrt{d}}(\frac{\vX\vX^T}{n}-\vSigma)\frac{\vW^T}{\sqrt{d}}\rVert_{\mathrm{op}}\le\lVert \frac{\vW^T}{\sqrt{d}} \frac{\vW}{\sqrt{d}}\rVert_{\mathrm{op}}\lVert(\frac{\vX\vX^T}{n}-\vSigma)\rVert_{\mathrm{op}}.
\end{align}
We can do the same operation by introducing $\vX=\vSigma\vZ$ with $\vZ\in\mathbb{R}^{d\times n}$ with standard Gaussian entries,
\begin{align}
    \lVert(\frac{\vX\vX^T}{n}-\vSigma)\rVert_{\mathrm{op}}=\lVert\vSigma^{1/2}(\frac{\vZ\vZ^T}{n}-\vI_d)\vSigma^{1/2}\rVert_{\mathrm{op}}\le\lVert(\frac{\vZ\vZ^T}{n}-\vI_d)\rVert_{\mathrm{op}}\lVert\vSigma\rVert_{\mathrm{op}}.
\end{align}
Among our assumptions, we had $\lVert\vSigma\rVert_{\mathrm{op}}<\mathcal{O}(1)$. The spectrum of $(\frac{\vX\vX^T}{n}-\vI_d)$ is the Marchenko-Pastur law whose largest eigenvalue is of order $\sqrt{d/n}$ while for $\frac{\vW^T\vW}{d}$ it is order $\frac{p}{d}$. The bound reads
\begin{align}
    \lVert \frac{\vW}{\sqrt{d}}(\frac{\vX\vX^T}{n}-\vSigma)\frac{\vW^T}{\sqrt{d}}\rVert_{\mathrm{op}}\le\mathcal{O}(\frac{p}{\sqrt{nd}}).
\end{align}

\paragraph*{Second term.}
\begin{align}
\lVert (\frac{\vOmega\vOmega^T}{n}-\vI_p)\rVert_{\mathrm{op}}.
\end{align}
We observe that the spectrum of $\vOmega\vOmega^T/n-\vI_p$ is Marchenko-Pastur and thus its largest eigenvalue is order $\mathcal{O}(p/n)$ yielding
\begin{align}
    \lVert (\frac{\vOmega\vOmega^T}{n}-\vI_p)\rVert_{\mathrm{op}}\le \mathcal{O}(p/n).
\end{align}
\paragraph*{Third term.}
\begin{align}
    \lVert\vOmega \vX^T\vW^T+\vW \vX\vOmega^T\rVert_{\mathrm{op}}.
\end{align}
We first bound the operator norm by the Frobenius norm.
\begin{align}
    \lVert\vOmega \vX^T\vW^T+\vW \vX\vOmega^T\rVert_{\mathrm{op}}\le 2\lVert\vOmega \vX^T\vW^T\rVert_{\mathrm{F}}.
\end{align}
We expand the square
\begin{align}
    \lVert\vOmega \vX^T\vW^T+\vW \vX\vOmega^T\rVert_{\mathrm{F}}^2=C\sum_{k=1}^d\sum_{i=1}^p(\sum_{\nu=1}^n\vOmega_i^\nu \vX_k^\nu\vW_{kl})^2.
\end{align}
The Central Limit Theorem yields
\begin{align}
    \sum_{\nu=1}^n\vOmega_i^\nu \vX_k^\nu\vW_{kl}=\mathcal{O}(\sqrt{n})\vW_{kl},
\end{align}
hence
\begin{align}
    \frac{1}{n\sqrt{d}}\lVert\vOmega \vX^T\vW^T+\vW \vX\vOmega^T\rVert_{\mathrm{op}}=\mathcal{O}(\frac{\sqrt{ndp}}{n\sqrt{d}})=\mathcal{O}(\sqrt{\frac{p}{n}})
\end{align}
Putting all the contributions together yields
\begin{align}
      \lVert (\vU-\tilde{\vU})\rVert_{\mathrm{op}}&\le \mathcal{O}(\frac{p}{\sqrt{dn}})=\mathcal{O}(\frac{\psi_p}{\sqrt{\psi_n}}).
\end{align}
\end{proof}

\begin{proposition}[Informal]
    On timescales $1\ll \tau\ll \psi_n$, both the train and test losses satisfy
    \begin{align}
        \mathcal{L}_\mathrm{train}\simeq\mathcal{L}_\mathrm{test}\simeq 1-\mathcal{O}(\Delta_t).
    \end{align}
\end{proposition}

\begin{proof}
    
According to the spectral analysis of $\vU$ conducted previously, there are two bulks in the spectrum that contribute to the dynamics: a first bulk with eigenvalues of order $\frac{\psi_p}{\psi_n}$ and a second bulk with eigenvalues of order $\psi_p$ in the $\psi_p, \psi_n\gg1$ limit. Hence, in the regime $1\ll\tau\ll \psi_n$, $e^{-\lambda\frac{\Delta_t\tau}{\psi_p}}\sim0$ if $\lambda$ is in the second bulk and is $e^{-\lambda\frac{\Delta_t\tau}{\psi_p}}\sim1$ if $\lambda$ is in the first bulk. We remind the expressions of the train and test loss
\begin{align}
    &\mathcal{L}_\mathrm{train}(\vA)=1+\frac{\Delta_t}{d} \Tr(\frac{\vA^T}{\sqrt{p}}\frac{\vA}{\sqrt{p}}\vU)+\frac{2\sqrt{\Delta_t}}{d}\Tr(\frac{\vA}{\sqrt{p}}\vV)\\
    &\mathcal{L}_\mathrm{test}(\vA)=1+\frac{\Delta_t}{d} \Tr(\frac{\vA^T}{\sqrt{p}}\frac{\vA}{\sqrt{p}}\tilde{\vU})+\frac{2\sqrt{\Delta_t}}{d}\Tr(\frac{\vA}{\sqrt{p}}\tilde{\vV}) 
\end{align}
and use the expression of $\vA(\tau)$ in Proposition~\ref{prop:A_tau} that we expand on the basis of eigenvectors $\{\vv_\lambda\}_{\lambda\in Sp(\vU)}$of $\vU$.
    \begin{align}
    \frac{\vA(\tau)}{\sqrt{p}}&=\frac{1}{\sqrt{\Delta_t}}\vV^T\vU^{-1}(e^{-\frac{2\Delta_t}{d}\vU \tau}-\vI_p)\\
    &=\frac{1}{\sqrt{\Delta_t}}\vV^T\vU^{-1}\sum_\lambda(e^{-\frac{2\Delta_t}{d}\lambda \tau}-1)\vv_\lambda\vv_\lambda^T\\
    &\sim -\frac{1}{\sqrt{\Delta_t}}\vV^T\vU^{-1}\sum_{\lambda\in \rho_2}\vv_\lambda\vv_\lambda^T,
\end{align}
where $\lambda\in\rho_2$ means that the eigenvalue $\lambda$ belongs to the second bulk. We also have that $\vV$ and $\tilde{\vV}$ have the same GEP $\frac{\mu_1(t)\sqrt{\Delta_t}}{\Gamma_t}\frac{\vW}{\sqrt{d}}$ and they thus cancel each other when computing the generalization loss $\mathcal{L}_{\mathrm{gen}}=\mathcal{L}_{\mathrm{test}}-\mathcal{L}_{\mathrm{train}}$. It reads
\begin{align}
    \mathcal{L}_\mathrm{gen}&=-\frac{\mu_1^2(t)\Delta_t}{\Gamma_t^2d}\Tr(\sum_{\lambda, \lambda'\in\rho_2}\vv_{\lambda'}\vv_{\lambda'}^T\vU^{-1}\frac{\vW\vW^T}{d}\vU^{-1}\vv_{\lambda}\vv_{\lambda}^T(\vU-\tilde{\vU}))\\
    &=-\frac{\mu_1^2\Delta_t}{\Gamma_t^2d}(\sum_{\lambda, \lambda'\in\rho_2}\vv_{\lambda'}^T\vU^{-1}\frac{\vW\vW^T}{d}\vU^{-1}\vv_{\lambda}\vv_{\lambda}^T(\vU-\tilde{\vU})\vv_{\lambda'})\\
    &=-\frac{\mu_1^2\Delta_t}{\Gamma_t^2d}(\sum_{\lambda, \lambda'\in\rho_2}\vv_{\lambda'}^T\frac{1}{\lambda'}\frac{\vW\vW^T}{d}\frac{1}{\lambda}\vv_{\lambda}\vv_{\lambda}^T(\vU-\tilde{\vU})\vv_{\lambda'})\label{eq:L_gen_ww}\\
\end{align}
We then use Lemma~\ref{lem:Eigenvectors_U} --- which states that the operator norm of $\vU-\tilde{\vU}$ in the subspace spanned by the eigenvectors of the second bulk is bounded by $\mathcal{O}(\frac{\psi_p}{\sqrt{\psi_n}})$ --- to bound $\mathcal{L}_{\mathrm{gen}}$,
\begin{align}
    \lvert \mathcal{L}_{\mathrm{gen}} \rvert &\le \lVert \frac{\mu_1^2\Delta_t}{\Gamma_t^2d}(\sum_{\lambda, \lambda'\in\rho_2}\vv_{\lambda'}^T\frac{1}{\lambda'}\frac{\vW\vW^T}{d}\frac{1}{\lambda}\vv_{\lambda}\vv_{\lambda}^T(\vU-\tilde{\vU})\vv_{\lambda'})\rVert_{\mathrm{op}}\\
    &\le \frac{\mu_1^2\Delta_t}{\Gamma_t^2d} d\frac{1}{\psi_p^2} \lVert \frac{\vW\vW^T}{d}\rVert_{\mathrm{op}}\frac{\psi_p}{\sqrt{\psi_n}}\le\mathcal{O}(\frac{d\psi_p^2}{d\psi_p^2\sqrt{\psi_n}})=\mathcal{O}(\frac{1}{\sqrt{\psi_n}}).
\end{align}
We also used the fact that the sums contain $d$ terms --- the only terms that matter are the diagonal ones --- and that the eigenvalues scale as $\psi_p$. The bound yield that $\mathcal{L}_{\mathrm{gen}}$ vanishes asymptotically in the large number of data and large number of parameters regime. Therefore, on the fast timescale we find $\mathcal{L}_\mathrm{train}\simeq \mathcal{L}_\mathrm{test}$. Let us now focus on $\mathcal{L}_\mathrm{train}$

\begin{align}
    \mathcal{L}_\mathrm{train}& = 1+\frac{\mu_1^2\Delta_t}{\Gamma_t^2 d}(\sum_{\lambda, \lambda'\in\rho_2}\vv_{\lambda'}^T\frac{1}{\lambda'}\frac{\vW\vW^T}{d}\frac{1}{\lambda}\vv_{\lambda}\vv_{\lambda}^T\vU\vv_{\lambda'})-\frac{2\Delta_t\mu_1^2}{\Gamma_t^2d}\sum_{\lambda\in\rho_2}\vv_\lambda^T\frac{\vW\vW^T}{d}\vU^{-1}\vv_\lambda\\
    &=1-\frac{\mu_1^2\Delta_t}{\Gamma_t^2d}\sum_{\lambda\in\rho_2}\frac{1}{\lambda}\vv_\lambda^T\frac{\vW\vW^T}{d}\vv_\lambda.
\end{align}
There are $d$ values in the sum and the eigenvalues of $\vU$ and $\frac{\vW\vW^T}{d}$ are both order $\mathcal{O}(\psi_p)$ hence the sum divided by $d$ is a positive $\mathcal{O}(1)$ quantity thus in this training time regime, $1\ll\tau\ll \psi_n$, we obtain: 
\begin{align}
    \mathcal{L}_\mathrm{train}\sim\mathcal{L}_\mathrm{test}=1-\mathcal{O}(\Delta_t).
\end{align}
\end{proof}

\section{Numerical experiments for Random Features}
\label{appendix:Num_exp_RF}
\paragraph*{Details on the numerical experiments.}
All the numerical experiments for the RFNN were conducted using $\sigma=\tanh$ and $\sigma_{\vx}=1$ unless specified. At each step, the gradient of the loss was computed using the full batch of data points. The train loss was estimated by adding noise to each data point $N=100$ times. The test loss was computed by drawing $n$ new points from the data distribution and noising each one $N$ times. The error on the score was evaluated by drawing 10,000 points from the noisy distribution $P_t=\mathcal{N}(0,\Gamma_t^2\vI_d)$.

\paragraph*{Effect of $t$.}
\begin{figure}
    \centering
    \includegraphics[width=0.49\linewidth]{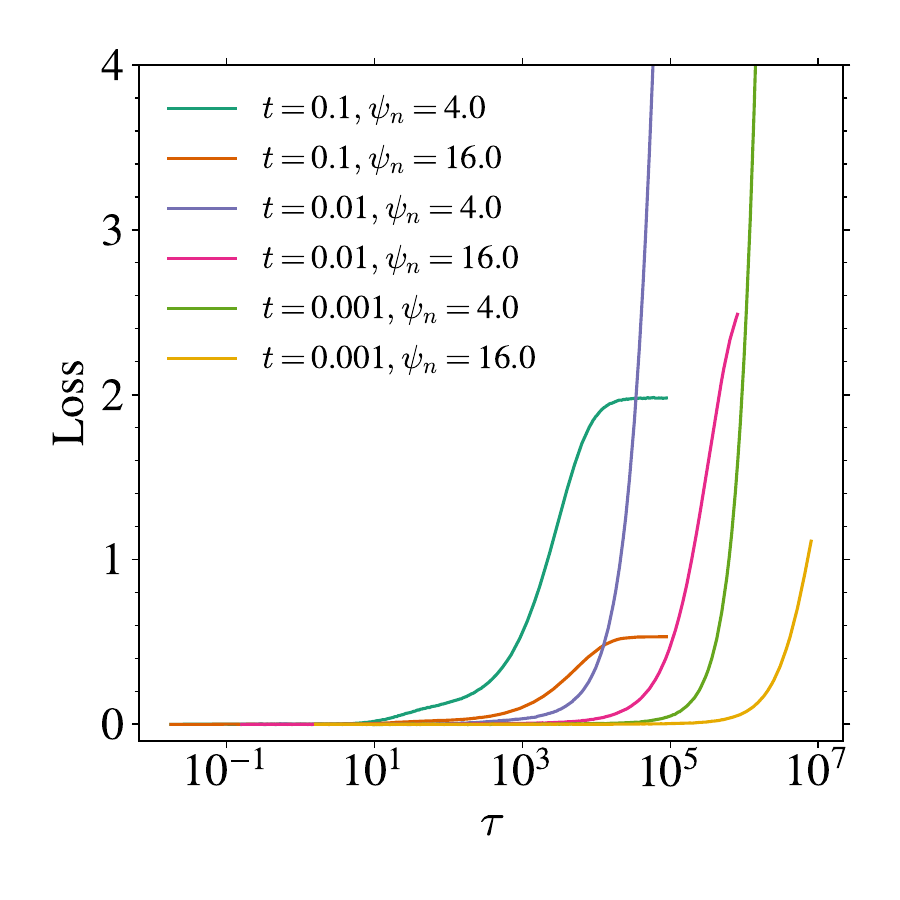}
    \includegraphics[width=0.49\linewidth]{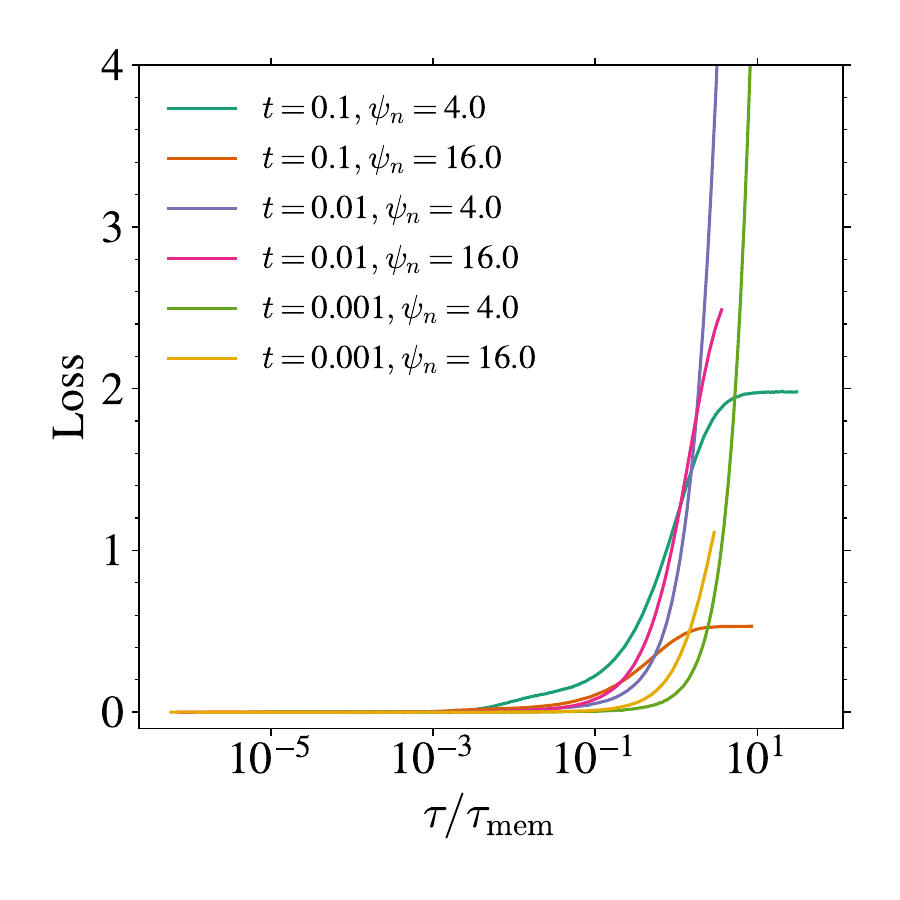}
    \caption{\textbf{Generalization loss for different diffusion times $t$.} Generalization loss $\mathcal{L}_{\mathrm{gen}}$ against (Left) training time $\tau$ and (Right) rescaled training time $\tau/\tau_{\mathrm{gen}}$ for different $\psi_p=32,d=100$ and different $\psi_n$ and $t$.}
    \label{fig:gen_loss_different_t}
\end{figure}
We present plots for different diffusion times $t$ in Fig.~\ref{fig:gen_loss_different_t} and show that the rescaling of the training times $\tau$ by $\tau_{\mathrm{mem}}=\psi_p/\Delta_t\lambda_\mathrm{min}$ also makes the loss curves collapse. Of particular interest is the behavior of \( \tau_{\mathrm{mem}} \), and more specifically the ratio \( \tau_{\mathrm{mem}} / \tau_{\mathrm{gen}} \), at small \( t \). Recall that
\[
\lambda_{\min} = s_t^2 + v_t^2 \left(1 - \sqrt{\frac{\psi_p}{\psi_n}}\right)^2.
\]
In the overparameterized regime \( p \gg n \), this ratio is independent of \( t \) since $v_t^2\sim\mu_*^2$ and $s_t^2\sim{t}$. However, when \( p \sim n \), a nontrivial scaling emerges: since \( \lambda_{\min} \sim s_t^2 \sim t \), it follows that
\[
\frac{\tau_{\mathrm{mem}}}{\tau_{\mathrm{gen}}} \sim \frac{1}{t},
\]
implying that the two timescales become increasingly separated. It is unclear whether this behavior is related to specific properties of the learned score function, and is related to the approach of the interpolation threshold. We leave this question for future investigation.

\begin{figure}[htbp]
  \centering
  \begin{minipage}[b]{0.49\textwidth}
    \centering
    \includegraphics[width=\textwidth]{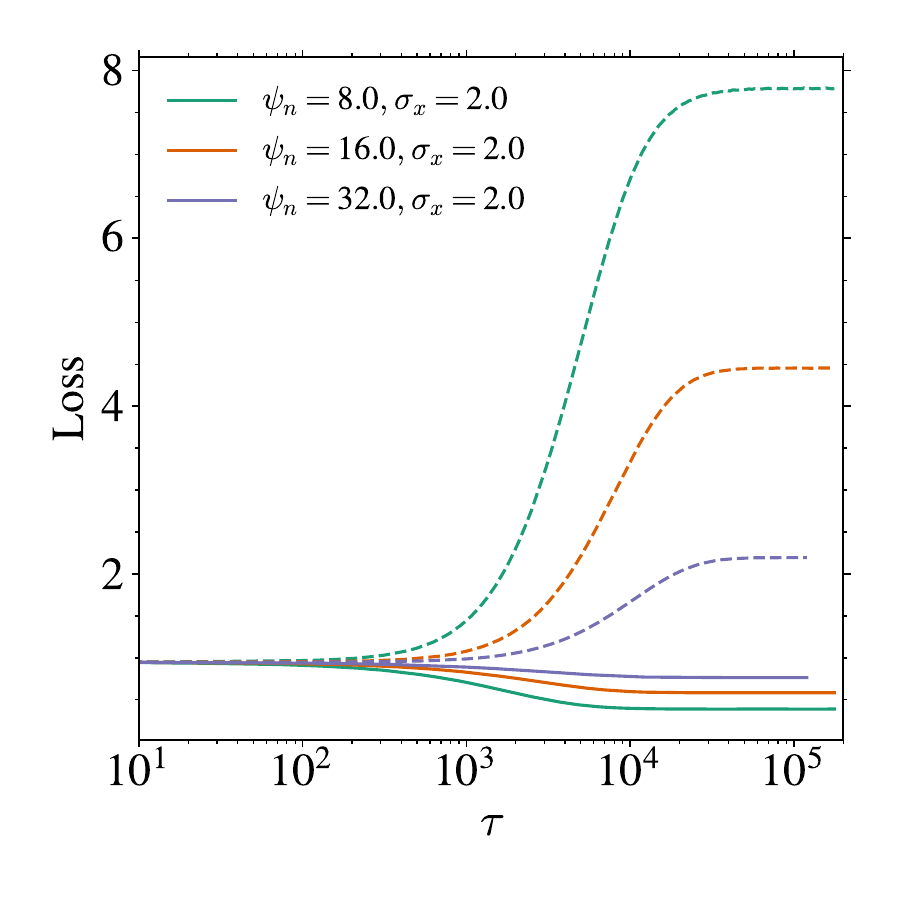}
  
  \end{minipage}
  \hfill
  \begin{minipage}[b]{0.49\textwidth}
    \centering
    \includegraphics[width=\textwidth]{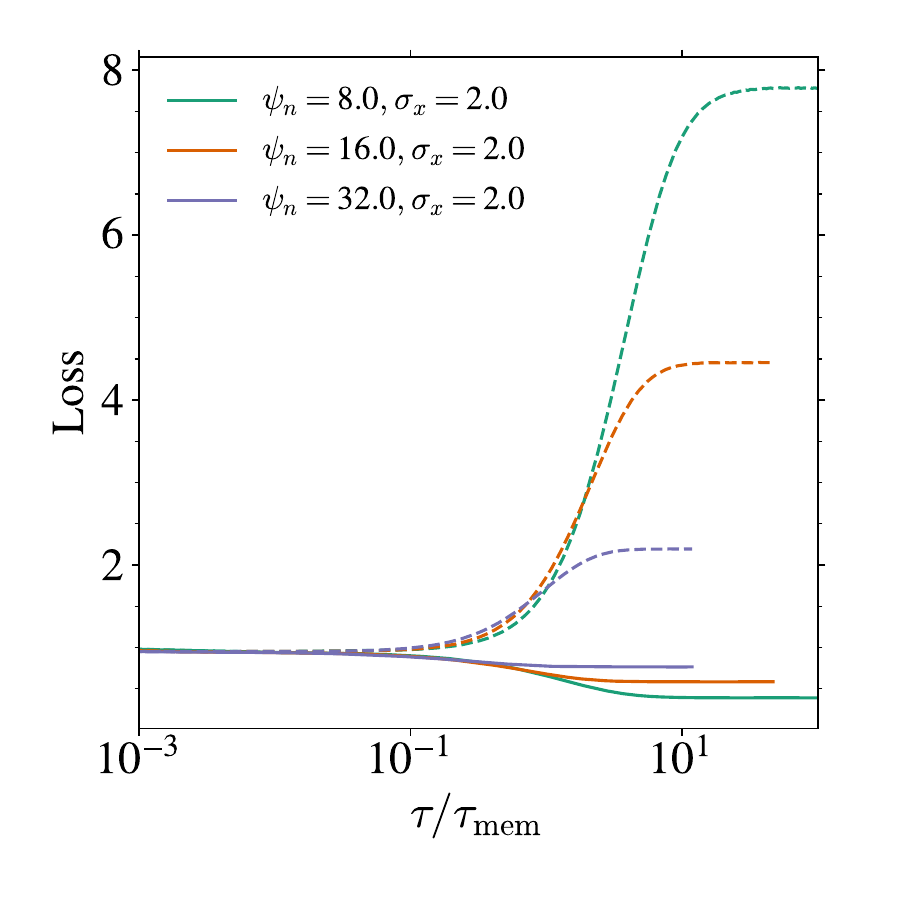}

  \end{minipage}


  \begin{minipage}[b]{0.49\textwidth}
    \centering
    \includegraphics[width=\textwidth]{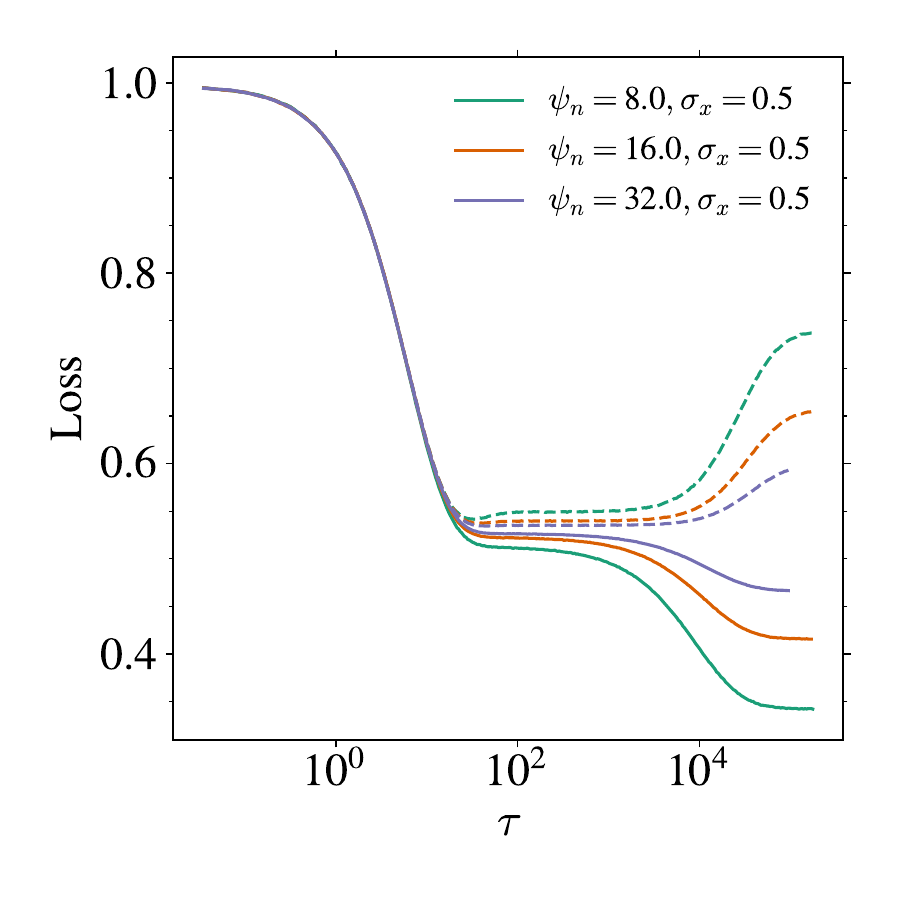}

  \end{minipage}
  \hfill
  \begin{minipage}[b]{0.49\textwidth}
    \centering
    \includegraphics[width=\textwidth]{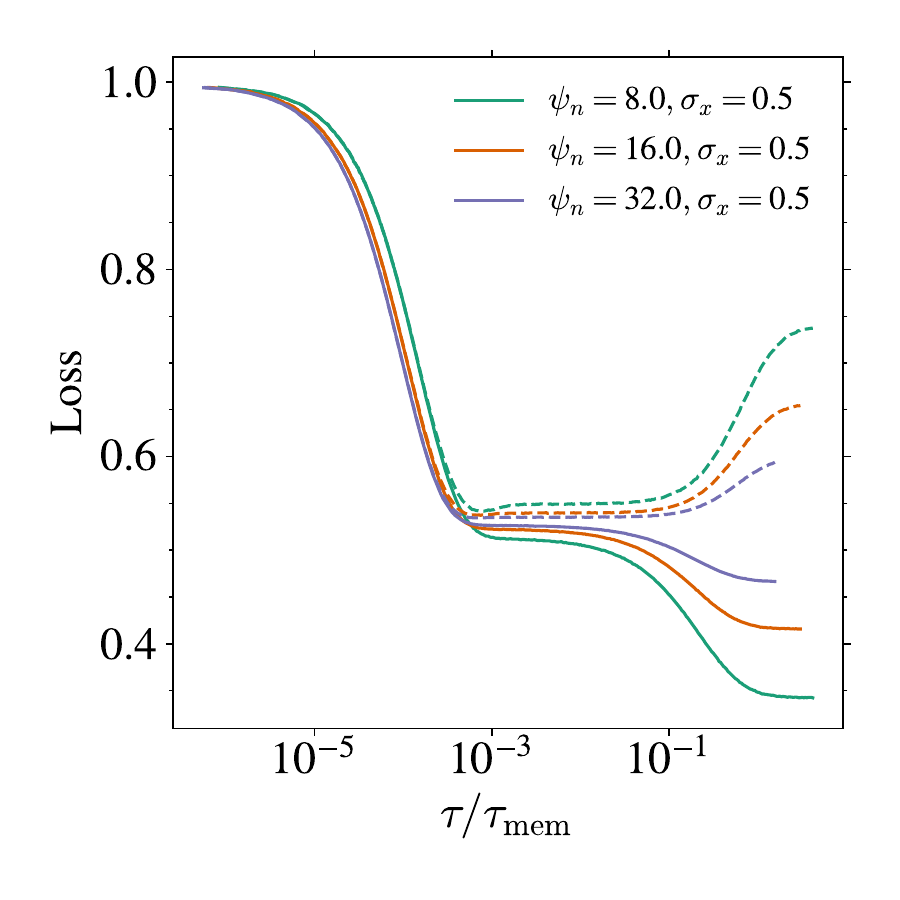}

  \end{minipage}

  \caption{\textbf{Different $\sigma_{\vx}^2$.} Train loss (solid line) and test loss (dotted line) for $\psi_p=64,t=0.1,d=100$, different $\psi_n$ and $\sigma_{\vx}=2.$(top) and $\sigma_{\vx}=0.5$(bottom) against the training time $\tau$ and the rescaled training time $\tau/\tau_{\mathrm{mem}}$.}
  \label{fig:curves:different_sigma_x}
\end{figure}

\paragraph*{Experiments with $\sigma_{\vx}^2\neq1$.}
In Fig.~\ref{fig:curves:different_sigma_x}, we present train and test loss curves for $\sigma_{\vx}\neq1$. We see that our prediction of the timescale of memorization computed in the MT holds for general data variance.

\begin{figure}[htp]
    \centering
    \includegraphics[width=.49\linewidth]{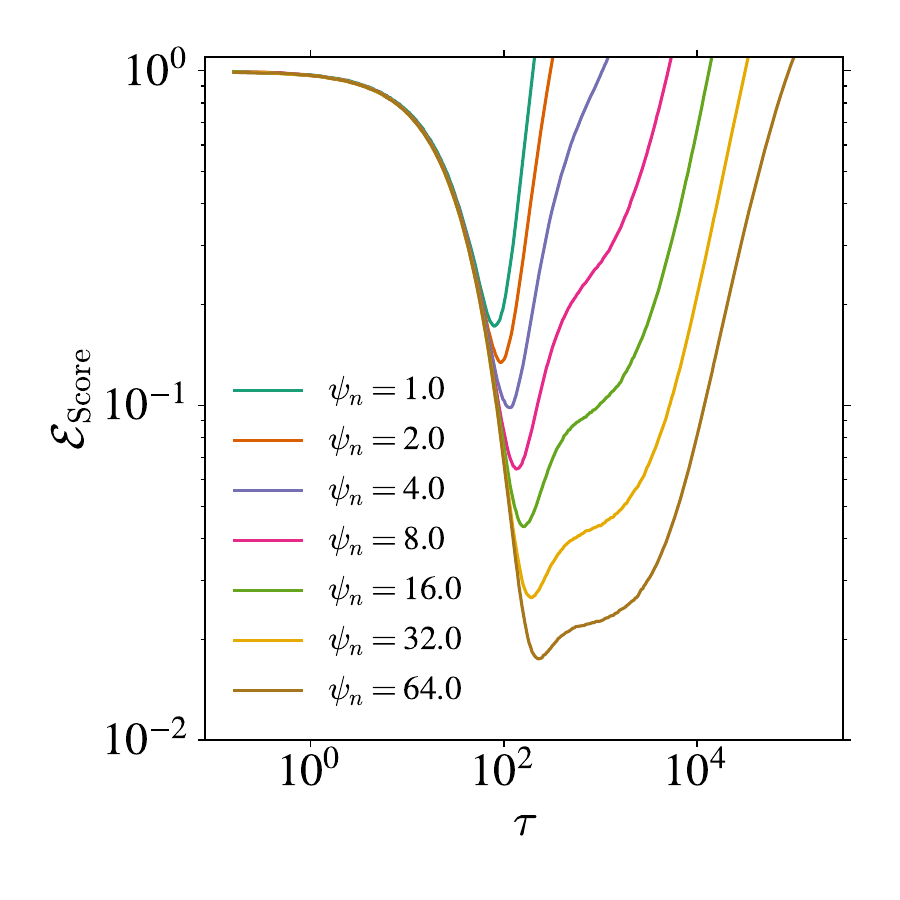}
    \includegraphics[width=.49\linewidth]{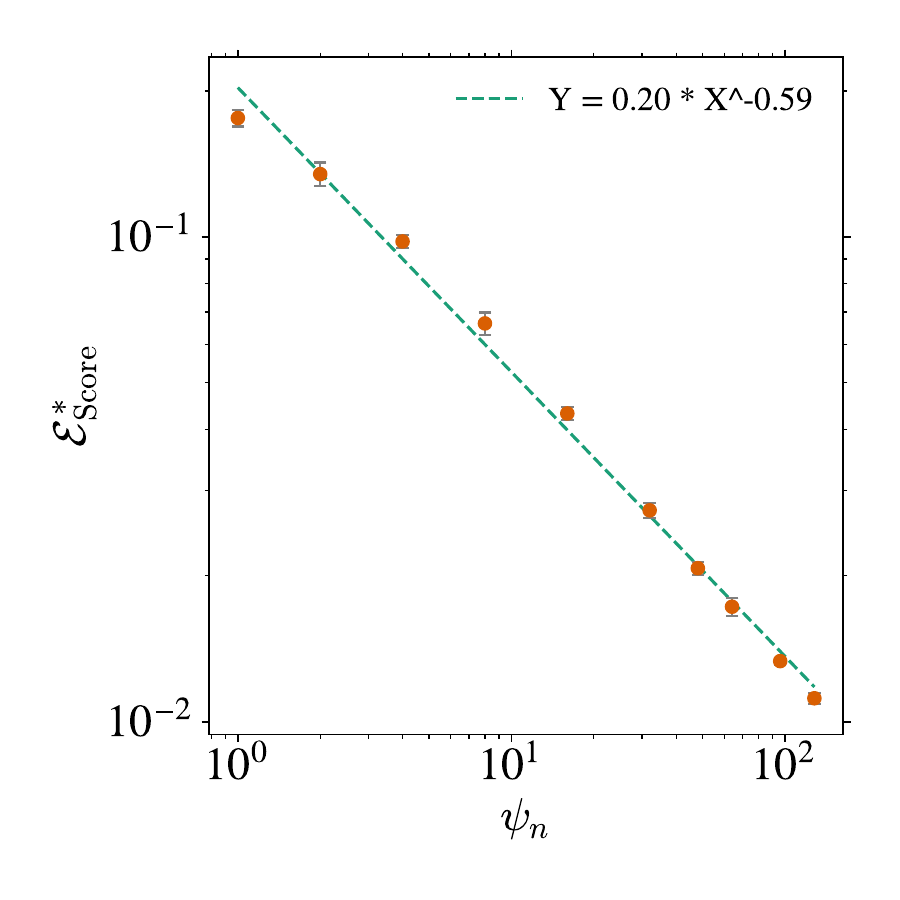}
    \caption{\textbf{Effect of $\psi_n$ on $\mathcal{E}_{\mathrm{Score}}^*$.} (Left) Error between the learned score and the true score $\mathcal{E}_{\mathrm{Score}}$ for $\psi_p = 32$, $t = 0.01$, and various values of $\psi_n$. (Right) Minimum score error $\mathcal{E}_{\mathrm{Score}}^* = \underset{\tau}{\min}[ \mathcal{E}_{\mathrm{Score}}(\tau)]$ as a function of $\psi_n$, showing a power-law decay with exponent approximately $-0.59$. The error bars correspond to thrice the standard deviation over 10 runs with new initial conditions.}
    \label{fig:Effect_n_Min_loss_macris}
\end{figure}

\paragraph*{Scaling of $\mathcal{E}_{\mathrm{score}}$ with $n$.}
In the RF model, the error with respect to the true score, as defined in the main text,
\begin{align}
    \mathcal{E}_{\mathrm{Score}} = \frac{1}{d} \mathbb{E}_{\vy \sim \mathcal{N}(0, \Gamma_t^2 \vI_p)} \left[ \left\lVert \vs_{\vA(\tau)}(\vy) + \frac{\vy}{\Gamma_t^2} \right\rVert^2 \right],
\end{align}
serves as a measure of the generalization capability of the generative process. As shown in \cite{song2021a}, the Kullback–Leibler divergence between the true data distribution $P_{\vx}$ and the generated distribution $\hat{P}$ can be upper bounded
\begin{align}
    \mathcal{D}_{\mathrm{KL}}(P_{\vx} \,\|\, \hat{P}) \leq \frac{d}{2} \int \dd t\, \mathcal{E}_{\mathrm{Score}}(\vA_t),
\end{align}
where the integral is taken over all estimations of the parameter matrix $\vA$ at all diffusion times $t$. This bound assumes that the reverse dynamics are integrated exactly, starting from infinite time. In practical settings, however, one typically relies on an approximate scheme and initiates the reverse process at a large but finite time $T$. A generalization of this bound under such conditions can be found in \cite{bortoli2022convergence}. We have numerically investigate the behaviour of $\mathcal{E}_{\mathrm{score}}$ on Fig.~\ref{fig:Effect_n_Min_loss_macris}. On the fast timescale $\tau_{\mathrm{gen}}$, it decreases until a minimal value $\mathcal{E}_{\mathrm{score}}^*$ that depends only on $\psi_n$ with a power-law $\psi_n^{-\eta}$ with $\eta \simeq 0.59$. We leave for future work performing an accurate numerical estimate of $\eta$ and a developing a theory for it.

\begin{figure}
    \centering
    \includegraphics[width=0.49\linewidth]{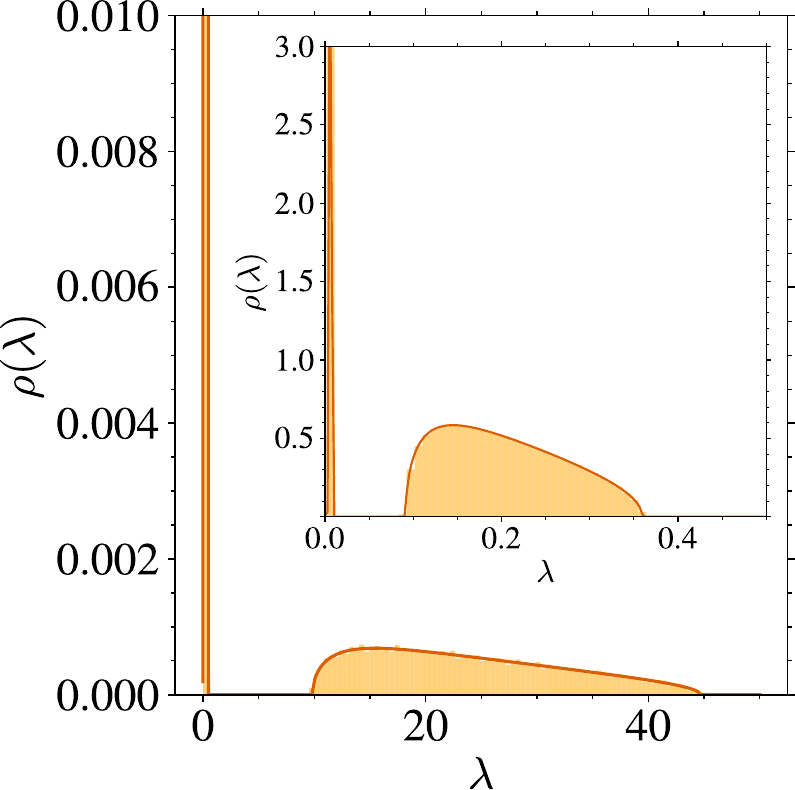}
    \includegraphics[width=0.49\linewidth]{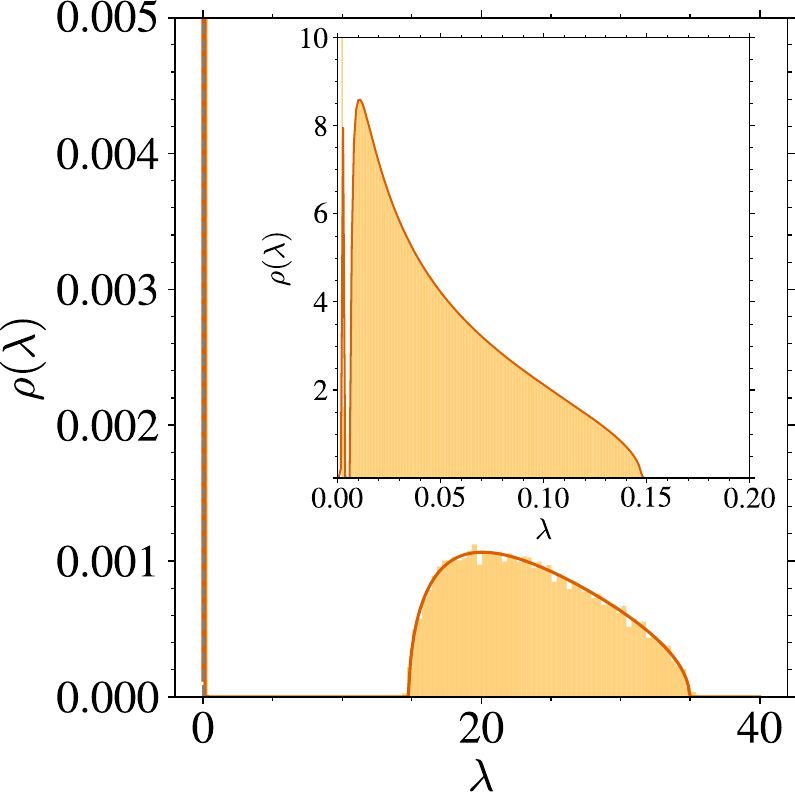}
   \caption{\textbf{Spectrum of $\vU$.} Solutions of the equations in Theorem~3.1. (solid lines) and empirical spectrum for $\rho_{\vSigma}(\lambda)=\delta(\lambda-1)$ and $d=100$ (histogram). (Left) $\psi_p=64$, $\psi_n=8$, $t=0.01$. (Right) $\psi_p=64$, $\psi_n=32$, $t=0.01$.}
    \label{fig:analytical_spectrum_appendix}
\end{figure}

\paragraph*{Spectrum of \texorpdfstring{$\vU$}{U}.} In Fig.~\ref{fig:analytical_spectrum_appendix}, we compare the solutions of the equations of Theorem~\ref{thm:Saddle_point_equations_new} to the histogram of finite size realizations of $\vU$.

\begin{figure}
    \centering
    \includegraphics[width=0.5\linewidth]{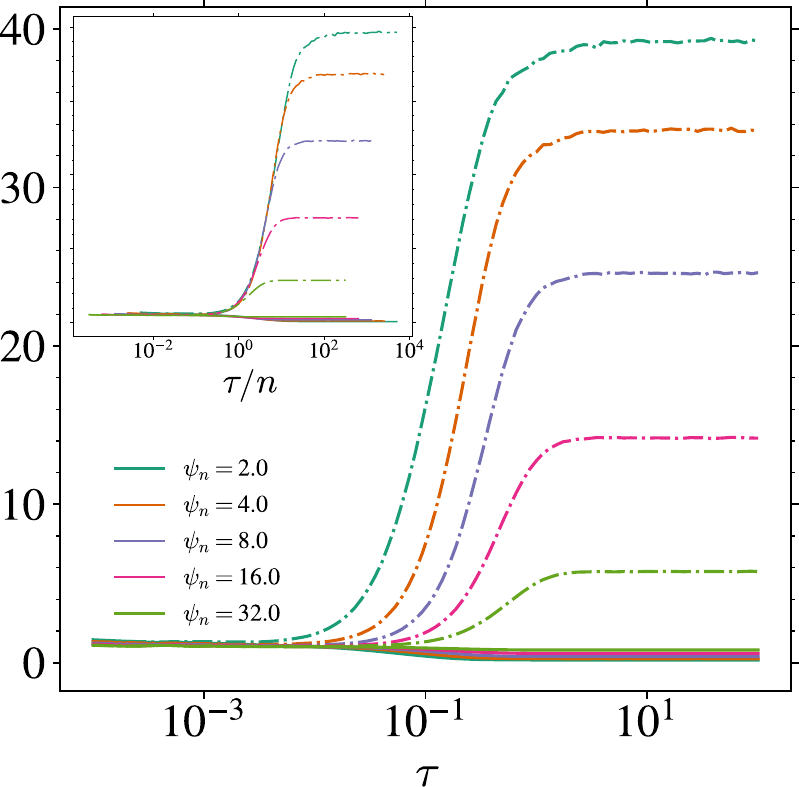}
    \caption{\textbf{Adam.} Train loss (solid line) and test loss (dotted line) at $t=0.01,d=100,\psi_p=64$ for several $\psi_n$ with the Pytorch \cite{pytorch_2019} implementation of Adam. The inset shows the effect of a rescaling of the training time by $n$.}
    \label{fig:Adam_RF}
\end{figure}

\paragraph*{Effect of Adam optimization.} Numerical experiments with RFNN on Gaussian data show that the linear scaling of the memorization time with $n$ holds also for the Adam optimizer as shown in Fig.\ref{fig:Adam_RF}.

\end{document}